\documentclass[10pt,twocolumn,letterpaper]{article}

\usepackage[pagenumbers]{cvpr} 

\usepackage{bm}
\usepackage{amssymb}
\usepackage{amsmath}
\usepackage{mathtools}
\usepackage{amsthm}
\usepackage{mathrsfs}

\DeclareMathAlphabet{\mathsfit}{\encodingdefault}{\sfdefault}{m}{sl}
\SetMathAlphabet{\mathsfit}{bold}{\encodingdefault}{\sfdefault}{bx}{n}

\usepackage{multicol}               
\usepackage{float}                  
\usepackage{booktabs}       
\usepackage{amsfonts}       
\usepackage{nicefrac}       
\usepackage{microtype}      
\usepackage{adjustbox}

\usepackage{bbding}
\usepackage[linesnumbered,ruled]{algorithm2e} 
\usepackage{array}
\usepackage{utfsym}
\usepackage{xcolor}         
\usepackage{color}
\usepackage{colortbl}  
\usepackage{pifont}
\usepackage{enumitem}
\usepackage{mdframed}
\usepackage{footnote}
\usepackage{multirow}

\makeatletter
\newcommand{\thickhline}{
    \noalign {\ifnum 0=`}\fi \hrule height 1pt
    \futurelet \reserved@a \@xhline
}

\definecolor{igray}{gray}{.95}
\definecolor{igray2}{gray}{.80}
\definecolor{A1color}{RGB}{252,63,77} 
\definecolor{A1color2}{RGB}{30,144,255} 
\definecolor{A2color}{RGB}{29,26,178}

\definecolor{cvprblue}{rgb}{0.21,0.49,0.74}
\usepackage[pagebackref,breaklinks,colorlinks,allcolors=cvprblue]{hyperref}

\def\paperID{612} 
\def\confName{CVPR}
\def\confYear{2025}

\title{Correlative and Discriminative Label Grouping for Multi-Label \\ Visual Prompt Tuning}

\author{
	Lei-Lei Ma$^{1}$\thanks{Equal contribution.},\ \
	Shuo Xu$^{1}$\footnotemark[1],\ \
	Ming-Kun Xie$^{2}$\thanks{Correspondence to: Haifeng Zhao (senith@163.com), Ming-Kun Xie (mkxie@nuaa.edu.cn)},\ \
	Lei Wang$^{3}$, \ \
	Dengdi Sun$^{1}$,\ \
	Haifeng Zhao$^{1}$\footnotemark[2]\\
        {\large $^1$Anhui Provincial Key Laboratory of Multimodal Cognitive Computation,}\\
        {\large School of Computer Science and Technology, Anhui University, Hefei, China}\\
	{\large$^2$Nanjing University of Aeronautics and Astronautics, Nanjing, China}\\
	{\large$^3$Nanjing University of Science and Technology, Nanjing, China}
}

\begin{document}

\maketitle
\begin{abstract}
Modeling label correlations has always played a pivotal role in multi-label image classification (MLC), attracting significant attention from researchers. 
However, recent studies have overemphasized co-occurrence relationships among labels, which can lead to overfitting risk on this overemphasis, resulting in suboptimal models.
To tackle this problem, we advocate for balancing correlative and discriminative relationships among labels to mitigate the risk of overfitting and enhance model performance. 
To this end, we propose the Multi-Label Visual Prompt Tuning framework, a novel and parameter-efficient method that groups classes into multiple class subsets according to label co-occurrence and mutual exclusivity relationships, and then models them respectively to balance the two relationships.
In this work, since each group contains multiple classes, multiple prompt tokens are adopted within Vision Transformer (ViT) to capture the correlation or discriminative label relationship within each group, and effectively learn correlation or discriminative representations for class subsets. On the other hand, each group contains multiple group-aware visual representations that may correspond to multiple classes, and the mixture of experts (MoE) model can cleverly assign them from the group-aware to the label-aware, adaptively obtaining label-aware representation, which is more conducive to classification.
Experiments on multiple benchmark datasets show that our proposed approach achieves competitive results and outperforms SOTA methods on multiple pre-trained models.
\vspace{-1.0em}
\end{abstract}    
\section{Introduction}
\label{sec:intro}

Multi-label image classification (MLC) aims to assign multiple labels to a single instance~\cite{zhang2013review}. 
As a fundamental machine learning algorithm~\cite{liu2021emerging}, MLC reflects real-world scenarios and is an essential component in many applications, \eg, face recognition \cite{chen2020label}, and scene understanding~\cite{zhang2024multi}, weakly-supervised semantic segmentation~\cite{chang2020weakly}.
Compared with single-label settings, the MLC presents significant challenges due to the exponential expansion of the output space, which usually requires large-scale datasets~\cite{wei2015hcp}.

In real-world scenarios, co-occurrence between labels is prevalent, and numerous studies~\cite{xie2021partial,zhang2020partial,li2022estimating} have leveraged label relationships to address the challenges resulting from complex label output spaces.
In the context of deep learning, pioneering works~\cite{wang2016cnn,wang2017lstm} employ recurrent neural networks (RNNs) to capture limited correlations between labels, while later studies~\citep{chen2019gcn, wu2020adahgnn} utilize graph convolutional
networks (GCNs) to capture global relationships in graphs or higher-order relationships in hypergraphs.
More recently, for better representations, most researchers employ attention architectures to decouple an image into multiple label-aware features, as seen in methods like SSGRL~\cite{chen2019iccv}, Q2L~\cite{liu2021query2label}, \etc. 

\begin{figure}[t]
    \centering
    \begin{subfigure}{0.49\linewidth}
        \begin{adjustbox}{raise=8pt,center}  
    		\includegraphics[width=0.67\linewidth]{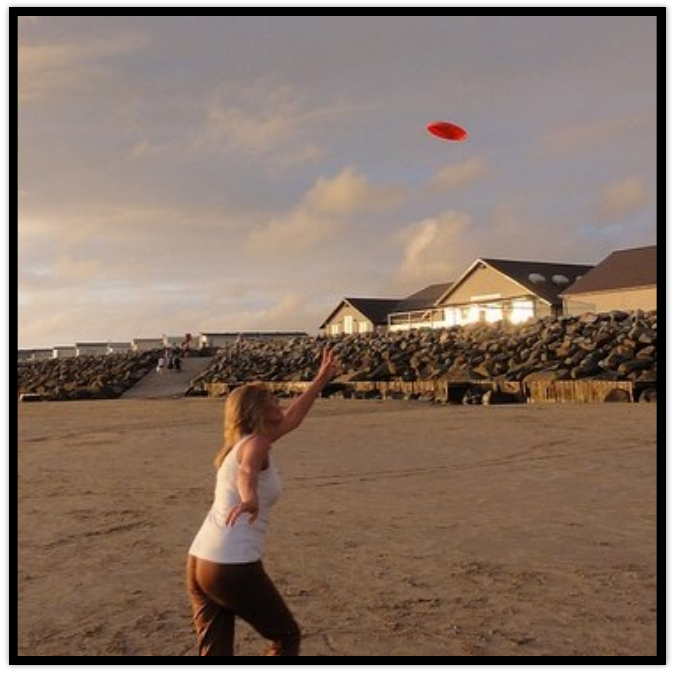}
        \end{adjustbox}
        \vspace{-1.5em}
		\caption{predicted image}
		\label{fig:motivation_subfig1}
    \end{subfigure}
    \hfill
    \begin{subfigure}{0.49\linewidth}
		\includegraphics[width=0.99\linewidth]{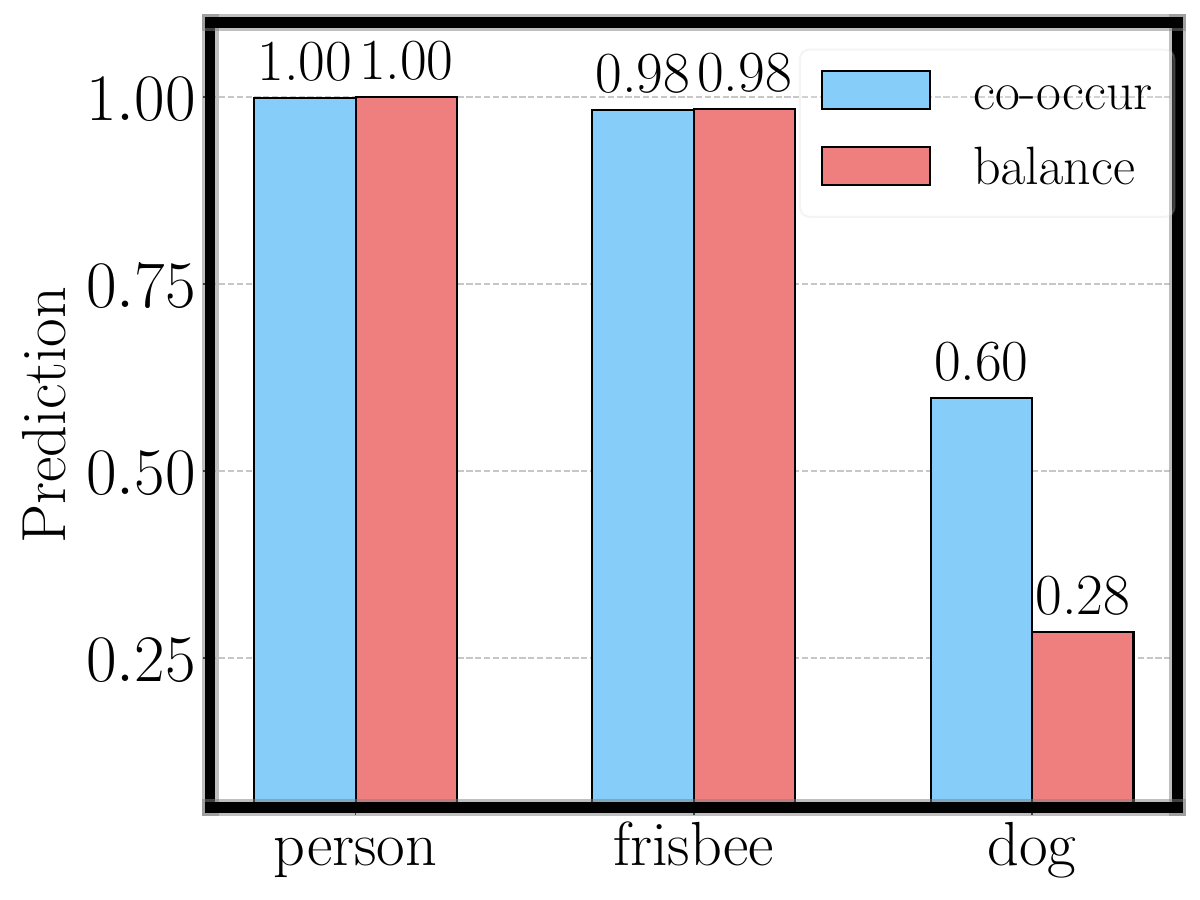}
        \vspace{-1.5em}
		\caption{predicted probability}
		\label{fig:motivation_subfig2}
    \end{subfigure}
\vspace{-1.0em}
\caption{This illustrates the negative impact of co-occurrence. Due to the high co-occurrence probability $p(\text{\ttfamily{dog}}|\text{\ttfamily{frisbee}})$, \eg, 0.23, the model incorrectly infers the presence of \text{\ttfamily{dog}} when the label \text{\ttfamily{frisbee}} appears.}
\label{fig:motivation}
\vspace{-1.0em}
\end{figure}

Despite the great advantages these methods showcase, two significant drawbacks require mitigation.
{\textbf{\ding{182}}} The pre-trained model in these methods needs to be fine-tuned, leading to increased computational and resource demands. Also, for large-scale pre-trained models (\eg, DINOv2~\cite{dinov22023oquab}), full fine-tuning can negatively affect performance on downstream tasks due to object tasks and data distributions \cite{hanfacing}.
{\textbf{\ding{183}}} More importantly, while modeling correlations between labels can enhance model performance, overemphasizing these correlations will lead to the risks of overfitting and may misguide the model to make erroneous inferences, resulting in sub-optimal overall performance~\citep{liu2022contextual}.
\Cref{fig:motivation} verifies this in an empirical example.
Recent attempts to address this issue, such as BoostMLC \cite{xuboosting} and PAT \cite{xiecounterfactual}, still involve considerable computational costs (\eg, multiple teacher models and more augmented images).

Recently, visual prompt tuning (VPT)~\cite{jia2022visual} has emerged as a parameter-efficient fine-tuning (PEFT)~\cite{ding2023parameter,fu2023effectiveness} method for adapting large-scale pre-trained vision transformer models to downstream tasks~\cite{yoo2023improving,wang2024revisiting,han20232vpt}.
Naturally, the success of VPT has inspired us to explore using PEFT to address the first issue. However, existing VPT-like methods primarily downstream tasks in single-label prediction~\cite{gao2022visual,park2024fair} or dense prediction tasks~\cite{kim2024eclipse,hong2024onetracker}, which may not suit MLC where the correlation between labels is not considered during the learning representation phase.

Regarding the second drawback, the key is to mitigate the adverse effects of overemphasizing co-occurrence relationships while preserving their advantages.
A Na\"ive idea is to model the correlative (CO) and discriminative (DC) relationships between labels separately and then balance them.
In terms of CO, a reasonable strategy is to highlight class pairs with high co-occurrence probabilities within a co-occurrence graph~\cite{chen2019gcn,chen2019iccv}, thereby modeling the CO between these classes. 
In contrast, DC tends to emphasize the characteristics of the object class to highlight distinguishability. Therefore, class pairs with low co-occurrence probability may have higher DC relationships (discriminability).
Existing studies~\cite{xuboosting,xiecounterfactual} tend to be one of them or require more complex models and computation. Therefore, to effectively balance these two relationships, we can naturally group the classes into multiple sets according to their co-occurrence probabilities. All classes are placed together in the CO groups via high probabilities. Conversely, for DC groups, it is advocated to divide all classes using low co-occurrence probabilities.

Drawing on the above investigations and analysis, we propose a multi-label visual prompt tuning (ML-VPT) framework that utilizes grouping classes to balance the two relationships simply and effectively.
\textbf{\emph{Firstly}}, we take advantage of VPT and introduce an equal number of visual prompt tokens to each group, which facilitates learning visual representations for each class subset, referred to as group-aware representations. Unlike mainstream methods \cite{chen2019iccv,liu2021query2label}, our approach offers two great advantages: It emphasizes directly modeling label relationships based on multiple prompt tokens for each group within the visual encoder. Our method maintains a balanced focus, not overemphasizing the CO relationship between labels while considering DC.  
\textbf{\emph{Secondly}}, each class subset comprises multiple classes, and each group incorporates multiple visual prompt tokens, resulting in each group containing multiple group-aware representations.  
To more effectively highlight the distinctions between classes, we employ a MoE model to selectively identify suitable visual representations for each class within the relevant group-aware representations.
MoE can simplify the mapping from group-aware to label-aware complexity and enable adaptive mapping tailored to specific images.

Overall, our contribution is two-fold:
1) We propose a novel framework for MLC that groups classes first and subsequently models them within ViT. The proposed method balances the CO and DC relationships to overcome the risk of co-occurrence overfitting and improve model performance.
2) The group-aware MoEs are implemented to map group-aware representations to label-aware ones. In a dynamic way, multiple appropriate experts with gating networks are chosen to construct label-aware representations. 
\section{Related Work}
\noindent \textbf{Multi-Label Image Classification.}
MLC aims to predict multiple labels associated with an instance simultaneously~\cite{xie2023class}.
The mainstream MLC methods can be roughly summarized into two types: modeling label correlations and learning label-aware representations. The former typically employs GCNs or RNNs to capture dependencies between labels~\cite{wang2016cnn,wang2017lstm,chen2019gcn,chen2019iccv}, while the latter often utilizes attention mechanisms to learn category-specific representations, which focus on effectively processing contextual information in images~\cite{ye2020attention,liu2021query2label,mldecoder2023ridnik,ma2023semantic,ma2024text}.
Very recently, limited studies have begun to investigate overfitting and causal interventions in the context of MLC.
For example, CDD~\cite{liu2022contextual} and IDA~\cite{liu2023causality} eliminate the contextual debiasing.
BoostMLC~\cite{xuboosting} and PAT~\cite{xiecounterfactual} attempt to eliminate the risk of overfitting resulting from overemphasizing co-occurrence, but involve complex calculation processes and computational overhead.

\noindent \textbf{Visual Prompt Tuning.}
To strike an optimal balance between computational cost and performance, Jia \etal pioneer VPT~\cite{jia2022visual}, which effectively adapts transformer models pre-trained on large-scale datasets for various downstream tasks.
Subsequently, research initiatives analogous to VPT emerged~\cite{yoo2023improving,wang2024revisiting,han20232vpt}, with many researchers employing VPT across various computer vision applications, such as classification~\cite{park2024fair}, test-time adaptation~\cite{gao2022visual}, segmentation~\cite{kim2024eclipse}, and continual learning~\cite{gao2023unified}.
However, these studies are unsuitable for MLC, as they fail to account for label correlations, whereas our approach explicitly considers them.

\noindent \textbf{Mixture of Experts.}
The MoE~\cite{jacobs1991adaptive} dynamically selects specialized expert networks to process input data, enhancing overall model performance.
MoE has been successfully applied across various fields, 
\eg, reinforcement learning~\cite{zheng2019self}, domain generalization~\cite{li2022sparse}, multimodal large language models~\cite{chen2024llava}, Re-identification~\cite{LYD2024}.
Recently, HQS~\cite{yinhybrid} treats each label as task-specialized and employs the MoE model for learning in MLC tasks.
However, unlike HQS, which assigns experts based on label-aware representation, our approach assigns experts based on group-aware representation, which can balance correlative (CO) and discriminative (DC) relationships to eliminate the risk of overfitting.
\begin{figure*}[ht]
    \centering
    \includegraphics[width=0.90\linewidth]{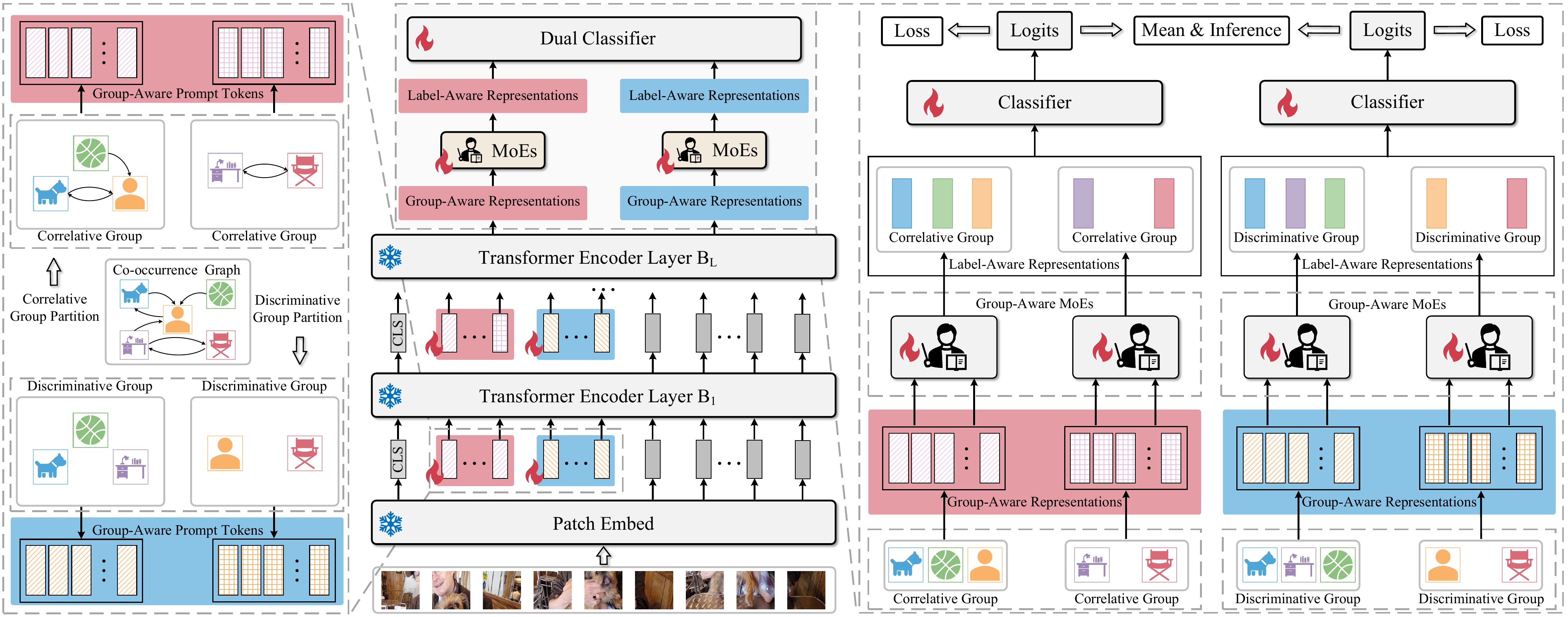}
    \vspace{-0.5em}
    \caption{
    The overview of our framework.
    We apply clustering strategies to the co-occurrence graph to group labels into co-occurrence groups (CO) and discriminative groups (DC). Then, we introduce two types of prompt tokens into ViT, corresponding to CO and DC, respectively. ViT subsequently generates two types of group-aware representations. To adaptively map these representations to label-aware ones, we design Group-Aware MoEs. Finally, two distinct classifiers are employed to classify the label-aware representations.
    }
    \label{fig:framework}
    \vspace{-1.0em}
\end{figure*}
\section{Preliminaries}
\subsection{Notations}
Let $\bm{x}\in \mathcal{X}$ denote the instance, and $\bm{y}\in \mathcal{Y}$ denote the corresponding label, where $\mathcal{X}=\mathbb{R}^d$ is the instance space, and $\mathcal{Y}=\{0,1\}^K$ is the label space containing the $K$ classes.
For a given instance $\bm{x}$, $y^{k} = 1$ denotes $k$-th label is a relevant label associated with the instance $\bm{x}$, and vice versa.
In MLC tasks, given a labeled dataset with $N$ samples $\mathcal{D} = \{(\bm{x}_i,\bm{y}_i)\}_{i=1}^N$.
Our goal is to train a model, denoted as $f(\bm{x}; \theta): \mathcal{X} \rightarrow \mathcal{Y}$, using a training dataset $\mathcal{D}$. The aim is to accurately predict all potential labels $\bm{y}$ for unseen instances $\bm{x}$, with $\theta$ representing the learnable model parameters.
Accordingly, the MLC expected risk function:
\begin{equation}\label{eq:full-cls-expected-risk}
\begin{aligned}
    \mathcal{R}\left(f\right) = ~& \underset{ \left(\bm{x}\!,\bm{y}\right)\sim p\left(\bm{x}\!,\bm{y}\right) }{\mathbb{E}}\left[\mathcal{L}\left(f\left( \bm{x} \right), \bm{y} \right)\right]~\text{,}
\end{aligned}
\end{equation}
where $\mathcal{L}:\mathcal{X} \times \mathcal{Y} \rightarrow \mathbb{R}^{+}$ indicates a multi-label loss function that turns MLC task into multiple more straightforward binary classification problems.

\subsection{Visual Prompt Tuning Revisit}
Visual Prompt Tuning is mainly implemented in ViTs~\cite{dosovitskiy2020image}, which usually consists of a patch embedding layer and $L$ stacked transformer blocks ({\rm Block}).
Given an image $\bm{x}$ that is divided into a set of non-overlapping patches, the patch embedding layer embedded each patch with the position information $\mathbf{E}_{\text{pos}}$ into a $D$-dimensional space, as follows:
\begin{align}
    \mathbf{E}_{0} = {\rm PatchEmbed}(\bm{x}) + \mathbf{E}_{\text{pos}}~\text{.}
\end{align}
Next, image patch embeddings $\mathbf{E}_i = \{\mathbf{e}_i^j \in \mathbb{R}^D | j \in \mathbb{N}, 1\leq j \leq N_{e} \} $ along with a class token $\mathrm{cls}_{i}$ is input into the ($i$+1)-th transformer block:
\begin{align}
    [\mathrm{cls}_{i}, \mathbf{E}_{i}]= {\rm Block}_{i}([\mathrm{cls}_{i-1}, \mathbf{E}_{i-1}])~\text{,}
\end{align}
where $[\cdot,\cdot]$ indicates stacking and concatenation on the sequence length dimension.

For VPT, Jia \etal~\cite{jia2022visual} advocate introducing a set of N learnable prompt tokens as input to $i$-th transformer block (${\rm Block}_{i}$), denoted as $\mathbf{P}_{i-1} = \{\mathbf{p}_{i-1,k} \in \mathbb{R}^D | k \in \mathbb{N}, 1 \leq k \leq N_p\}$. The whole VPT is formulated as follows:
\begin{align}
    [\mathrm{cls}_{i}, \_, \mathbf{E}_{i}]\!=\!{\rm Block}_{i}([\mathrm{cls}_{i-1}, \mathbf{P}_{i-1}, \mathbf{E}_{i-1}])~\text{.}
\end{align}

\subsection{How to Group Labels?}
To group the labels, we take an approach~\cite{xuboosting}, which performs spectral clustering on the co-occurrence graph $\mathcal{G}^{+}$ and dis-occurrence graph $\mathcal{G}^{-}$.
Specifically, we define a co-occurrence matrix $\mathbf{S}\in\mathbb{R}^{K\times K}$ and an affinity matrix, 
\begin{align}
\mathbf{M}=
\begin{cases}
  \mathbf{M}^+= (\sqrt[\tau]{\mathbf{S}}+\sqrt[\tau]{\mathbf{S}}^\top)~/~{2},       & \mathcal{G}=\mathcal{G}^+ \\
  \mathbf{M}^-= \mathbf{I}-(\sqrt[\tau]{\mathbf{S}}+\sqrt[\tau]{\mathbf{S}}^\top)~/~{2}, & \mathcal{G}=\mathcal{G}^-
\end{cases}
\end{align}
where $\tau$ is a hyper-parameter. 
By applying spectral clustering to $\mathbf{M}$, we divide the classes $\mathcal{C}$ into several subsets within both CO and DC, designated as $\{\mathcal{C}^{+}_{t}\}^{N_{c}}_{t=1}$ and $\{\mathcal{C}^{-}_{t}\}^{N_{c}}_{t=1}$, respectively, where $N_{c}$ is the number of subsets in each set.
Additionally, the number of classes $|\mathcal{C}^{+}_{t}|$/$|\mathcal{C}^{-}_{t}|$ within these subsets is expected to vary. More can be found in Appendix.
\section{Method}
In this section, we describe our novel framework, named ML-VPT, with group-aware visual prompt tuning (in \S \ref{sec:groupvpt}) and group-aware MoE (in \S \ref{sec:moe_groupvpt}) for MLC, depicting the overall overview in \cref{fig:framework}.
In the end, we describe in \S \ref{sec:optimization} how our CO and DC jointly optimize and infer.

\subsection{Overview}
To address the co-occurrence overfitting problem as well as high resource demands in training, we propose a novel multi-label learning method with VPT, and the key is to balance CO and DC, effectively reducing the overfitting risk resulting from overemphasized co-occurrence.
To this end, first, we divide all classes into $N_c$ CO groups and $N_c$ DC groups as class subsets, according to the co-occurrence graph.

Secondly, to model CO and DC between labels, we add two sets of prompt tokens inside ViT, called CO and DC tokens $\mathbf{P}^{+}$/$\mathbf{P}^{-}$, by leveraging the capabilities of advanced VPT.
In each subset, these tokens are evenly distributed across several groups, with each group associated with a class subset.
Within the ViT, extended prompt tokens learn CO and DC relationships among labels, aiming to mitigate the adverse effects of CO relationships.
Upon completing those, the ViT generates multiple group-aware representations $\mathbf{Z}^{+}$/$\mathbf{Z}^{-}$ for class subsets in CO and DC groups:
\begin{align}
\mathbf{Z}^{+} \cup \mathbf{Z}^{-} = {\rm ViT}_{\Phi}([\mathbf{P}^{+}, \mathbf{P}^{-}], \bm{x})~\text{.}
\end{align}
These representations are shared among various classes within the same grouping strategy, which complicates the representation from group-aware to label-aware. 

Subsequently, to enhance the learning of label-aware representations $\mathbf{C}^{+}$/$\mathbf{C}^{-}$ from group-aware representations for each class, we use MoEs to adaptively map these representations from group-aware to label-aware:
\begin{align}
\mathbf{C}^{+} \!\!=\! {\rm MoE}\biguplus{\rm Gate}(\mathbf{Z}^{+})\text{,}~~\mathbf{C}^{-} \!\!=\! {\rm MoE}\biguplus{\rm Gate}(\mathbf{Z}^{-})~\text{,}
\end{align}
where $\biguplus$ indicates the weighted sum, MoE refers to the mixture-of-experts model, and Gate is the gating network.

Finally, we employ two sets of classifiers with non-shared parameters, each making predictions in their respective, and then combine the results using a weighted sum.

\subsection{Group-Aware Visual Prompt Tuning}\label{sec:groupvpt}
Learning the relationship between labels has long been a key focus in MLC \cite{chen2019gcn,liu2021query2label}. However, overemphasizing these correlations often results in inferring objects primarily from co-occurring ones, which is unsuitable for all real-world scenarios \cite{xuboosting}. On the other hand, learning discriminative representations for each object independently can lead to incorrect inferences when context or co-occurring objects are absent \cite {wu2020adahgnn}. To address this issue, we decompose label relationships into two aspects: \textbf{\ding{182}} In the CO, learning shared information for each group is utilized to model the relationships between co-occurring labels. {\textbf{\ding{183}}} Conversely, in the DC, learning unique information for each group is employed to define the relationships between labels that do not co-occur. 

Specifically, we define two types of prompt tokens, referred to as CO group prompts $\{\mathbf{p}^{+}_{t}\}^{N_{c}}_{t=1}$ and DC group prompts $\{\mathbf{p}^{-}_{t}\}^{N_{c}}_{t=1}$.
On the one hand, from the perspective of grouping class, each prompt token $\mathbf{p}^{+}_{t}$/$\mathbf{p}^{-}_{t}$ uniquely maps to a specific class subset $\mathcal{C}^{+}_{t}$/$\mathcal{C}^{-}_{t}$. 
On the other hand, in terms of representation learning, the prompt token $\mathbf{p}^{+}_{t}$/$\mathbf{p}^{-}_{t}$ is associated with all classes, contained by $\mathcal{C}^{+}_{t}$/$\mathcal{C}^{-}_{t}$, thus forming a one-to-many relationship. 
For example, if $\mathcal{C}^{+}_{t}$ includes the classes {\ttfamily{\{bicycle;car\}}}, then $\mathbf{p}^{+}_t$ is exclusively associated with these classes, which can indicate vision representations of these classes.
Similarly, $\mathbf{p}^{-}_t$ corresponds to $\mathcal{C}^{-}_{t}$.
In the context of prompt tuning, we introduce a set of CO group prompts $\mathbf{P}_{i-1}^{+} = \{\mathbf{p}_{i-1,t}^{+} \in \mathbb{R}^D | t \in \mathbb{N}, 1 \leq t \leq N_c\}$ and a set of DC group prompts $\mathbf{P}_{i-1}^{-} = \{\mathbf{p}_{i-1,t}^{-} \in \mathbb{R}^D | t \in \mathbb{N}, 1 \leq t \leq N_c\}$ into each transformer block (${\rm Block}_{i}$), which is designed to facilitate the learning of visual representations pertinent to specific class subsets $\mathcal{C}^{+}_{t}$ and $\mathcal{C}^{-}_{t}$\footnote{The index $t$, in class subsets $\mathcal{C}_{t}^{+}$ and $\mathcal{C}_{t}^{-}$, may not be the same. However, for simplicity, we will uniformly denote it as $t$ in our analysis.}.
It is expressed in the following mathematical form:
\begin{align}\label{eq:group-vpt}
[\mathrm{cls}_{i}, \_, \_, \mathbf{E}_{i}] = {\rm Block}_{i}([\mathrm{cls}_{i-1},\mathbf{P}_{i-1}^{+},\mathbf{P}_{i-1}^{-},\mathbf{E}_{i-1}])~\text{.}
\end{align}

Leveraging the powerful long-range modeling advantages of the transformer, each $\mathbf{p}^{+}_{i,t}$/$\mathbf{p}^{+}_{i,t}$ can aggregate semantic information about the class subset $\mathcal{C}_{t}$ from the image embedding $\mathbf{E}_{i}$.
In CO, $\mathbf{p}_{i,t}^{+}$ focuses on learning shared information within the class subset $\mathcal{C}_{t}^{+}$. Conversely, in DC, due to the classes that do not co-occur in subsets $\mathcal{C}_{t}^{-}$, the DC group prompts $\mathbf{p}_{i,t}^{-}$ seek to capture the distinct information unique to each class within the subset.
However, due to MSA, prompt tokens $\mathbf{p}_{i,t}$ may transfer information to each other. To mitigate this effect, we introduce a predefined mask to prevent information exchange between prompt tokens across groups, making them more discriminative.

\subsection{Group-Aware Mixture-of-Experts}\label{sec:moe_groupvpt}
As in \S \ref{sec:groupvpt}, each class subset was associated with one prompt token.
However, this simplification does not accurately represent the real-world scenarios.
For example, in the CO, the classes {\ttfamily\{\text{traffic light};\text{stop sign};\text{car}\}} exhibit a strong co-occurrence relationship and are classified into the same group. However, when only one prompt token is used, it becomes challenging to capture the co-occurrence relationships between labels within the same group.
In contrast, for the DC, the classes {\ttfamily\{\text{skateboard};\text{skis};\text{refrigerator}\}}, which are divided into the same group due to their solid discrimination relationships, tend to show significant visual differences \cite{zhan2017inductive} in the images. 
When only one prompt token is assigned to each group, the model tends to share the group's common semantic information, which in turn impedes its ability to capture the unique representations for each class.

To this end, we expand the number of prompt tokens for each group. Since each class possesses unique characteristics, it is essential to select a group-aware representation that is appropriate for each class. To achieve this, we frame the selection of group-aware representations as a multi-task learning problem, using the MoEs to play its advantages in multi-task learning~\cite{li2022sparse}. And each subtask is associated with multiple group-aware representations, which are then mapped to label-aware representations via the MoE model. 

\noindent \textbf{More Prompt Tokens for Each Group.}
Specifically, we extend each prompt token $\mathbf{p}_{i-1,t}$ from 1 to $N_{m}$, defined as follows: $\mathbf{p}_{i-1,t} = \{\mathbf{p}_{i-1,t}^{e} \in \mathbb{R}^D | e \in \mathbb{N}, 1 \leq e \leq N_m\}$.
Moreover, $\mathbf{p}_{i-1,t}^{e,+}$ indicates the  correlative group prompt tokens, while $\mathbf{p}_{i-1,t}^{e,-}$ is the opposite.
To reduce additional parameters, we advocate introducing MoE after the last transformer block.
Then, the final transformer block’s output can be expressed as: 
$[\mathrm{cls}_{o}, \mathbf{Z}^{+}, \mathbf{Z}^{-}, \mathbf{E}_{o}]$, where $\mathbf{Z}^{+}$ corresponds to the output of the CO and $\mathbf{Z}^{-}$ is the opposite. More specifically, $\mathbf{z}_{t}^{e,+}$ is related to the $e$-th prompt token in the $t$-th discriminative group.
And, we refer to $\mathbf{z}_{t}^{e}$ as the $e$-th representation corresponding to the class subset $\mathcal{C}_{t}$.

\noindent \textbf{Mixture-of-Experts.}
Subsequently, we introduce correlative and discriminative group experts $\mathcal{E}$, and each expert $E_{t}^{e}(\cdot)$ is implemented as an FFN with a ReLU activation and residual connection:
\begin{align}
    \mathcal{E}^{+} = \{ {E}_{t}^{e,+} | e, t \in \mathbb{N}, 1 \leq e \leq N_m, 1 \leq t \leq N_c\}~\text{,}\\
    \mathcal{E}^{-} = \{ {E}_{t}^{e,-} | e, t \in \mathbb{N}, 1 \leq e \leq N_m, 1 \leq t \leq N_c\}~\text{,}\\
    E_{t}^{e}(\mathbf{z}_{t}^{e}) = \mathbf{z}_{t}^{e} \!+\! \sigma(\mathbf{z}_{t}^{e} \mathbf{W}_{t,{\rm dn}}^{e} \!+\! \mathbf{b}_{t,{\rm dn}}^{e}) \mathbf{W}_{t,{\rm up}}^{e} \!+\! \mathbf{b}_{t,{\rm up}}^{e}~\text{,}
\end{align}
where $\mathbf{W}_{t}^{e}$ and $\mathbf{b}_{t}^{e}$ are the weights and biases of the experts, and $\sigma(\cdot)$ is a activation.
To make lightweightness, the hidden dimension is kept small, \eg, 5. 
The two groups of experts don’t share parameters.

\noindent \textbf{Gating Networks.}
We employ two distinct strategies for expert selection: {\textbf{\ding{182}}} Each subset within the grouping class utilizes a routing network to learn varying weights for multiple group-aware representations $\mathbf{z}_{t}^{e}$, followed by weighted merging by multiple experts. {\textbf{\ding{183}}}
The second strategy considers each class independently; here, different classes within the corresponding class subset group employ different dynamic routing networks to determine diverse weights for multiple visual representations $\mathbf{z}_{t}^{e}$, subsequently merging these using multiple experts.
We opt for the second strategy, as different classes necessitate distinct information, rendering identical shared visual representations insufficient.

For each class $k$, we define a gating network $g^{+}_{k}(\cdot)$/$g^{-}_{k}(\cdot)$ in CO and DC groups, respectively, implemented as a simple fully connected network:
\begin{align}
    \bm{w}_{k}^{+} &= {\rm softmax}(\mathbf{z}_{t}^{e,+}\mathbf{W}_{k}^{+} + \mathbf{b}_{k}^{+} ) \in \mathbb{R}^{N_{m}}\text{,}~ k \in \mathcal{C}_{t}^{+}~\text{,} \\
    \bm{w}_{k}^{-} &= {\rm softmax}(\mathbf{z}_{t}^{e,-}\mathbf{W}_{k}^{-} + \mathbf{b}_{k}^{-} ) \in \mathbb{R}^{N_{m}}\text{,}~ k \in \mathcal{C}_{t}^{-}~\text{,}
\end{align}
where $\bm{w}_{k}^{+} \!=\! \{w_{k}^{e,+}\}^{N_e}_{e=1}$ and $\bm{w}_{k}^{-} \!=\! \{w_{k}^{e,-}\}^{N_e}_{e=1}$  denote the weights associated with class $k$ in class subset $\mathcal{C}^{+}_{t}$ and $\mathcal{C}^{-}_{t}$.
%

\noindent \textbf{Adaptive Label-Aware Representation.}
Adaptively aggregate group-aware representations into label-aware representations. The group-aware MoE is formalized as follows:
\begin{align}
     \mathbf{c}_{k}^{+} & = \sum\nolimits_{e=1}^{N_e} {w}_{k}^{e,+} E_{t}^{e,+}(\mathbf{z}_{t}^{e,+})~\text{,} \\
     \mathbf{c}_{k}^{-} & = \sum\nolimits_{e=1}^{N_e} {w}_{k}^{e,-} E_{t}^{e,-}(\mathbf{z}_{t}^{e,-})~\text{.}
\end{align}
where $\mathbf{c}_{k}^{+}$ and $\mathbf{c}_{k}^{-}$ indicate the label-aware representations for class $k$, obtained from CO and DC groups, respectively.
\subsection{Optimization and Inference}\label{sec:optimization}
In this work, we adapt dual classification heads to mitigate the negative impacts arising from the accumulated prediction errors from a single classification head \cite{xiecounterfactual}.
The logits for both correlative and discriminative groups can be predicted using two linear classifiers with a sigmoid as follows:
\begin{align}
    \hat{{y}}_{k}^{+} = \sigma(\mathbf{W}_{k}^{\top}\mathbf{c}_{k}^{+} + {b}_{k}^{+})~, \quad
    \hat{{y}}_{k}^{-} = \sigma(\mathbf{W}_{k}^{\top}\mathbf{c}_{k}^{-} + {b}_{k}^{-})~.
\end{align}
Here, $\hat{{y}}_{k}^{+}$ and $\hat{{y}}_{k}^{-}$ represent the predictions for class $k$ in the two types of , while $f^{+}(\bm{x})= \hat{\bm{y}}^{+}$ and $f^{-}(\bm{x}) = \hat{\bm{y}}^{-}$ denote the predictions for all classes about instance $\bm{x}$, respectively.

Referring to Eq.~\eqref{eq:full-cls-expected-risk}, the overall expected risk associated with our proposed method can be reformulated as follows:
\begin{align}\label{eq:our-full-cls-expected-risk}
\mathcal{R}\!\left(f\right)\!=\!\!\underset{ \left(\bm{x}\!,\bm{y}\right)\sim p\left(\bm{x}\!,\bm{y}\right) }{\mathbb{E}}\!\!\left[\mathcal{L}\left(f^{+}\left(\bm{x} \right), \bm{y} \right)\!\!+\!\mathcal{L}\left(f^{-}\left(\bm{x} \right), \bm{y} \right)\right]\text{,}
\end{align}
where $\mathcal{L}(\cdot,\cdot)$ is ASL loss function~\cite{ben2022multi}.
Given an instance $\bm{x}$, the model's final prediction is $f(\bm{x})= \hat{\bm{y}} = 0.5 \cdot (\hat{\bm{y}}^{+} + \hat{\bm{y}}^{-})$.
\section{Experiments}

\begin{table*}[!ht]
\renewcommand\arraystretch{0.55}
\centering
\caption{Comparison of our method with SOTA models on \texttt{COCO} at 224 $\times$ 224 and 448 $\times$ 448 resolution. All metrics are in \%.
Since mAP, CF1, and OF1 are among the most important evaluation metrics, they are highlighted in dark gray in our method.}
\vspace{-1.0em}
\resizebox{0.95\linewidth}{!}
{\begin{tabular}{c|c|c|c|c|c|c|c|c|c|c|c|c|c|c|c|c} 
\hline\thickhline
\multicolumn{1}{c|}{\multirow{2}{*}{\cellcolor{igray}}}
&
\multicolumn{1}{c|}{\multirow{2}{*}{\cellcolor{igray}}}
&
\multicolumn{1}{c|}{\multirow{2}{*}{\cellcolor{igray}}}
&
%
\multicolumn{7}{c|}{\cellcolor{igray}Resolution: 224 $\times$ 224} & \multicolumn{7}{c}{\cellcolor{igray}Resolution: 448 $\times$ 448} \\ 
\multirow{-2.5}{*}{\cellcolor{igray}Method} & 
\multirow{-2.5}{*}{\cellcolor{igray}Backbone} &  \multirow{-2.5}{*}{\cellcolor{igray}Pre-trained Data} &\cellcolor{igray}mAP &  \cellcolor{igray}CP & \cellcolor{igray}CR & \cellcolor{igray}CF1 & \cellcolor{igray}OP & \cellcolor{igray}OR & \cellcolor{igray}OF1 &  \cellcolor{igray}mAP & \cellcolor{igray}CP & \cellcolor{igray}CR & \cellcolor{igray}CF1 & \cellcolor{igray}OP & \cellcolor{igray}OR & \cellcolor{igray}OF1 \\ 
\hline
\midrule
VPT     & \multicolumn{1}{c|}{\multirow{4}{*}{}}     & \multicolumn{1}{c|}{\multirow{4}{*}{}} & 78.0 & 71.7 & 72.3 & 72.0 & 74.6 & 75.4 & 75.0 & 82.6 & 76.1 & 76.7 & 76.4  & 78.2 & 79.1 & 78.7 \\
GateVPT &       &                                        & 75.6 & 69.7 & 70.3 & 70.0 & 72.9 & 73.7 & 73.3 & 80.4 & 74.2 & 74.7 & 74.4  & 76.6 & 77.5 & 77.0 \\
E2VPT   &       &                                        & 77.3 & 71.6 & 72.2 & 71.9 & 74.5 & 75.4 & 75.0 & 81.7 & 75.3 & 75.9 & 75.6  & 77.7 & 78.5 & 78.1 \\
Ours    & \multirow{-5.3}{*}{ViT-B}     & \multirow{-5.3}{*}{ImageNet 1K}        & \cellcolor{igray2}79.6 & 73.1 & 73.6 & \cellcolor{igray2}73.3 & 75.7 & 76.5 & \cellcolor{igray2}76.1 & \cellcolor{igray2}83.6 & 76.9 & 77.5 & \cellcolor{igray2}77.2  & 79.1 & 79.9 & \cellcolor{igray2}79.5 \\
\midrule
VPT     & \multicolumn{1}{c|}{\multirow{4}{*}{}}       & \multicolumn{1}{c|}{\multirow{4}{*}{}} & 71.0 & 66.0 & 66.6 & 66.3 & 71.1 & 72.0 & 71.6 & 72.2 & 66.9 & 67.6 & 67.3 & 72.3 & 73.1 & 72.7 \\
GateVPT &         &                                        & 66.5 & 62.4 & 62.9 & 62.6 & 68.2 & 69.0 & 68.6 & 69.1 & 64.5 & 65.1 & 64.8 & 70.1 & 70.9 & 70.5  \\
E2VPT   &         &                                        & 69.8 & 65.0 & 65.6 & 65.3 & 70.4 & 71.2 & 70.8 & 73.0 & 67.8 & 68.4 & 68.1 & 72.8 & 73.6 & 73.2  \\
Ours    & \multirow{-5.3}{*}{MAE}       &  \multirow{-5.3}{*}{ImageNet 1K}       & \cellcolor{igray2}75.2 & 69.3 & 69.9 & \cellcolor{igray2}69.6 & 74.0 & 74.9 & \cellcolor{igray2}74.4 & \cellcolor{igray2}78.8 & 72.8 & 73.4 & \cellcolor{igray2}73.1 & 76.8 & 77.6 & \cellcolor{igray2}77.2  \\
\midrule
VPT     & \multicolumn{1}{c|}{\multirow{4}{*}{}}   & \multicolumn{1}{c|}{\multirow{4}{*}{}} & 73.0 & 67.6 & 68.1 & 67.9 & 71.8 & 72.6 & 72.2 & 75.9 & 70.2 & 70.8 & 70.5 & 74.1 & 74.9 & 74.5 \\
GateVPT &    &                                        & 70.9 & 65.7 & 66.3 & 66.0 & 70.3 & 71.2 & 70.8 & 74.9 & 69.4 & 70.0 & 69.7 & 73.4 & 74.2 & 73.8  \\
E2VPT   &    &                                        & 72.7 & 67.3 & 67.9 & 67.6 & 71.7 & 72.5 & 72.1 & 76.0 & 70.3 & 70.9 & 70.6 & 74.2 & 75.0 & 74.6  \\
Ours    & \multirow{-5.3}{*}{MoCo v3}   & \multirow{-5.3}{*}{ImageNet 1K}        & \cellcolor{igray2}75.1 & 69.1 & 69.7 & \cellcolor{igray2}69.4 & 73.2 & 74.0 & \cellcolor{igray2}73.6 & \cellcolor{igray2}77.3 & 71.4 & 72.0 & \cellcolor{igray2}71.7 & 75.2 & 76.0 & \cellcolor{igray2}75.6  \\
\midrule
VPT     & \multicolumn{1}{c|}{\multirow{4}{*}{}} & \multicolumn{1}{c|}{\multirow{4}{*}{}} & 81.0 & 74.5 & 75.1 & 74.8 & 77.3 & 78.1 & 77.7 & 85.0 & 78.4 & 79.0 & 78.7  & 80.5 & 81.4 & 81.0\\
GateVPT &  &                                        & 80.8 & 74.2 & 74.8 & 74.5 & 76.9 & 77.8 & 77.3 & 84.5 & 77.8 & 78.4 & 78.1  & 80.1 & 80.9 & 80.5 \\
E2VPT   &  &                                        & 81.9 & 75.4 & 76.0 & 75.7 & 77.9 & 78.8 & 78.3 & 85.2 & 78.6 & 79.2 & 78.9  & 80.7 & 81.6 & 81.1 \\
Ours    & \multirow{-5.3}{*}{ViT-B-21k} & \multirow{-5.3}{*}{ImageNet 21K}       & \cellcolor{igray2}83.0 & 76.2 & 76.7 & \cellcolor{igray2}76.4 & 78.6 & 79.5 & \cellcolor{igray2}79.0 & \cellcolor{igray2}86.4 & 79.7 & 80.3 & \cellcolor{igray2}80.0  & 81.6 & 82.5 & \cellcolor{igray2}82.0 \\
\midrule
VPT     & \multicolumn{1}{c|}{\multirow{4}{*}{}}  & \multicolumn{1}{c|}{\multirow{4}{*}{}} & 86.1 & 79.5 & 80.1 & 79.8 & 81.9 & 82.8 & 82.4 & 89.7 & 83.2 & 83.8 & 83.5  & 85.2 & 86.1 & 85.7 \\
GateVPT &   &                                        & 85.6 & 79.0 & 79.6 & 79.3 & 81.6 & 82.5 & 82.0 & 89.1 & 82.5 & 83.1 & 82.8  & 84.5 & 85.5 & 85.0 \\
E2VPT   &   &                                        & 86.3 & 79.8 & 80.4 & 80.1 & 82.2 & 83.1 & 82.6 & 89.6 & 83.1 & 83.7 & 83.4  & 85.2 & 86.1 & 85.6 \\
Ours    & \multirow{-5.3}{*}{DINOv2/B}  & \multirow{-5.3}{*}{LVD-142M \cite{dinov22023oquab}}           & \cellcolor{igray2}87.5 & 80.8 & 81.4 & \cellcolor{igray2}81.1 & 83.0 & 83.9 & \cellcolor{igray2}83.4 & \cellcolor{igray2}90.6 & 84.2 & 84.8 & \cellcolor{igray2}84.5 & 86.0 & 86.9 & \cellcolor{igray2}86.4 \\
\midrule
VPT     & \multicolumn{1}{c|}{\multirow{4}{*}{}}  & \multicolumn{1}{c|}{\multirow{4}{*}{}} & 80.1 & 73.8 & 74.4 & 74.1 & 77.1 & 77.9 & 77.5 & 84.5 & 78.2 & 78.8 & 78.5 & 81.1 & 82.0 & 81.5 \\ 
GateVPT &   &                                        & 79.4 & 73.2 & 73.8 & 73.5 & 76.5 & 77.3 & 76.9 & 83.3 & 77.0 & 77.7 & 77.3  & 80.0 & 80.9 & 80.4 \\
E2VPT   &   &                                        & 80.6 & 74.2 & 74.8 & 74.5 & 77.5 & 78.4 & 77.9 & 84.3 & 78.1 & 78.8 & 78.5  & 81.0 & 81.9 & 81.4 \\
Ours    & \multirow{-5.3}{*}{DINOv2/S}  & \multirow{-5.3}{*}{LVD-142M \cite{dinov22023oquab}}           & \cellcolor{igray2}83.4 & 76.6 & 77.1 & \cellcolor{igray2}76.8 & 79.6 & 80.5 & \cellcolor{igray2}80.0 & \cellcolor{igray2}87.4 & 80.8 & 81.4 & \cellcolor{igray2}81.1  & 83.2 & 84.1 & \cellcolor{igray2}83.7\\
\bottomrule
\end{tabular}}
\label{tab:coco}
\vspace{-1.0em}
\end{table*}

\subsection{Experimental Settings}
\noindent \textbf{Dataset and Evaluation Metric.}
In this work, we evaluate the effectiveness of the proposed method on four benchmark datasets, including Pascal VOC 2007 (\texttt{VOC07}) \cite{everingham2010pascal}, MS-COCO 2014 (\texttt{COCO}) \cite{lin2014microsoft}, NUS-WIDE (\texttt{NUS}) \cite{chua2009nus}, and Visual Genome (\texttt{VG256}) \cite{krishna2017visual}. 
For a fair comparison, we utilize widely adopted metrics: Mean Average Precision (mAP) across all classes.
Additionally, consistent with previous works \cite{chen2019gcn, ma2023semantic}, we also showcase overall precision (OP), recall (OR), F1-measure (OF1), as well as per-category precision (CP), recall (CR), and F1-measure (CF1) for detailed comparisons. 
It is important to emphasize that mAP, OF1, and CF1 are the most important metrics among these. 

\noindent \textbf{Implementation Details.}
Following previous MLC methods \cite{ridnik2021asymmetric,liu2021query2label,ma2023semantic}, we adopt a similar experimental setup.
The AdamW \cite{loshchilov2018decoupled} optimizer with the one-cycle policy lr\_schedule \cite{smith2019super} is applied to train the model with maximal learning rate of 0.0005.
All models are trained for 40 epochs with the early stopping.
The batch size is set to 64.
For data augmentation, we apply RandAugment \cite{cubuk2020randaugment} and Cutout \cite{devries2017improved}.
To make the model more robust, we apply an exponential moving average to the model parameters $\theta$, using a decay rate of 0.9997.

\noindent \textbf{Pre-trained Models.}
To verify the robustness of our proposed method, we utilized a series of pre-trained models as backbone networks, including ViT \cite{dosovitskiy2020image}, MAE \cite{he2022masked}, MoCo v3 \cite{chen2021empirical}, and DINOv2 \cite{dinov22023oquab}.
For ViT, we employ ViT-B and ViT-B-21k, supervised training on ImageNet 1k \cite{russakovsky2015imagenet}, and ImageNet 21k \cite{russakovsky2015imagenet} respectively.
MAE and MoCo v3 are trained on ImageNet 1k for autoregressive and contrastive self-supervised training, respectively.
For DINO v2, we utilize DINOv2/B and DINOv2/S, the former is the base model and the latter is the small variant, trained on a larger dataset for self-supervised learning.

\noindent \textbf{Comparing Methods.}
To evaluate the performance of our model, we carried out a comprehensive set of experiments involving the following methods: VPT \cite{jia2022visual}, GateVPT \cite{yoo2023improving}, and E2VPT \cite{han20232vpt}.
These methods are primarily designed for single-label tasks, and the $\mathrm{cls}$ token output of the model is input into a linear classifier and optimized using the ASL~\cite{ben2022multi} to adapt it to multi-label tasks.
In our proposed methods, by default, the number of groups in the CO and DC groups is 5, and the number of experts in each group is 3.
In the work, all comparing methods are setup and compared fairly.

\subsection{Compared to State-of-the-Art (SOTA) Results}
\noindent \textbf{Performance on MS-COCO 2014.}
In \cref{tab:coco}, we report the comparison results between our method and the state-of-the-art method on the \texttt{COCO}.
This table demonstrates that, under the same setting (\ie, the same pre-trained backbone and resolution), our method generally surpasses SOTA methods in terms of mean mAP, CF1, and OF1 at both 224$\times$224 and 448$\times$448 resolutions.
Specifically, at the 224$\times$224 resolution, our method achieves improvements over the SOTA methods as follows: the mAP increases by 1.2\%-8.7\%, the CF1 by 0.7\%-7.0\%, and the OF1 by 0.7\%-5.8\%.
Similarly, at the 448$\times$448 resolution, improvements are noted in the mAP from 0.4\% to 6.6\%, the CF1 from 0.8\% to 5.8\%, and the OF1 score from 0.7\% to 4.5\%, also compared with the SOTA method.
These results demonstrate the effectiveness of our method. Moreover, it should be noted that GateVPT and E2VPT, which are modified from VPT, do not consistently outperform VPT. For instance, at a 224$\times$224 resolution using ViT-B, MAE, and MoCo v3, VPT demonstrates superior performance compared to these two modifications. A potential explanation could be that GateVPT and E2VPT are tailored for single-label tasks and do not account for label correlations during the feature extraction phase.
The best mAP is 90.6\% achieved by our method, using DINOv2/B at a resolution of 448$\times$448, which can be attributed to the superior capabilities of DINOv2/B and higher resolution.

\begin{table}[!t]
\renewcommand\arraystretch{0.50}
\centering
\caption{Comparison of our method with SOTA models on \texttt{VOC07} at 224 $\times$ 224 and 448 $\times$ 448 resolution. All metrics are in \%.}
\vspace{-1.0em}
\resizebox{0.9\linewidth}{!}
{\begin{tabular}{c|c|c|c|c|c|c|c} 
\hline\thickhline
\multicolumn{1}{c|}{\multirow{2}{*}{\cellcolor{igray}}}
&
\multicolumn{1}{c|}{\multirow{2}{*}{\cellcolor{igray}}}
& 
\multicolumn{3}{c|}{\cellcolor{igray}Resolution: 224} & \multicolumn{3}{c}{\cellcolor{igray}Resolution: 448} \\ 
\multirow{-2.5}{*}{\cellcolor{igray}Method} & \multirow{-2.5}{*}{\cellcolor{igray}Backbone} & \cellcolor{igray}mAP &  \cellcolor{igray}CF1 & \cellcolor{igray}OF1 & \cellcolor{igray}mAP & \cellcolor{igray}CF1 & \cellcolor{igray}OF1 \\
\hline
\midrule
VPT      & \multicolumn{1}{c|}{\multirow{4}{*}{}}       & 90.4 & 83.4 & 85.7 & 93.4 & 87.4 & 89.1 \\
GateVPT  &        & 87.4 & 80.9 & 82.9 & 89.2 & 83.1 & 85.2 \\
E2VPT    &        & 85.8 & 79.4 & 83.7 & 92.4 & 86.5 & 88.3 \\
Ours     & \multirow{-5.5}{*}{ViT-B}       & 92.9 & 86.4 & 88.1 & 94.3 & 88.6 & 90.1 \\
\midrule
VPT      & \multicolumn{1}{c|}{\multirow{4}{*}{}}         & 80.3 & 74.1 & 79.2 & 82.8 & 75.9 & 81.6 \\
GateVPT  &         & 85.1 & 75.4 & 80.5 & 86.1 & 79.2 & 84.2 \\
E2VPT    &         & 83.3 & 76.7 & 81.5 & 84.6 & 78.2 & 83.2 \\
Ours     & \multirow{-5.5}{*}{MAE}         & 89.2 & 82.7 & 86.4 & 90.4 & 83.9 & 87.4 \\
\midrule
VPT      & \multicolumn{1}{c|}{\multirow{4}{*}{}}     & 86.2 & 79.7 & 83.6 & 88.4 & 82.3 & 85.7 \\ 
GateVPT  &       & 83.5 & 77.1 & 81.2 & 87.7 & 81.1 & 84.7 \\
E2VPT    &       & 85.9 & 79.2 & 83.1 & 88.8 & 82.4 & 85.6 \\
Ours     & \multirow{-5.5}{*}{MoCo v3}     & 91.1 & 84.2 & 86.8 & 92.7 & 86.3 & 88.6 \\
\midrule
VPT      & \multicolumn{1}{c|}{\multirow{4}{*}{}}   & 94.2 & 88.0 & 89.6 & 94.4 & 88.7 & 90.4 \\
GateVPT  &     & 94.3 & 88.5 & 89.7 & 95.2 & 89.6 & 91.1 \\
E2VPT    &     & 94.1 & 87.9 & 89.4 & 95.1 & 89.3 & 90.7 \\
Ours     & \multirow{-5.5}{*}{ViT-B-21k}   & 95.0 & 89.2 & 90.3 & 95.6 & 90.0 & 91.3 \\
\midrule
VPT      & \multicolumn{1}{c|}{\multirow{4}{*}{}}    & 95.6 & 90.2 & 91.6 & 96.1 & 91.0 & 92.3 \\
GateVPT  &      & 95.5 & 89.7 & 91.2 & 96.2 & 91.1 & 92.5 \\
E2VPT    &      & 95.4 & 89.9 & 91.4 & 96.1 & 91.1 & 92.4 \\
Ours     & \multirow{-5.5}{*}{DINOv2/B}    & 96.4 & 91.0 & 92.3 & 97.0 & 92.1 & 93.1 \\
\midrule
VPT      & \multicolumn{1}{c|}{\multirow{4}{*}{}}     & 92.8 & 86.4 & 88.5 & 93.8 & 88.1 & 90.1 \\
GateVPT  &      & 92.1 & 85.5 & 87.7 & 93.3 & 87.3 & 89.3 \\
E2VPT    &      & 89.6 & 83.1 & 86.3 & 91.2 & 84.8 & 88.0 \\
Ours     & \multirow{-5.5}{*}{DINOv2/S}    & 94.3 & 88.3 & 90.3 & 95.7 & 90.2 & 92.0 \\
\bottomrule
\end{tabular}}
\label{tab:voc}
\vspace{-0.5em}
\end{table}
\noindent \textbf{Performance on Pascal VOC 2007.}
For \texttt{VOC07}, we present the results of both our method and the comparison method in \cref{tab:voc}.
For clarity and brevity, unlike previous methods \cite{chen2019gcn,liu2021query2label,wang2025splicemix}, we report the same evaluation metric as those used in the \texttt{COCO} and do not provide the Average Precision (AP) for each class.
\Cref{tab:voc} illustrates that in the same settings, our method generally outperforms other approaches in mAP, OF1, and CF1 at both 224$\times$224 and 448$\times$448 resolutions. Specifically, at the 224$\times$224 resolution, our method shows improvements of 0.7\%-8.9\% in mAP, 0.7\%-8.6\% in CF1, and 0.6\%-7.2\% in OF1 compared to SOTA methods. Similarly, at the 448$\times$448 resolution, the enhancements are 0.4\%-7.6\% in mAP, 0.4\%-8.0\% in CF1, and 0.2\%-5.8\% in OF1 compared to SOTA methods.
These results robustly confirm the effectiveness of the proposed method.
\begin{table}[!t]
\renewcommand\arraystretch{0.50}
\centering
\caption{Comparison of our method with SOTA models on \texttt{NUS} and \texttt{VG256} at 224 $\times$ 224 resolution. All metrics are in \%.}
\vspace{-1.0em}
\resizebox{0.9\linewidth}{!}
{\begin{tabular}{c|c|c|c|c|c|c|c} 
\hline\thickhline
\multicolumn{1}{c|}{\multirow{2}{*}{\cellcolor{igray}}}
&
\multicolumn{1}{c|}{\multirow{2}{*}{\cellcolor{igray}}}
& 
\multicolumn{3}{c|}{\cellcolor{igray} NUS} & \multicolumn{3}{c}{\cellcolor{igray} VG256 } \\ 
\multirow{-2.5}{*}{\cellcolor{igray}Method} & \multirow{-2.5}{*}{\cellcolor{igray}Backbone} & \cellcolor{igray}mAP &  \cellcolor{igray}CF1 & \cellcolor{igray}OF1 & \cellcolor{igray}mAP & \cellcolor{igray}CF1 & \cellcolor{igray}OF1 \\
\hline
\midrule
VPT      & \multicolumn{1}{c|}{\multirow{4}{*}{}}        & 65.2 & 62.8 & 73.8 & 42.2 & 44.0 & 57.5  \\
GateVPT  &         & 62.8 & 61.5 & 72.9 & 41.1 & 43.0 & 56.5  \\
E2VPT    &         & 63.9 & 62.2 & 73.7 & 42.2 & 44.2 & 57.6  \\ 
Ours     & \multirow{-5.5}{*}{ViT-B}        & 65.7 & 63.3 & 74.1 & 44.2 & 45.8 & 58.8  \\
\midrule
VPT      & \multicolumn{1}{c|}{\multirow{4}{*}{}}          & 60.3 & 59.3 & 73.0 & 37.9 & 40.3 & 55.2  \\
GateVPT  &            & 58.2 & 57.6 & 71.9 & 35.3 & 38.2 & 53.2  \\
E2VPT    &            & 60.6 & 59.4 & 73.0 & 37.6 & 40.2 & 55.0  \\
Ours     & \multirow{-5.5}{*}{MAE}          & 61.9 & 60.7 & 73.5 & 40.5 & 42.7 & 57.3  \\
\midrule
VPT      & \multicolumn{1}{c|}{\multirow{4}{*}{}}       & 62.7 & 61.3 & 73.8 & 40.4 & 42.6 & 56.5  \\
GateVPT  &       & 61.8 & 60.5 & 72.9 & 39.6 & 41.8 & 55.8  \\
E2VPT    &       & 62.8 & 61.2 & 73.5 & 40.4 & 42.5 & 56.5  \\
Ours     & \multirow{-5.5}{*}{MoCo v3}      & 63.0 & 61.1 & 73.6 & 41.0 & 43.1 & 57.1  \\
\midrule
VPT      & \multicolumn{1}{c|}{\multirow{4}{*}{}}    & 67.5 & 64.8 & 74.7 & 46.0 & 47.4 & 60.2  \\
GateVPT  &     & 66.9 & 64.4 & 74.3 & 45.4 & 46.8 & 59.5  \\
E2VPT    &     & 67.7 & 64.9 & 74.7 & 46.0 & 47.4 & 60.1  \\
Ours     & \multirow{-5.5}{*}{ViT-B-21k}    & 68.2 & 65.0 & 75.0 & 47.6 & 48.6 & 61.2  \\
\midrule
VPT      & \multicolumn{1}{c|}{\multirow{4}{*}{}}     & 68.1 & 65.1 & 75.0 & 49.3 & 50.3 & 62.9  \\
GateVPT  &      & 67.2 & 64.6 & 74.5 & 48.6 & 49.7 & 62.4  \\
E2VPT    &      & 67.9 & 65.0 & 75.0 & 49.2 & 50.3 & 62.9  \\
Ours     & \multirow{-5.5}{*}{DINOv2/B}     & 68.7 & 65.5 & 75.2 & 50.9 & 51.7 & 64.1  \\
\midrule
VPT      & \multicolumn{1}{c|}{\multirow{4}{*}{}}     & 65.3 & 63.2 & 74.3 & 45.2 & 46.7 & 60.2  \\
GateVPT  &      & 64.5 & 62.6 & 73.7 & 44.3 & 45.9 & 59.3  \\
E2VPT    &      & 65.4 & 63.1 & 74.3 & 45.1 & 46.7 & 60.1  \\
Ours     & \multirow{-5.5}{*}{DINOv2/S}     & 66.9 & 64.3 & 74.8 & 47.7 & 48.7 & 61.9  \\
\bottomrule
\end{tabular}}
\label{tab:nusvg256}
\vspace{-1.5em}
\end{table}


\noindent \textbf{Performance on NUS-WIDE and Visual Genome.}
In \cref{tab:nusvg256}, we show the experimental results for \texttt{NUS} and \texttt{VG256}. For the \texttt{VG256} setup, we follow previous work \cite{xiecounterfactual}. Our method achieves the best performance and outperforms SOTA methods in mAP, CF1, and OF1, under the same pre-training model and resolution setting.
\subsection{Diagnostic Experiments}
\begin{figure}[t]
	\centering
	\begin{subfigure}{0.49\linewidth}
		\includegraphics[width=0.98\linewidth]{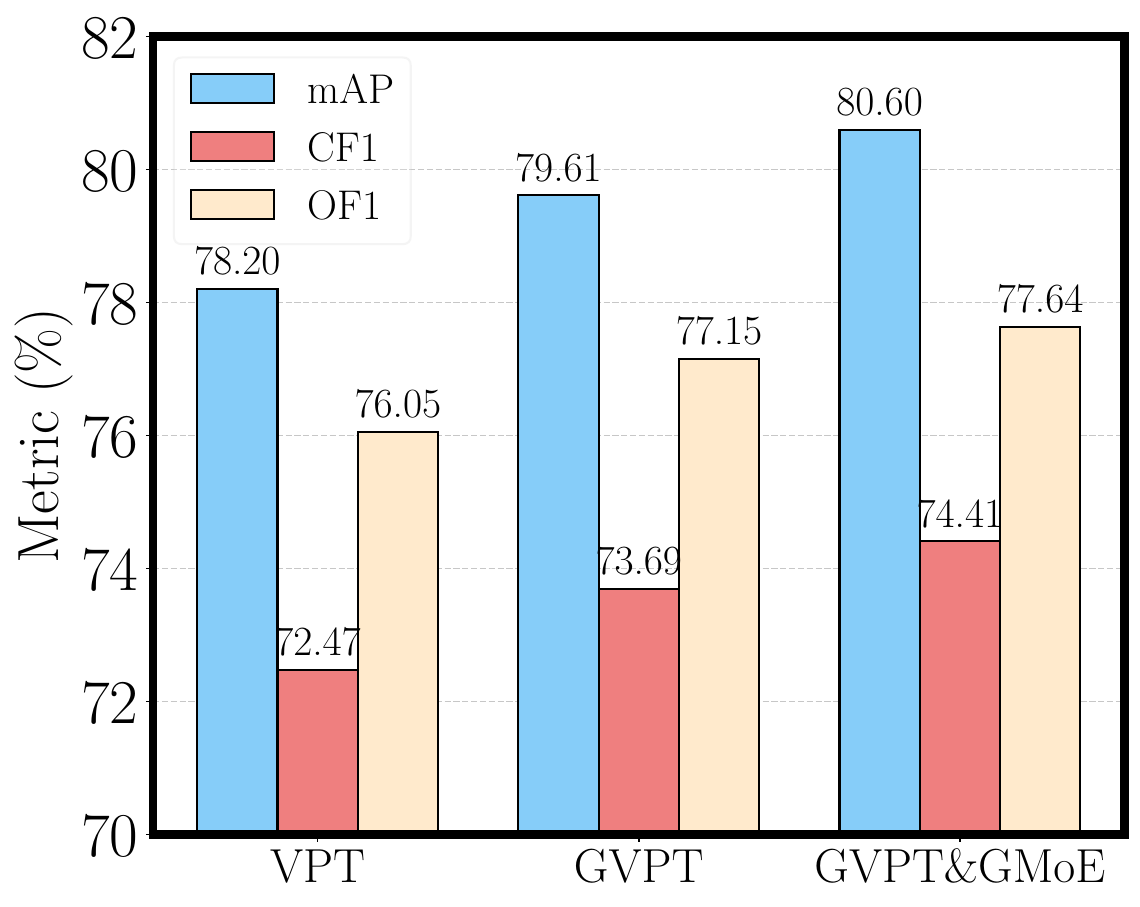}
            \vspace{-0.5em}
		\caption{evaluation on \texttt{COCO}}
		\label{fig:coco_expert_component}
	\end{subfigure}
	\hfill
	\begin{subfigure}{0.49\linewidth}
		\includegraphics[width=0.98\linewidth]{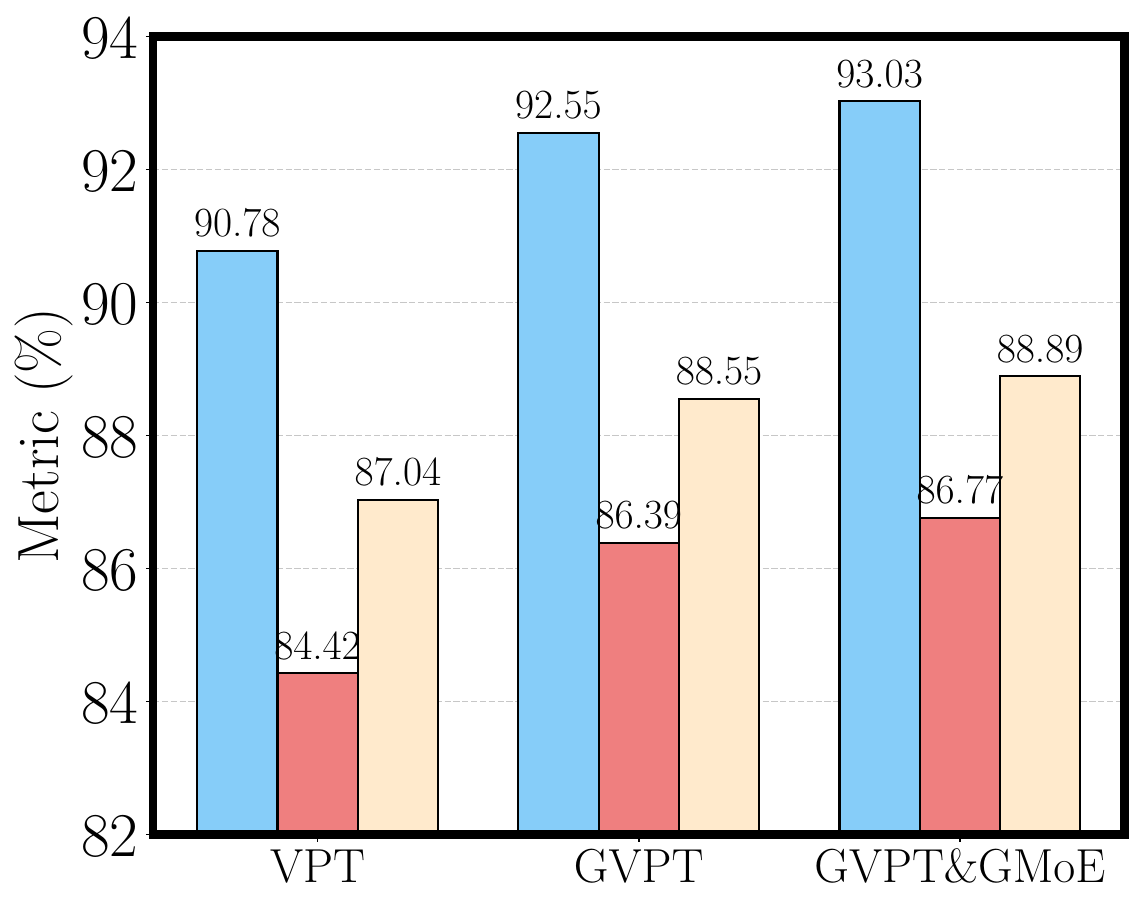}
            \vspace{-0.5em}
		\caption{evaluation on \texttt{VOC07}}
		\label{fig:voc_expert_component}
	\end{subfigure}
        \vspace{-1.0em}
    \caption{The analysis of different components. VPT stands for the vanilla model, GVPT\&GMoE combines both GVPT and GMoE.
    Note that each model uses 30 additional prompt tokens.}
    \label{fig:ablation_components}
    \vspace{-1.0em}
\end{figure}

\noindent \textbf{Impact of Different Components.}
To investigate the impact of different components of our method, including Group-Level Visual Prompt Tuning (GVPT) and Group-Level Mixture-of-Experts (GMoE), we conducted experiments on two benchmark datasets: \texttt{COCO} and \texttt{VOC07}.
The \cref{fig:ablation_components} illustrates the performance enhancements achieved by GVPT on the \texttt{COCO}, with improvements of 1.41\% in mAP, 1.22\% in CF1, and 1.1\% in OF1, respectively.
Moreover, incorporating GMoE into GVPT further improves performance by  0.99\%, 0.72\%, and 0.49\%, respectively, compared to GVPT alone.
On the \texttt{VOC07}, GVPT led to improvements of 1.77\%, 1.97\%, and 1.55\% in mAP, CF1, and OF1, respectively.  
However, GMoE has limited improvement in model performance, \eg, 0.48\% in mAP.
These results reveal that the grouping classes in \texttt{VOC07} have a relatively weak ability to learn label correlations within each group, as each group contains relatively few classes, and the distinctions between multiple tokens within each group may be minimal.

\begin{figure}[t]
	\centering
	\begin{subfigure}{0.49\linewidth}
		\includegraphics[width=0.98\linewidth]{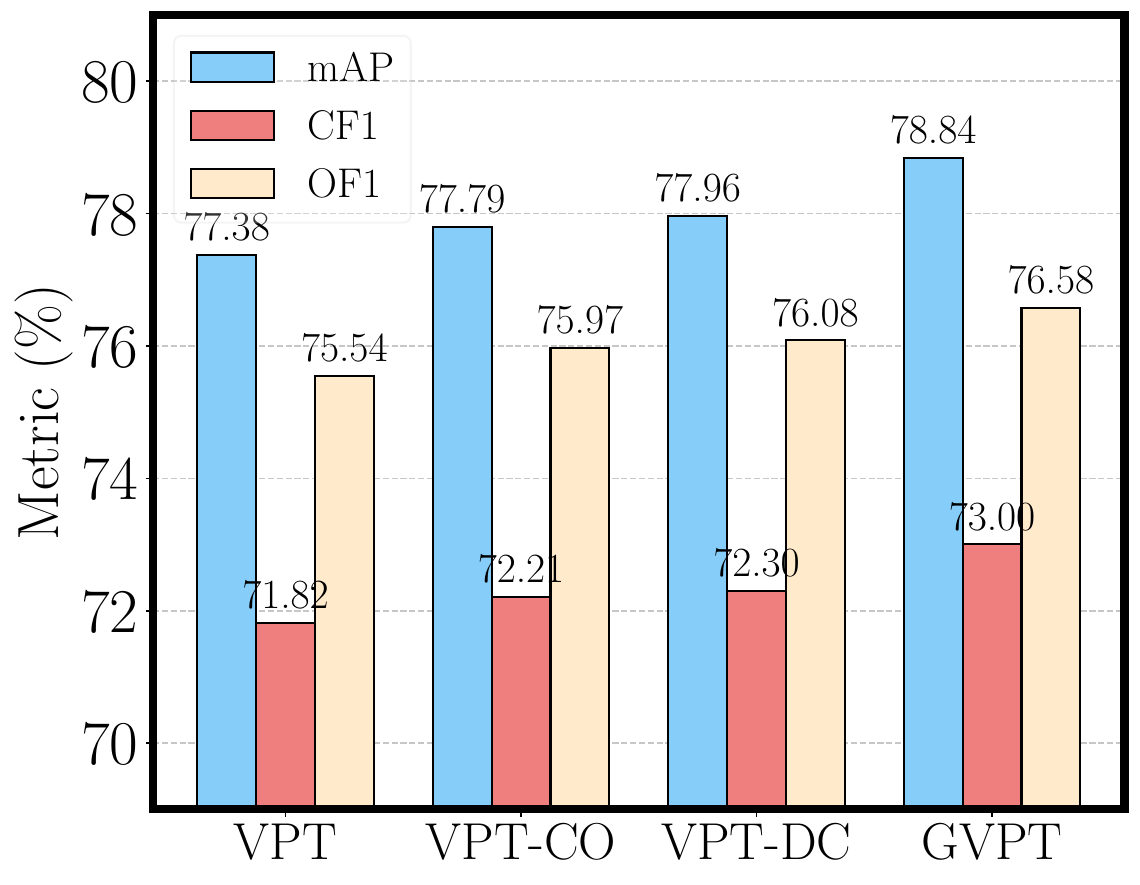}
            \vspace{-0.5em}
		\caption{evaluation on \texttt{COCO}}
		\label{fig:coco_group_component}
	\end{subfigure}
	\hfill
	\begin{subfigure}{0.49\linewidth}
		\includegraphics[width=0.98\linewidth]{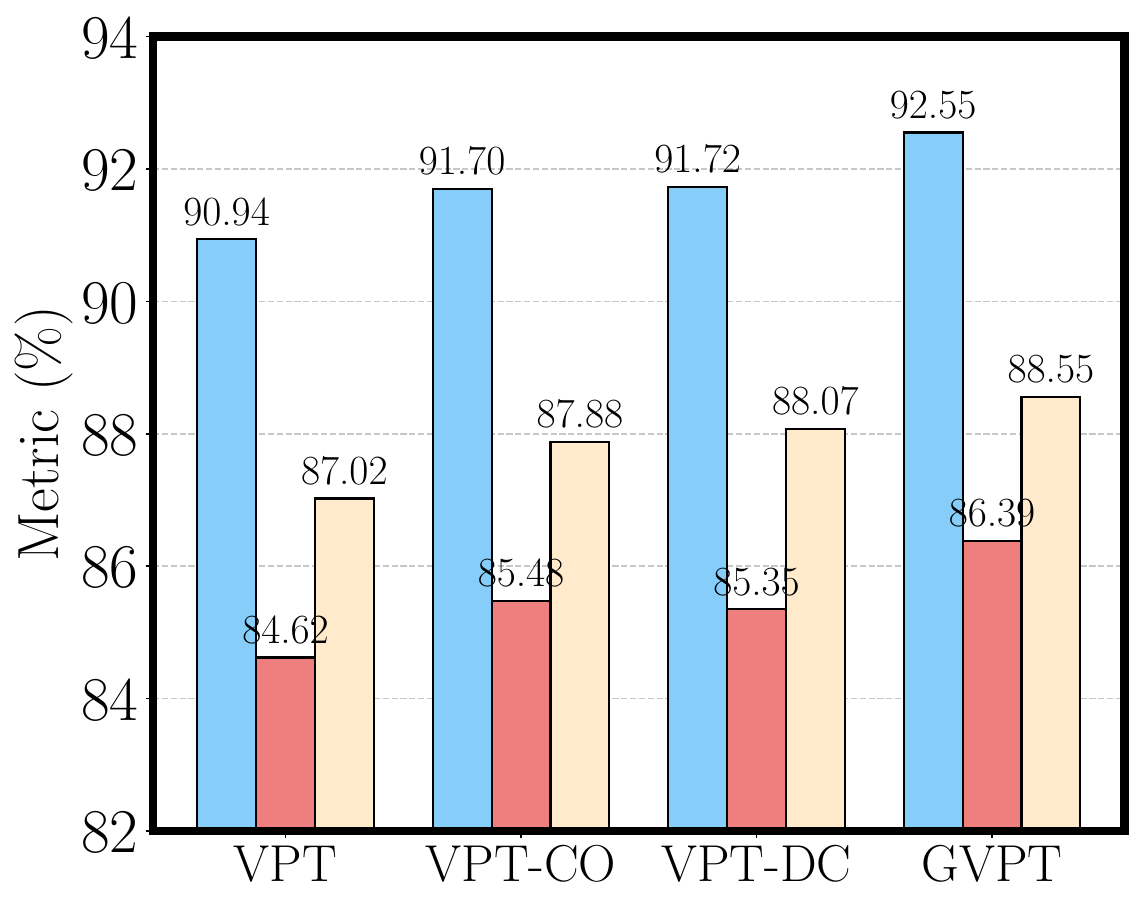}
            \vspace{-0.5em}
		\caption{evaluation on \texttt{VOC07}}
		\label{fig:voc_group_component}
	\end{subfigure}
        \vspace{-1.0em}
	\caption{The analysis of different grouping strategies. VPT is vanilla mode; VPT-CO indicates the VPT uses the CO groups. VPT-DC applies DC groups in VPT. GVPT incorporates both. Note that each model uses 10 additional prompt tokens and excludes MoE.}
    \label{fig:ablation_grouping}
    \vspace{-1.0em}
\end{figure}

\noindent \textbf{Grouping Strategies.}
To study the impact of the grouping strategy, we use four settings: vanilla VPT, using only the CO group, using only the DC group, and using both. As can be seen from \cref{fig:ablation_grouping}, on \texttt{COCO} and \texttt{VOC07}, using the CO group and the DC group, respectively, has a certain improvement. When using both modes, the model is improved even more. We analyze that emphasizing correlative and discriminative relationships among labels has a certain gain on the MLC. While emphasizing the co-occurrence relationship, cleverly using the discriminative relationship will further compete for the model's performance.

\begin{figure}[t]
	\centering
	\begin{subfigure}{0.49\linewidth}
		\includegraphics[width=0.98\linewidth]{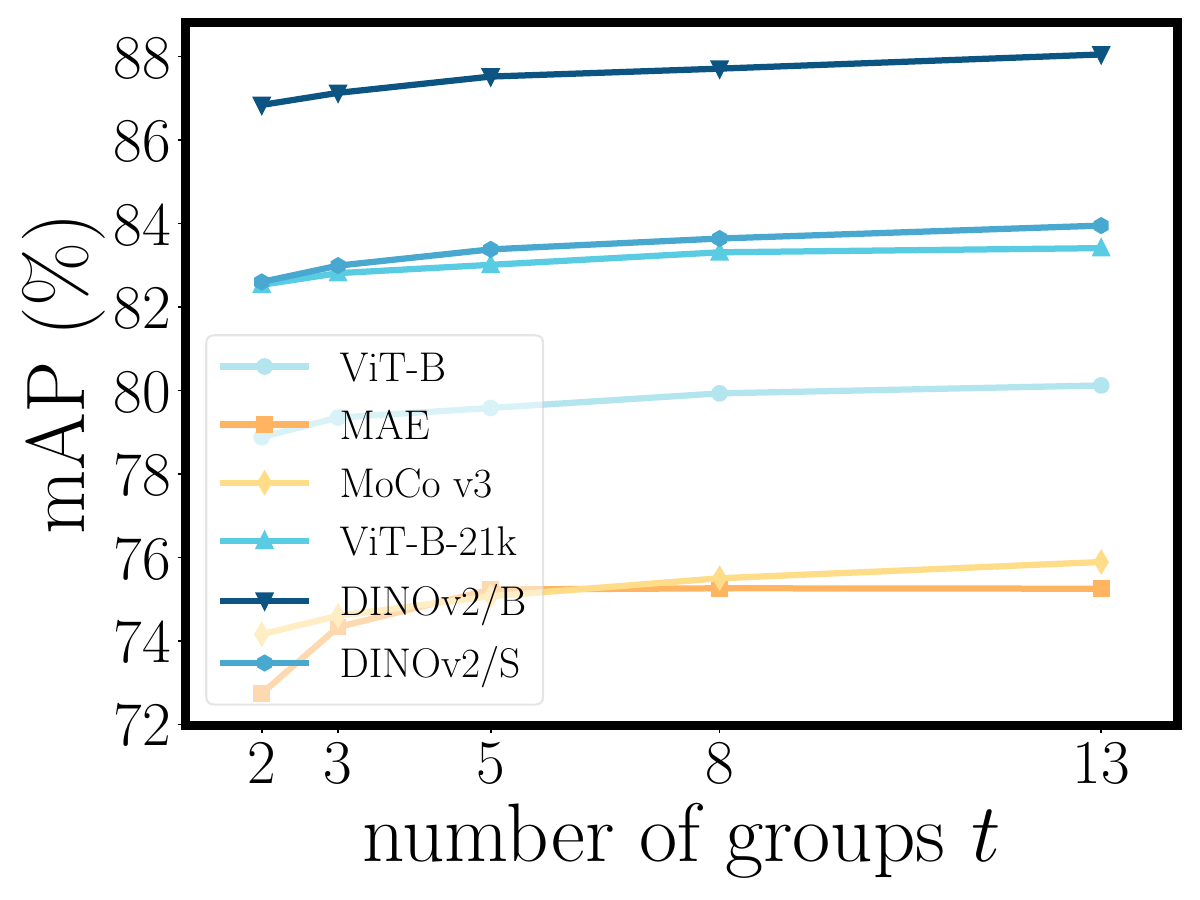}
            \vspace{-0.5em}
		\caption{mAP on \texttt{COCO}}
		\label{fig:coco_map_group}
	\end{subfigure}
	\hfill
	\begin{subfigure}{0.49\linewidth}
		\includegraphics[width=0.98\linewidth]{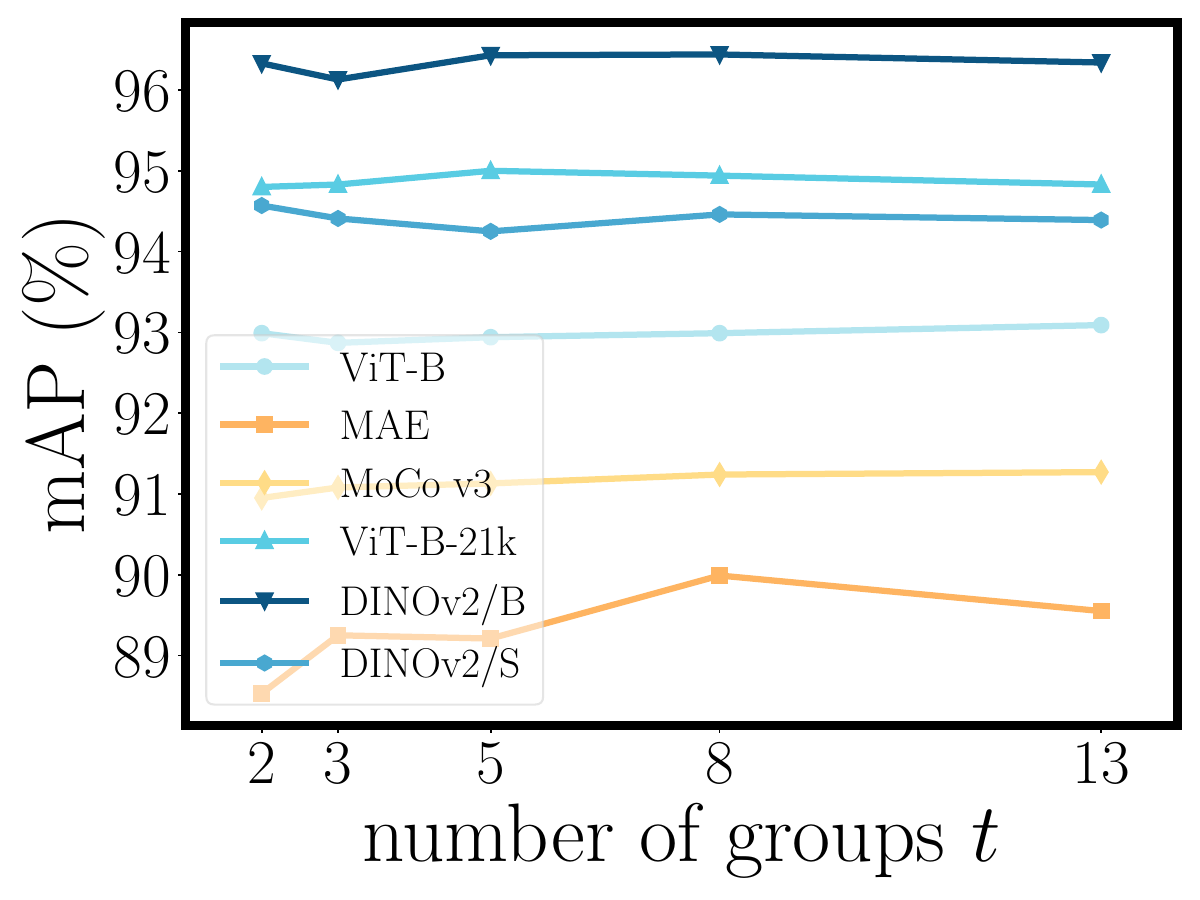}
            \vspace{-0.5em}
		\caption{mAP on \texttt{VOC07}}
		\label{fig:vco_map_group}
	\end{subfigure}
        \vspace{-1.0em}
	\caption{The performance curve varies with the increase in the number of groups. More can be found in Appendix.}
    \label{fig:ablation_number_group}
    \vspace{-1.0em}
\end{figure}

\noindent \textbf{Number of Groups.}
In this study, our main idea is to group classes within the co-occurrence graph. 
We then aim to establish both a CO and DC relationship.
Hence, additional research is required to explore the impact of group quantity on model performance.
As illustrated in \cref{fig:ablation_number_group}, we present comprehensive ablation studies on the \texttt{COCO} and \texttt{VOC07}.
For the \texttt{COCO}, model performance improves with the increased number of groups when the same pre-trained model is used. 
Notably, when pre-training with MoCo v3, optimal performance is achieved with 5 groups.
For \texttt{VOC07}, optimal results are achieved when the number of groups is set to either 5 or 8. 
The potential reason is the dataset's relatively small number of 20 classes, which likely impedes the effective learning of label correlations in the CO group.  
Additionally, the average number of labels per image in \texttt{VOC07} is only 1.5, leading to weaker correlations between labels in many images compared to those in \texttt{COCO}.

\begin{figure}[t]
	\centering
	\begin{subfigure}{0.49\linewidth}
		\includegraphics[width=0.98\linewidth]{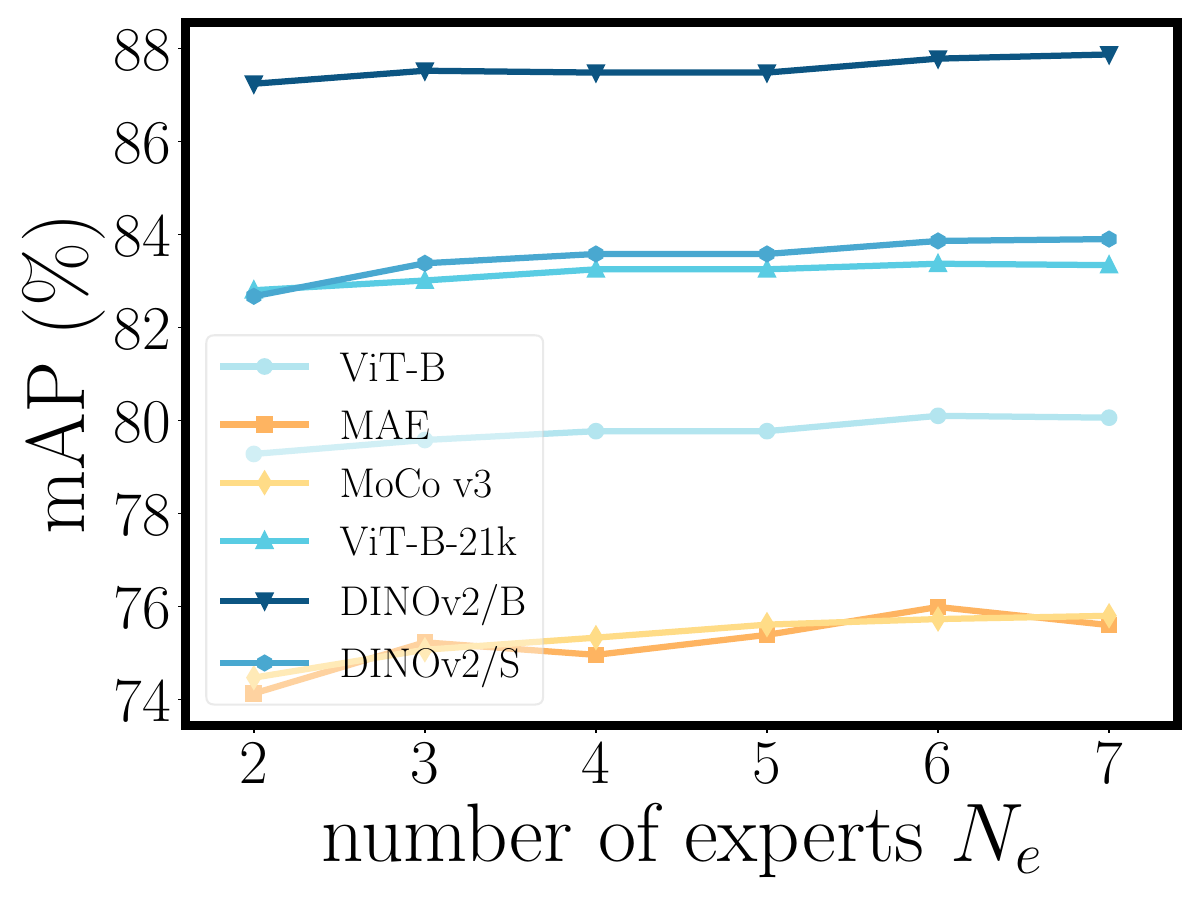}
            \vspace{-0.5em}
		\caption{mAP on \texttt{COCO}}
		\label{fig:coco_map_expert}
	\end{subfigure}
	\hfill
	\begin{subfigure}{0.49\linewidth}
		\includegraphics[width=0.98\linewidth]{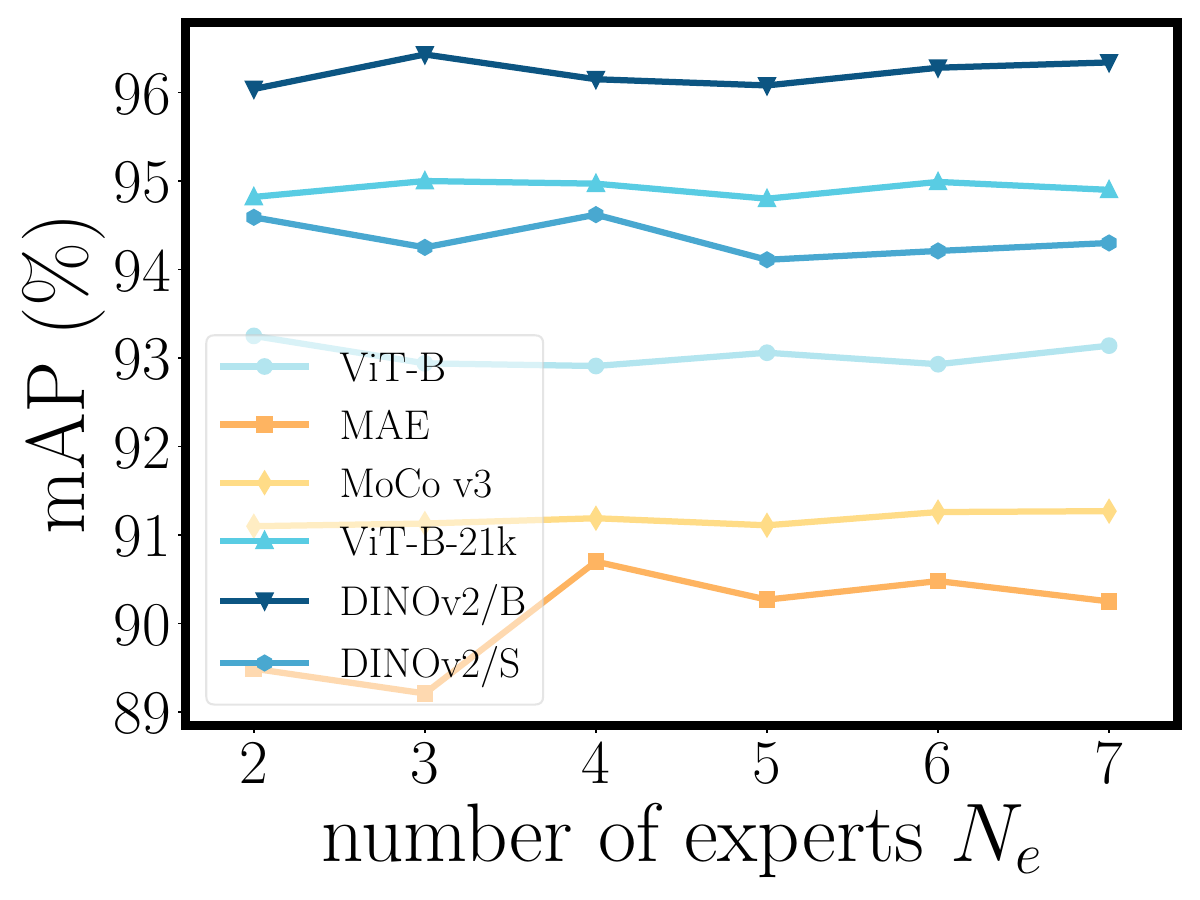}
            \vspace{-0.5em}
		\caption{mAP on \texttt{VOC07}}
		\label{fig:voc_map_expert}
	\end{subfigure}
        \vspace{-1.0em}
	\caption{The performance curve varies with the increase in the number of experts, in the mAP, More can be found in Appendix.}
    \label{fig:ablation_number_expert}
    \vspace{-1.0em}
\end{figure}

\noindent \textbf{Number of Experts.}
The number of experts corresponds to the number of group-level representatives within each class group, which reflects the capacity to model label relationships within each group.
Accordingly, we study the influence of the number of experts on model performance. 
As shown in \cref{fig:ablation_number_expert}, on the \texttt{COCO}, model performance marginally improves as the number of experts increases. However, the performance on \texttt{VOC07} shows slight fluctuations. This phenomenon may be attributed to the small number of classes in each group, which does not require more prompt tokens to learn the relationship between labels. Therefore, adding more experts yields limited benefits and could potentially detract from model performance.
\begin{table}[h]
\centering{
\vspace{-1.0em}
\caption{Computation cost and parameters number on \texttt{COCO}.}
\vspace{-1.0em}
\renewcommand\arraystretch{0.50}
\resizebox{0.95\linewidth}{!}{
    \begin{tabular}{c|c|c|c|c}
    \toprule
    Method             & \# Total param.  & \# Learnable param.  & FLOPs    &  mAP \\
    \midrule
    Linear Probing     & 86.64 M          & 0.06 M               & 21.96 G  &  77.8\% \\
    Fine Tuning        & 86.64 M          & 86.64 M              & 21.96 G  &  84.2\% \\
    VPT                & 86.91 M          & 0.33 M               & 24.51 G  &  86.1\% \\
    E2VPT              & 87.01 M          & 0.43 M               & 24.55 G  &  86.3\% \\ 
    GateVPT            & 86.67 M          & 0.08 M               & 24.51 G  &  85.6\% \\
    GVPT (Ours)        & 86.98 M          & 0.39 M               & 24.51 G  &  87.2\% \\
    ML-VPT (Ours)      & 87.60 M          & 1.02 M               & 24.51 G  &  87.5\% \\ 
    \bottomrule                      	 
    \end{tabular}
}
\vspace{-1em}
\label{tab:parameters_table1}
}
\end{table}

\begin{figure}[t]
    \centering
    \includegraphics[width=0.99\linewidth]{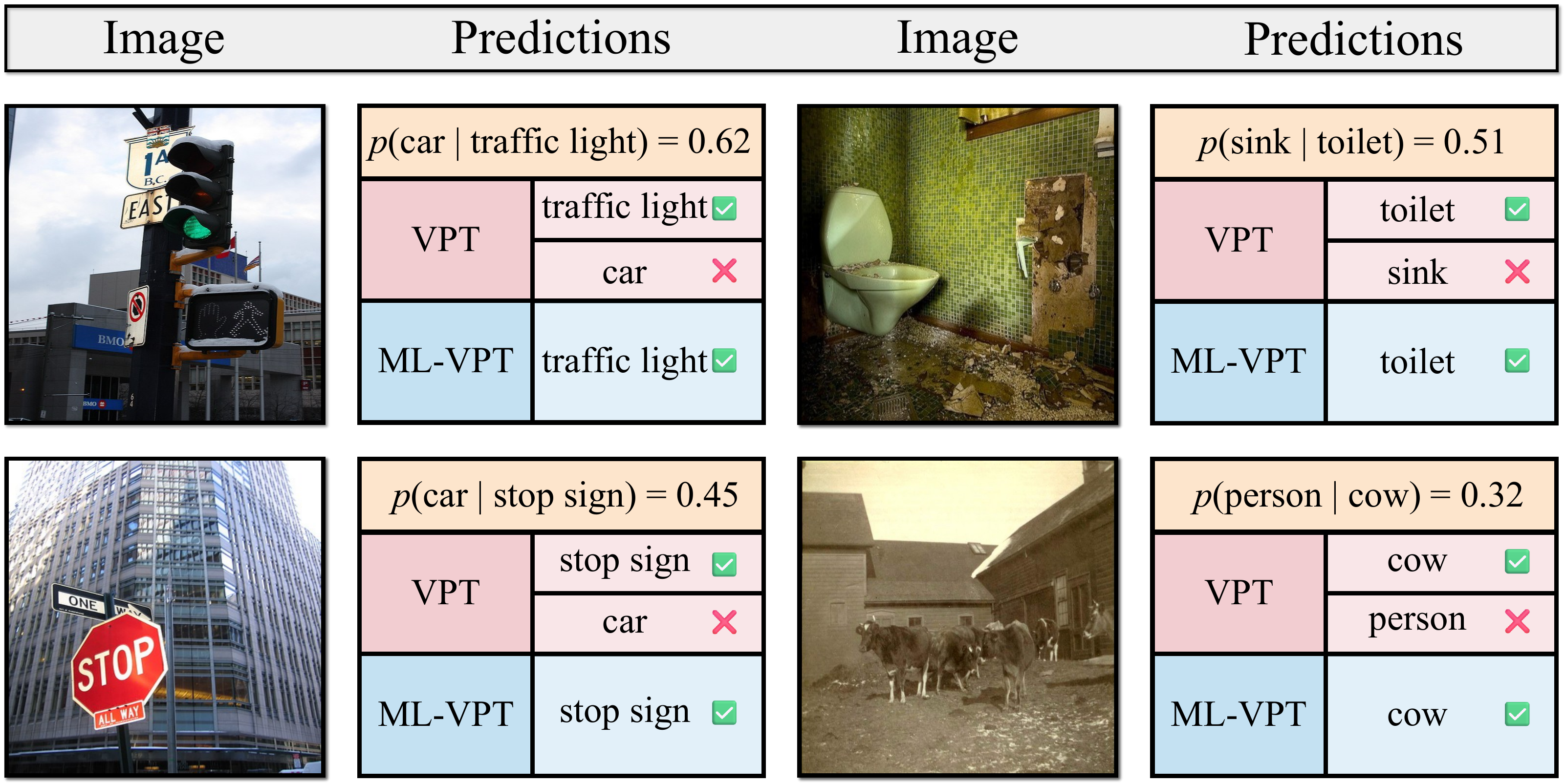}
    \vspace{-0.5em}
    \caption{Visualization of model predictions on \texttt{COCO}. Each example includes an input image, the co-occurrence probability, and the predictions of the two models. 
}
\label{fig:casestudy_mainbody}
\vspace{-1.0em}
\end{figure}

\noindent \textbf{Computation Cost and Parameters.}
To verify the first issue mentioned in \S \ref{sec:intro}, we use DINOv2/B as a pre-trained model in several methods to compare the computational overhead and parameters. 
As shown in \cref{tab:parameters_table1}, FineTuning not only requires more learnable parameters but also reduces model performance. Our method achieves the highest result with limited computational resources and parameters.

\subsection{Case Study}
To verify that our method can alleviate the co-occurrence overfitting problem, we select label pairs with high co-occurrence probabilities to study. As shown in \cref{fig:casestudy_mainbody}, ML-VPT is compared with VPT on \texttt{COCO}. Under the influence of high co-occurrence probabilities, VPT is prone to over-inference, resulting in model misprediction. For example, in the second example, when {\ttfamily\text{toilet}} appears, VPT infers that there is a {\ttfamily\text{sink}} through co-occurrence classes and other contextual information (the {\ttfamily\text{sink}} has been removed). However, our method can correctly predict all labels.
Similarly, in outdoor scenes, {\ttfamily\text{traffic light}} and {\ttfamily\text{stop sign}} are both misclassified as {\ttfamily\text{car}}, while in animal-related classes, the presence of a {\ttfamily\text{cow}} leads to the false inference of a {\ttfamily\text{person}}. These examples empirically demonstrate that ML-VPT can mitigate co-occurrence overfitting.

\section{Conclusion} \label{sec:conclusion}
In MLC, mainstream methods tend to learn co-occurrence relationships between labels. However, an overemphasis on these relationships can lead to model overfitting. To address this issue, we introduce a novel Multi-Label Visual Prompt Tuning method designed to balance both correlative and discriminative relationships among labels. 
Our method offers two advantages: i) We employ the concept of grouping classes to independently model the correlative and discriminative relationships between labels, thus reducing the risk of overfitting. ii) While the grouping idea allows multiple classes to share the same group information, we further enhance our approach by integrating a group-aware MoE model that dynamically maps optimal multiple group-aware representations to a label-aware representation for each class.
\section*{Acknowledgement}
This work was supported in part by the National Natural Science Foundation of China (No.62076005, No.62472004, No.61876002), the Provincial Quality Project of Education in the New Era in 2023 (Postgraduate Education No.2023lhpysfjd009) and the University Synergy Innovation Program of Anhui Province, China (GXXT-2021-002, GXXT-2022-029, GXXT-2022-031).
We thank the reviewers for their comments and suggestions.
We also thank the High-performance Computing Platform of Anhui University for providing computational resources for this project.

{
    \small
    \bibliographystyle{ieeenat_fullname}
    \bibliography{main}

\begin{thebibliography}{61}
\providecommand{\natexlab}[1]{#1}
\providecommand{\url}[1]{\texttt{#1}}
\expandafter\ifx\csname urlstyle\endcsname\relax
  \providecommand{\doi}[1]{doi: #1}\else
  \providecommand{\doi}{doi: \begingroup \urlstyle{rm}\Url}\fi

\bibitem[Ben-Baruch et~al.(2022)Ben-Baruch, Ridnik, Friedman, Ben-Cohen, Zamir, Noy, and Zelnik-Manor]{ben2022multi}
Emanuel Ben-Baruch, Tal Ridnik, Itamar Friedman, Avi Ben-Cohen, Nadav Zamir, Asaf Noy, and Lihi Zelnik-Manor.
\newblock Multi-label classification with partial annotations using class-aware selective loss.
\newblock In \emph{CVPR}, pages 4764--4772, 2022.

\bibitem[Chang et~al.(2020)Chang, Wang, Hung, Piramuthu, Tsai, and Yang]{chang2020weakly}
Yu-Ting Chang, Qiaosong Wang, Wei-Chih Hung, Robinson Piramuthu, Yi-Hsuan Tsai, and Ming-Hsuan Yang.
\newblock Weakly-supervised semantic segmentation via sub-category exploration.
\newblock In \emph{CVPR}, pages 8991--9000, 2020.

\bibitem[Chen et~al.(2020)Chen, Wang, Chen, Shi, Geng, and Rui]{chen2020label}
Shikai Chen, Jianfeng Wang, Yuedong Chen, Zhongchao Shi, Xin Geng, and Yong Rui.
\newblock Label distribution learning on auxiliary label space graphs for facial expression recognition.
\newblock In \emph{CVPR}, pages 13984--13993, 2020.

\bibitem[Chen et~al.(2024)Chen, Jie, and Ma]{chen2024llava}
Shaoxiang Chen, Zequn Jie, and Lin Ma.
\newblock Llava-mole: Sparse mixture of lora experts for mitigating data conflicts in instruction finetuning mllms.
\newblock \emph{arXiv preprint arXiv:2401.16160}, 2024.

\bibitem[Chen et~al.(2019{\natexlab{a}})Chen, Xu, Hui, Wu, and Lin]{chen2019iccv}
Tianshui Chen, Muxin Xu, Xiaolu Hui, Hefeng Wu, and Liang Lin.
\newblock Learning semantic-specific graph representation for multi-label image recognition.
\newblock In \emph{ICCV}, pages 522--531, 2019{\natexlab{a}}.

\bibitem[Chen et~al.(2021)Chen, Xie, and He]{chen2021empirical}
Xinlei Chen, Saining Xie, and Kaiming He.
\newblock An empirical study of training self-supervised vision transformers.
\newblock In \emph{ICCV}, 2021.

\bibitem[Chen et~al.(2019{\natexlab{b}})Chen, Wei, Wang, and Guo]{chen2019gcn}
Zhao-Min Chen, Xiu-Shen Wei, Peng Wang, and Yanwen Guo.
\newblock Multi-label image recognition with graph convolutional networks.
\newblock In \emph{CVPR}, pages 5177--5186, 2019{\natexlab{b}}.

\bibitem[Chua et~al.(2009)Chua, Tang, Hong, Li, Luo, and Zheng]{chua2009nus}
Tat-Seng Chua, Jinhui Tang, Richang Hong, Haojie Li, Zhiping Luo, and Yan-Tao Zheng.
\newblock Nus-wide: A real-world web image database from national university of singapore.
\newblock In \emph{CIVR}, Santorini, Greece., 2009.

\bibitem[Cubuk et~al.(2020)Cubuk, Zoph, Shlens, and Le]{cubuk2020randaugment}
Ekin~D Cubuk, Barret Zoph, Jonathon Shlens, and Quoc~V Le.
\newblock Randaugment: Practical automated data augmentation with a reduced search space.
\newblock In \emph{CVPRW}, pages 702--703, 2020.

\bibitem[DeVries and Taylor(2017)]{devries2017improved}
Terrance DeVries and Graham~W Taylor.
\newblock Improved regularization of convolutional neural networks with cutout.
\newblock \emph{arXiv preprint arXiv:1708.04552}, 2017.

\bibitem[Ding et~al.(2023)Ding, Qin, Yang, Wei, Yang, Su, Hu, Chen, Chan, Chen, et~al.]{ding2023parameter}
Ning Ding, Yujia Qin, Guang Yang, Fuchao Wei, Zonghan Yang, Yusheng Su, Shengding Hu, Yulin Chen, Chi-Min Chan, Weize Chen, et~al.
\newblock Parameter-efficient fine-tuning of large-scale pre-trained language models.
\newblock \emph{Nature Machine Intelligence}, 5\penalty0 (3):\penalty0 220--235, 2023.

\bibitem[Dosovitskiy(2020)]{dosovitskiy2020image}
Alexey Dosovitskiy.
\newblock An image is worth 16x16 words: Transformers for image recognition at scale.
\newblock \emph{arXiv preprint arXiv:2010.11929}, 2020.

\bibitem[Everingham et~al.(2015)Everingham, Eslami, Van~Gool, Williams, Winn, and Zisserman]{everingham2010pascal}
Mark Everingham, SM~Ali Eslami, Luc Van~Gool, Christopher~KI Williams, John Winn, and Andrew Zisserman.
\newblock The pascal visual object classes challenge: A retrospective.
\newblock \emph{IJCV}, 111:\penalty0 98--136, 2015.

\bibitem[Fu et~al.(2023)Fu, Yang, So, Lam, Bing, and Collier]{fu2023effectiveness}
Zihao Fu, Haoran Yang, Anthony Man-Cho So, Wai Lam, Lidong Bing, and Nigel Collier.
\newblock On the effectiveness of parameter-efficient fine-tuning.
\newblock In \emph{Proceedings of the AAAI conference on artificial intelligence}, pages 12799--12807, 2023.

\bibitem[Gao et~al.(2023)Gao, Zhao, Sun, Xi, Zhang, Ghanem, and Zhang]{gao2023unified}
Qiankun Gao, Chen Zhao, Yifan Sun, Teng Xi, Gang Zhang, Bernard Ghanem, and Jian Zhang.
\newblock A unified continual learning framework with general parameter-efficient tuning.
\newblock In \emph{ICCV}, pages 11483--11493, 2023.

\bibitem[Gao et~al.(2022)Gao, Shi, Zhu, Wang, Tang, Zhou, Li, and Metaxas]{gao2022visual}
Yunhe Gao, Xingjian Shi, Yi Zhu, Hao Wang, Zhiqiang Tang, Xiong Zhou, Mu Li, and Dimitris~N Metaxas.
\newblock Visual prompt tuning for test-time domain adaptation.
\newblock \emph{arXiv preprint arXiv:2210.04831}, 2022.

\bibitem[Han et~al.(2023)Han, Wang, Cui, Cao, Wang, Qi, and Liu]{han20232vpt}
Cheng Han, Qifan Wang, Yiming Cui, Zhiwen Cao, Wenguan Wang, Siyuan Qi, and Dongfang Liu.
\newblock E\^{} 2vpt: An effective and efficient approach for visual prompt tuning.
\newblock In \emph{ICCV}, pages 17445--17456, 2023.

\bibitem[Han et~al.(2024)Han, Wang, Cui, Wang, Huang, Qi, and Liu]{hanfacing}
Cheng Han, Qifan Wang, Yiming Cui, Wenguan Wang, Lifu Huang, Siyuan Qi, and Dongfang Liu.
\newblock Facing the elephant in the room: Visual prompt tuning or full finetuning?
\newblock In \emph{ICLR}, 2024.

\bibitem[He et~al.(2022)He, Chen, Xie, Li, Doll{\'a}r, and Girshick]{he2022masked}
Kaiming He, Xinlei Chen, Saining Xie, Yanghao Li, Piotr Doll{\'a}r, and Ross Girshick.
\newblock Masked autoencoders are scalable vision learners.
\newblock In \emph{CVPR}, 2022.

\bibitem[Hong et~al.(2024)Hong, Yan, Zhang, Li, Zhou, Guo, Jiang, Chen, Li, Chen, et~al.]{hong2024onetracker}
Lingyi Hong, Shilin Yan, Renrui Zhang, Wanyun Li, Xinyu Zhou, Pinxue Guo, Kaixun Jiang, Yiting Chen, Jinglun Li, Zhaoyu Chen, et~al.
\newblock Onetracker: Unifying visual object tracking with foundation models and efficient tuning.
\newblock In \emph{CVPR}, pages 19079--19091, 2024.

\bibitem[Jacobs et~al.(1991)Jacobs, Jordan, Nowlan, and Hinton]{jacobs1991adaptive}
Robert~A Jacobs, Michael~I Jordan, Steven~J Nowlan, and Geoffrey~E Hinton.
\newblock Adaptive mixtures of local experts.
\newblock \emph{Neural computation}, 3\penalty0 (1):\penalty0 79--87, 1991.

\bibitem[Jia et~al.(2022)Jia, Tang, Chen, Cardie, Belongie, Hariharan, and Lim]{jia2022visual}
Menglin Jia, Luming Tang, Bor-Chun Chen, Claire Cardie, Serge Belongie, Bharath Hariharan, and Ser-Nam Lim.
\newblock Visual prompt tuning.
\newblock In \emph{ECCV}, pages 709--727, 2022.

\bibitem[Kim et~al.(2024)Kim, Yu, and Hwang]{kim2024eclipse}
Beomyoung Kim, Joonsang Yu, and Sung~Ju Hwang.
\newblock Eclipse: Efficient continual learning in panoptic segmentation with visual prompt tuning.
\newblock In \emph{CVPR}, pages 3346--3356, 2024.

\bibitem[Krishna et~al.(2017)Krishna, Zhu, Groth, Johnson, Hata, Kravitz, Chen, Kalantidis, Li, Shamma, et~al.]{krishna2017visual}
Ranjay Krishna, Yuke Zhu, Oliver Groth, Justin Johnson, Kenji Hata, Joshua Kravitz, Stephanie Chen, Yannis Kalantidis, Li-Jia Li, David~A Shamma, et~al.
\newblock Visual genome: Connecting language and vision using crowdsourced dense image annotations.
\newblock \emph{IJCV}, 123:\penalty0 32--73, 2017.

\bibitem[Li et~al.(2023)Li, Shen, Yang, Wang, Ren, Che, Zhang, and Liu]{li2022sparse}
Bo Li, Yifei Shen, Jingkang Yang, Yezhen Wang, Jiawei Ren, Tong Che, Jun Zhang, and Ziwei Liu.
\newblock Sparse mixture-of-experts are domain generalizable learners.
\newblock In \emph{ICML}, 2023.

\bibitem[Li et~al.(2022)Li, Xia, Zhang, Zhan, Ge, and Liu]{li2022estimating}
Shikun Li, Xiaobo Xia, Hansong Zhang, Yibing Zhan, Shiming Ge, and Tongliang Liu.
\newblock Estimating noise transition matrix with label correlations for noisy multi-label learning.
\newblock \emph{NeurIPS}, 35:\penalty0 24184--24198, 2022.

\bibitem[Lin et~al.(2014)Lin, Maire, Belongie, Hays, Perona, Ramanan, Dollar, and Zitnick]{lin2014microsoft}
Tsung-Yi Lin, Michael Maire, Serge Belongie, James Hays, Pietro Perona, Deva Ramanan, Piotr Dollar, and Larry Zitnick.
\newblock Microsoft coco: Common objects in context.
\newblock In \emph{ECCV}, 2014.

\bibitem[Liu et~al.(2024)Liu, Ye, and Du]{LYD2024}
Fangyi Liu, Mang Ye, and Bo Du.
\newblock Learning a generalizable re-identification model from unlabelled data with domain-agnostic expert.
\newblock \emph{Visual Intelligence}, 2\penalty0 (1):\penalty0 28, 2024.

\bibitem[Liu et~al.(2022)Liu, Liu, Li, Hou, Yu, and Yang]{liu2022contextual}
Ruyang Liu, Hao Liu, Ge Li, Haodi Hou, TingHao Yu, and Tao Yang.
\newblock Contextual debiasing for visual recognition with causal mechanisms.
\newblock In \emph{CVPR}, pages 12755--12765, 2022.

\bibitem[Liu et~al.(2023)Liu, Huang, Li, and Li]{liu2023causality}
Ruyang Liu, Jingjia Huang, Thomas~H Li, and Ge Li.
\newblock Causality compensated attention for contextual biased visual recognition.
\newblock In \emph{ICLR}, 2023.

\bibitem[Liu et~al.(2021{\natexlab{a}})Liu, Zhang, Yang, Su, and Zhu]{liu2021query2label}
Shilong Liu, Lei Zhang, Xiao Yang, Hang Su, and Jun Zhu.
\newblock Query2label: A simple transformer way to multi-label classification.
\newblock \emph{arXiv preprint arXiv:2107.10834}, 2021{\natexlab{a}}.

\bibitem[Liu et~al.(2021{\natexlab{b}})Liu, Wang, Shen, and Tsang]{liu2021emerging}
Weiwei Liu, Haobo Wang, Xiaobo Shen, and Ivor~W Tsang.
\newblock The emerging trends of multi-label learning.
\newblock \emph{IEEE TPAMI}, 44\penalty0 (11):\penalty0 7955--7974, 2021{\natexlab{b}}.

\bibitem[Loshchilov and Hutter(2018)]{loshchilov2018decoupled}
Ilya Loshchilov and Frank Hutter.
\newblock Decoupled weight decay regularization.
\newblock In \emph{ICLR}, 2018.

\bibitem[Ma et~al.(2023)Ma, Sun, Wang, Zhao, and Luo]{ma2023semantic}
Leilei Ma, Dengdi Sun, Lei Wang, Haifeng Zhao, and Bin Luo.
\newblock Semantic-aware dual contrastive learning for multi-label image classification.
\newblock In \emph{ECAI}, pages 1656--1663, 2023.

\bibitem[Ma et~al.(2024)Ma, Xie, Wang, Fu, Sun, and Zhao]{ma2024text}
Leilei Ma, Hongxing Xie, Lei Wang, Yanping Fu, Dengdi Sun, and Haifeng Zhao.
\newblock Text-region matching for multi-label image recognition with missing labels.
\newblock In \emph{ACM MM}, 2024.

\bibitem[Oquab et~al.(2023)Oquab, Darcet, Moutakanni, Vo, Szafraniec, Khalidov, Fernandez, Haziza, Massa, El-Nouby, et~al.]{dinov22023oquab}
Maxime Oquab, Timoth{\'e}e Darcet, Th{\'e}o Moutakanni, Huy Vo, Marc Szafraniec, Vasil Khalidov, Pierre Fernandez, Daniel Haziza, Francisco Massa, Alaaeldin El-Nouby, et~al.
\newblock Dinov2: Learning robust visual features without supervision.
\newblock \emph{arXiv preprint arXiv:2304.07193}, 2023.

\bibitem[Park and Byun(2024)]{park2024fair}
Sungho Park and Hyeran Byun.
\newblock Fair-vpt: Fair visual prompt tuning for image classification.
\newblock In \emph{CVPR}, pages 12268--12278, 2024.

\bibitem[Ridnik et~al.(2021)Ridnik, Ben-Baruch, Zamir, Noy, Friedman, Protter, and Zelnik-Manor]{ridnik2021asymmetric}
Tal Ridnik, Emanuel Ben-Baruch, Nadav Zamir, Asaf Noy, Itamar Friedman, Matan Protter, and Lihi Zelnik-Manor.
\newblock Asymmetric loss for multi-label classification.
\newblock In \emph{CVPR}, pages 82--91, 2021.

\bibitem[Ridnik et~al.(2023)Ridnik, Sharir, Ben-Cohen, Ben-Baruch, and Noy]{mldecoder2023ridnik}
Tal Ridnik, Gilad Sharir, Avi Ben-Cohen, Emanuel Ben-Baruch, and Asaf Noy.
\newblock Ml-decoder: Scalable and versatile classification head.
\newblock In \emph{WACV}, pages 32--41, 2023.

\bibitem[Russakovsky et~al.(2015)Russakovsky, Deng, Su, Krause, Satheesh, Ma, Huang, Karpathy, Khosla, Bernstein, et~al.]{russakovsky2015imagenet}
Olga Russakovsky, Jia Deng, Hao Su, Jonathan Krause, Sanjeev Satheesh, Sean Ma, Zhiheng Huang, Andrej Karpathy, Aditya Khosla, Michael Bernstein, et~al.
\newblock Imagenet large scale visual recognition challenge.
\newblock \emph{IJCV}, 115:\penalty0 211--252, 2015.

\bibitem[Shao et~al.(2019)Shao, Li, Zhang, Peng, Yu, Zhang, Li, and Sun]{shao2019objects365}
Shuai Shao, Zeming Li, Tianyuan Zhang, Chao Peng, Gang Yu, Xiangyu Zhang, Jing Li, and Jian Sun.
\newblock Objects365: A large-scale, high-quality dataset for object detection.
\newblock In \emph{CVPR}, pages 8430--8439, 2019.

\bibitem[Smith and Topin(2017)]{smith2019super}
Leslie~N Smith and Nicholay Topin.
\newblock Super-convergence: Very fast training of neural networks using large learning rates.
\newblock \emph{arXiv preprint arXiv:1708.07120}, 2017.

\bibitem[Von~Luxburg(2007)]{von2007tutorial}
Ulrike Von~Luxburg.
\newblock A tutorial on spectral clustering.
\newblock \emph{Statistics and computing}, 17:\penalty0 395--416, 2007.

\bibitem[Wang et~al.(2016)Wang, Yang, Mao, Huang, Huang, and Xu]{wang2016cnn}
Jiang Wang, Yi Yang, Junhua Mao, Zhiheng Huang, Chang Huang, and Wei Xu.
\newblock Cnn-rnn: A unified framework for multi-label image classification.
\newblock In \emph{CVPR}, pages 2285--2294, 2016.

\bibitem[Wang et~al.(2025)Wang, Zhan, Ma, Tao, Ding, and Gong]{wang2025splicemix}
Lei Wang, Yibing Zhan, Leilei Ma, Dapeng Tao, Liang Ding, and Chen Gong.
\newblock Splicemix: A cross-scale and semantic blending augmentation strategy for multi-label image classification.
\newblock \emph{IEEE TMM}, 2025.

\bibitem[Wang et~al.(2024)Wang, Cheng, Fang, Zhang, Duan, and Wang]{wang2024revisiting}
Yuzhu Wang, Lechao Cheng, Chaowei Fang, Dingwen Zhang, Manni Duan, and Meng Wang.
\newblock Revisiting the power of prompt for visual tuning.
\newblock In \emph{ICML}, pages 50233--50247, 2024.

\bibitem[Wang et~al.(2017)Wang, Chen, Li, Xu, and Lin]{wang2017lstm}
Zhouxia Wang, Tianshui Chen, Guanbin Li, Ruijia Xu, and Liang Lin.
\newblock Multi-label image recognition by recurrently discovering attentional regions.
\newblock In \emph{ICCV}, pages 464--472, 2017.

\bibitem[Wei et~al.(2015)Wei, Xia, Lin, Huang, Ni, Dong, Zhao, and Yan]{wei2015hcp}
Yunchao Wei, Wei Xia, Min Lin, Junshi Huang, Bingbing Ni, Jian Dong, Yao Zhao, and Shuicheng Yan.
\newblock Hcp: A flexible cnn framework for multi-label image classification.
\newblock \emph{IEEE TPAMI}, 38\penalty0 (9):\penalty0 1901--1907, 2015.

\bibitem[Wu et~al.(2020)Wu, Chen, Li, Xiao, and Hu]{wu2020adahgnn}
Xiangping Wu, Qingcai Chen, Wei Li, Yulun Xiao, and Baotian Hu.
\newblock Adahgnn: Adaptive hypergraph neural networks for multi-label image classification.
\newblock In \emph{ACM MM}, pages 284--293, 2020.

\bibitem[Xie and Huang(2021)]{xie2021partial}
Ming-Kun Xie and Sheng-Jun Huang.
\newblock Partial multi-label learning with noisy label identification.
\newblock \emph{IEEE TPAMI}, 44\penalty0 (7):\penalty0 3676--3687, 2021.

\bibitem[Xie et~al.(2023)Xie, Xiao, Liu, Niu, Sugiyama, and Huang]{xie2023class}
Ming-Kun Xie, Jiahao Xiao, Hao-Zhe Liu, Gang Niu, Masashi Sugiyama, and Sheng-Jun Huang.
\newblock Class-distribution-aware pseudo-labeling for semi-supervised multi-label learning.
\newblock \emph{NeurIPS}, 36:\penalty0 25731--25747, 2023.

\bibitem[Xie et~al.(2024)Xie, Xiao, Peng, Niu, Sugiyama, and Huang]{xiecounterfactual}
Ming-Kun Xie, Jia-Hao Xiao, Pei Peng, Gang Niu, Masashi Sugiyama, and Sheng-Jun Huang.
\newblock Counterfactual reasoning for multi-label image classification via patching-based training.
\newblock In \emph{ICML}, 2024.

\bibitem[Xu et~al.(2022)Xu, Huang, Zhou, Huangfu, Zeng, and Liu]{xuboosting}
Jiazhi Xu, Sheng Huang, Fengtao Zhou, Luwen Huangfu, Daniel Zeng, and Bo Liu.
\newblock Boosting multi-label image classification with complementary parallel self-distillation.
\newblock In \emph{IJCAI}, 2022.

\bibitem[Ye et~al.(2020)Ye, He, Peng, Wu, and Qiao]{ye2020attention}
Jin Ye, Junjun He, Xiaojiang Peng, Wenhao Wu, and Yu Qiao.
\newblock Attention-driven dynamic graph convolutional network for multi-label image recognition.
\newblock In \emph{ECCV}, pages 649--665. Springer, 2020.

\bibitem[Yin et~al.(2024)Yin, Gan, He, Gao, and Zhang]{yinhybrid}
Zihao Yin, Chen Gan, Kelei He, Yang Gao, and Junfeng Zhang.
\newblock Hybrid sharing for multi-label image classification.
\newblock In \emph{ICLR}, 2024.

\bibitem[Yoo et~al.(2023)Yoo, Kim, Jung, Lee, and Yoon]{yoo2023improving}
Seungryong Yoo, Eunji Kim, Dahuin Jung, Jungbeom Lee, and Sungroh Yoon.
\newblock Improving visual prompt tuning for self-supervised vision transformers.
\newblock In \emph{ICML}, pages 40075--40092, 2023.

\bibitem[Zhan and Zhang(2017)]{zhan2017inductive}
Wang Zhan and Min-Ling Zhang.
\newblock Inductive semi-supervised multi-label learning with co-training.
\newblock In \emph{KDD}, pages 1305--1314, 2017.

\bibitem[Zhang and Fang(2020)]{zhang2020partial}
Min-Ling Zhang and Jun-Peng Fang.
\newblock Partial multi-label learning via credible label elicitation.
\newblock \emph{IEEE TPAMI}, 43\penalty0 (10):\penalty0 3587--3599, 2020.

\bibitem[Zhang and Zhou(2013)]{zhang2013review}
Min-Ling Zhang and Zhi-Hua Zhou.
\newblock A review on multi-label learning algorithms.
\newblock \emph{IEEE TKDE}, 26\penalty0 (8):\penalty0 1819--1837, 2013.

\bibitem[Zhang et~al.(2024)Zhang, Li, Xie, Zhuang, Guo, and Li]{zhang2024multi}
Yuqi Zhang, Xiucheng Li, Hao Xie, Weijun Zhuang, Shihui Guo, and Zhijun Li.
\newblock Multi-label action anticipation for real-world videos with scene understanding.
\newblock \emph{IEEE TIP}, 2024.

\bibitem[Zheng et~al.(2019)Zheng, Yuan, Zhu, Lin, Cheng, Shi, and Ye]{zheng2019self}
Zhuobin Zheng, Chun Yuan, Xinrui Zhu, Zhihui Lin, Yangyang Cheng, Cheng Shi, and Jiahui Ye.
\newblock Self-supervised mixture-of-experts by uncertainty estimation.
\newblock In \emph{AAAI}, pages 5933--5940, 2019.

\end{thebibliography}
}

\clearpage
\setcounter{page}{1}
\maketitlesupplementary

\section{Supplementary Method}
\subsection{Divide All Classes into Subsets}\label{app:derive-dp}
\begin{figure}[t]
	\centering
	\begin{subfigure}{0.99\linewidth}
		\includegraphics[width=0.98\linewidth]{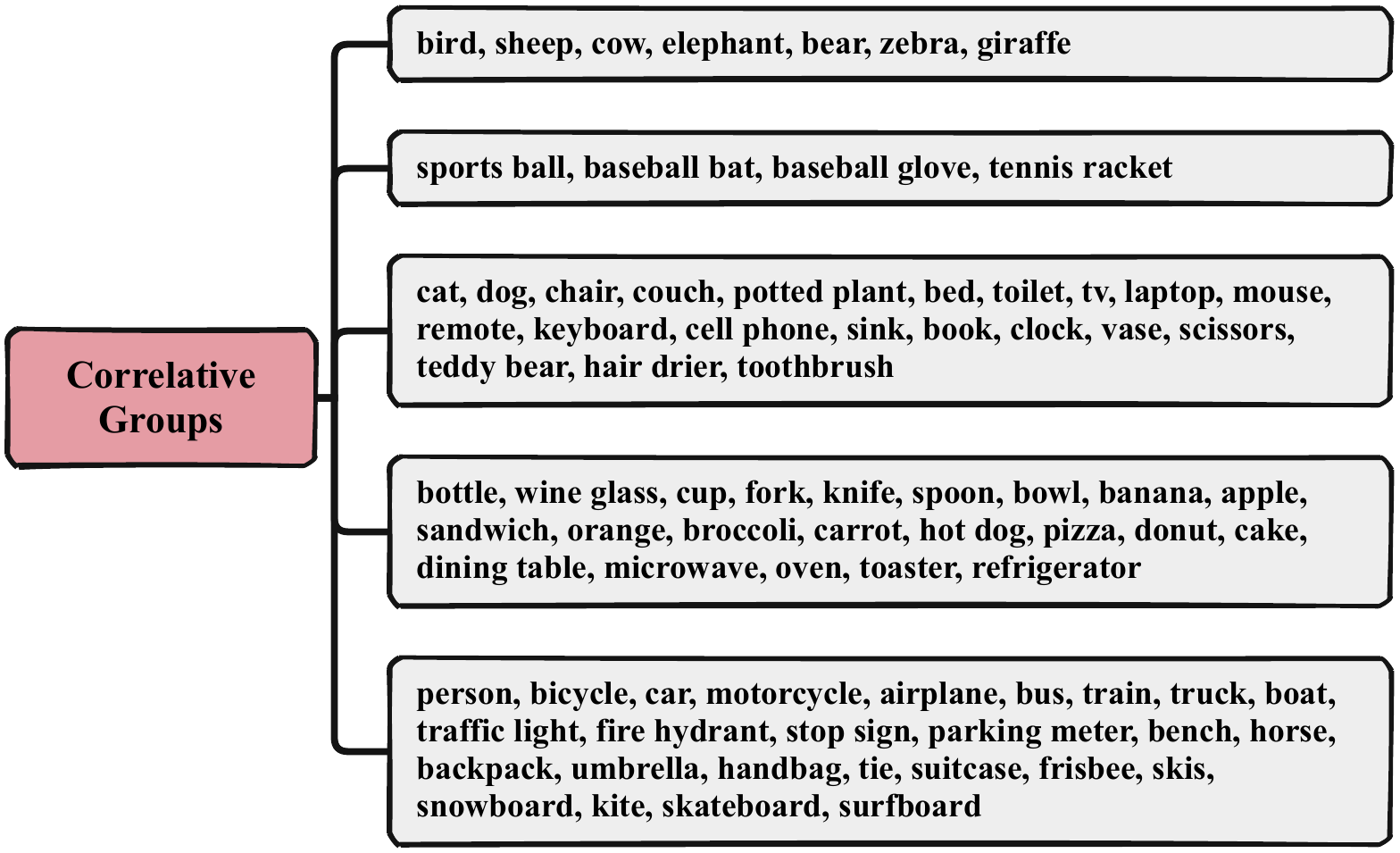}
            \vspace{-0.0em}
		\caption{correlative groups on \texttt{COCO}.}
		\label{fig:app_correlativegroups_name}
	\end{subfigure}
	\hfill
	\begin{subfigure}{0.99\linewidth}
		\includegraphics[width=0.98\linewidth]{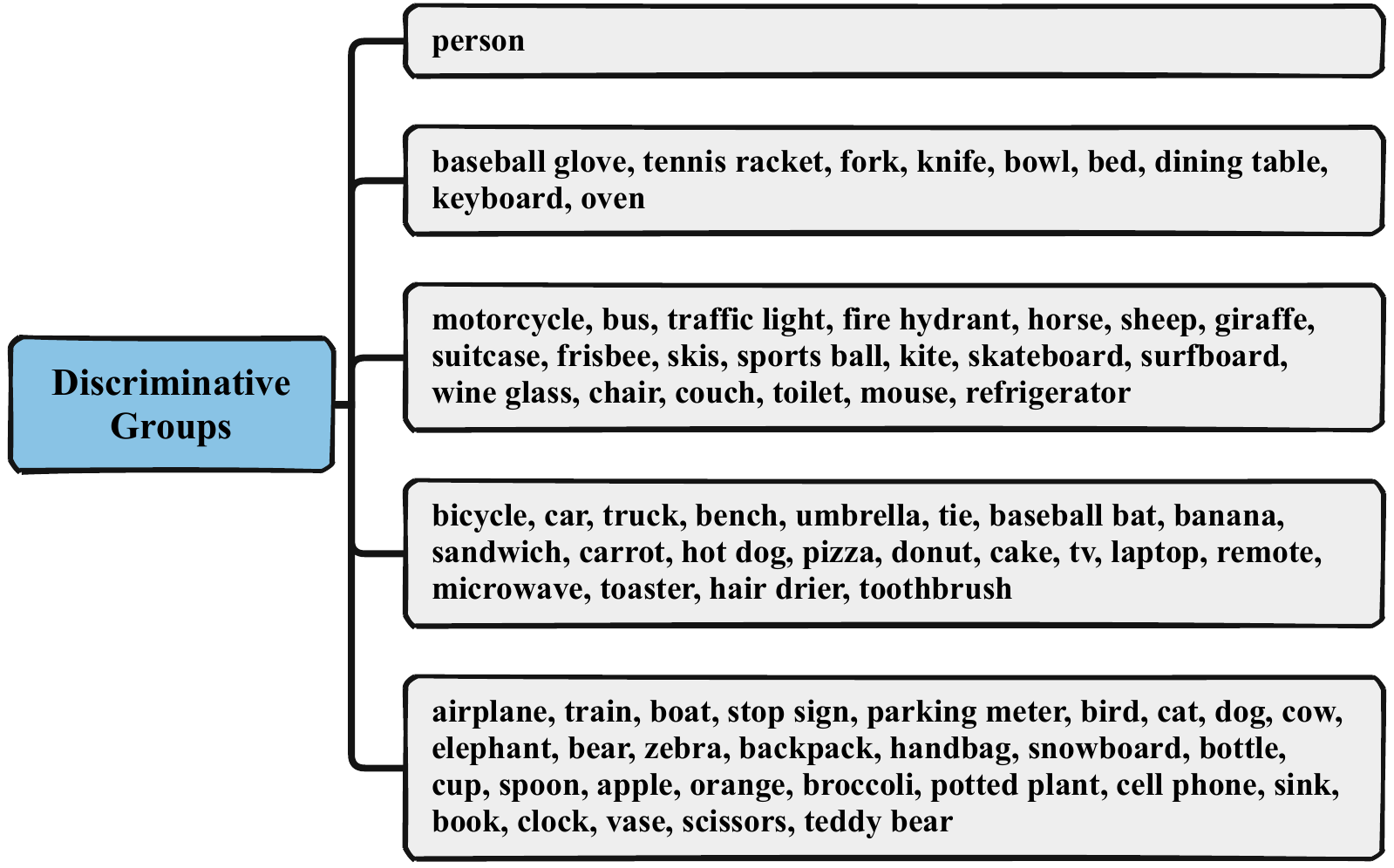}
            \vspace{-0.0em}
		\caption{discriminative groups on \texttt{COCO}.}
		\label{fig:app_discriminativegroups_name}
	\end{subfigure}
        \vspace{-0.0em}
    \caption{An example of grouping all classes in both co-occurrence and dis-occurrence graphs within the \texttt{COCO}, with each graph divided into 5 groups.}
    \label{fig:app_groups_name}
    \vspace{-1.0em}
\end{figure}

In this section, we present a grouping strategy for multi-label image classification (MLC), which divides the labels into several subgroups. Each subgroup is dedicated to addressing different label relationships.
And, a similar method has been shown in BootMLC~\cite{xuboosting}.
Each of these simpler subtasks is processed individually and in parallel, with the correlations among the labels being modeled within each subtask. These are then integrated to formulate a comprehensive solution to the original task.
In our setting, we decompose the modeling of label correlation into co-occurrence and dis-occurrence. Concretely, we construct a co-occurrence graph $\mathcal{G}^+$ and a dis-occurrence graph $\mathcal{G}^-$ to encode the correlative representations between labels and the discriminative representations of each label. Firstly, we count the co-occurrence of label pairs to obtain the co-occurrence matrix $\mathbf{S}\in\mathbb{R}^{K\times K}$, and $\mathbf{S}_{i,j}$ represents the probability of occurrence of label $\bm{y}_{j}$ when label $\bm{y}_{i}$ is present. Subsequently, a smoothing operation and a symmetrization are employed to derive the affinity matrix as follows:
\begin{align*}
\mathbf{M}=
\begin{cases}
  \mathbf{M}^+= (\sqrt[\tau]{\mathbf{S}}+\sqrt[\tau]{\mathbf{S}}^\top)~/~{2},       & \mathcal{G}=\mathcal{G}^+ \\
  \mathbf{M}^-= \mathbf{I}-(\sqrt[\tau]{\mathbf{S}}+\sqrt[\tau]{\mathbf{S}}^\top)~/~{2}, & \mathcal{G}=\mathcal{G}^-
\end{cases}
\end{align*}
where $\tau$ is a positive hyper-parameter to adjust the distribution of co-occurrence matrix $\mathbf{S}$, and the $\sqrt[\tau]{\mathbf{S}}^\top$ denotes its transpose. The symmetrization ensures an undirected graph, with bidirectional connection strengths. The affinity matrix $\mathbf{M}^+$ and $\mathbf{M}^-$ are utilized to encode the co-occurrence relationship and dis-occurrence between categories respectively. Then we leverage them to generate the Laplacian matrix and treat the decomposition problem as a spectral clustering~\cite{von2007tutorial} problem as follows:
\begin{align*}
    \hat{\mathbf{F}}\leftarrow\mathop{\arg\min}\limits_{F} \text{Trace}(\mathbf{F}^{T}\mathbf{L}_{\mathrm{syn}}\mathbf{F}), \quad\text{s.t.}~~\mathbf{F}^{T}\mathbf{F}=\mathbf{I},
\end{align*}
where $\mathbf{L}_{\mathrm{syn}} = \mathbf{I} - \mathbf{D}^{-\frac{1}{2}}\mathbf{M}\mathbf{D}^{-\frac{1}{2}}$ represents the normalized Laplacian matrix and
$\mathbf{D}$ is the degree matrix of graph $\mathcal{G}$. $\mathbf{F}$ is the learned graph embedding of vertices (categories), and $\hat{\mathbf{F}}$ indicates the top-$k$ minimum eigenvectors of $\mathbf{L}_{\mathrm{syn}}$. We cluster the $\hat{\mathbf{F}}$ via the $k$-means algorithm into clusters $\{\mathcal{C}_t\}_{t=1}^T$. Ultimately, the original task can be decomposed into $T$ sub-tasks $\{\mathcal{T}_t\}_{t=1}^T$ according to the clustered class subset. Correspondingly, sub-tasks $\{\mathcal{T}^+_t\}_{t=1}^T$ derived from the graph $\mathcal{G}^+$, acts as a guide for the model to learn the shared representations under co-occurrence relationship, whereas sub-tasks $\{\mathcal{T}^{-}_t\}_{t=1}^T$ generated from graph $\mathcal{G}^-$ promotes the model to focus on learning discriminative representations for each class by masking the correlations among labels. In \cref{fig:app_groups_name}, we present an example of grouping classes on \texttt{COCO}.

\subsection{Optimizing Objective in MLC}\label{loss:cls}
Most MLC methods often use Binary Cross-Entropy (BCE) as the loss function, which converts the problem into several binary classification tasks:
\begin{equation*}\label{eq:bce}
\begin{aligned}			
&\mathcal{L}_{\mathrm{BCE}}(f(\bm{x}),\bm{y})\!=\!\sum\nolimits_{k=1}^L[{y}_k\ell_{1}(f_k(\bm{x})) \!+\!(1\!-\!{y}_k)\ell_{0}(f_k(\bm{x}))].
\end{aligned}
\end{equation*}
Here, $\ell_{1}(f_k)\!=\!-(1-f_k)\log(f_k)$ and $\ell_{0}(f_k)\!=\!-f_k\log(1-f_k)$ represent the losses calculated on positive and negative labels. To simplify the notation for clarity, we let $f(\bm{x})$ represent the probability distribution of example $\bm{x}$ and $f_{k}(\bm{x})$ denote the probability of the $k$-th class.

In MLC, the imbalance between positive and negative classes in each instance is a common issue. To address this, we adopt the asymmetric loss (ASL)~\citep{ridnik2021asymmetric} as the multi-label classification loss, as suggested by prior researchs~\citep{liu2021query2label,ma2023semantic,xiecounterfactual}. ASL dynamically down-weights the easy negative samples and directs the optimization process towards the positive samples:
\label{eq:asl}
\begin{align*}
\mathcal{L}_{\mathrm{ASL}}(f(\bm{x}),\bm{y}) \!&=\! \sum\nolimits_{k=1}^{K} [y_k\ell_1(f_k(\bm{x})) \!+\! (1\!-\!y_k)\ell_0(f_k(\bm{x}))]\text{,} \\
~\mathrm{s.t.}\quad\ell_1(f_k) &= -(1-f_k)^{\lambda_1}\log(f_k)~\text{,} \\
\ell_0(f_k) &= -(f_k)^{\lambda_0}\log(1-f_k)~\text{.}
\end{align*}
$\ell_1$ and $\ell_0$ calculate the positive and negative class losses, respectively. $\lambda_1 \geq 0$ and $\lambda_0 \geq 0$ are the hyperparameters of the positive and negative classes. Additionally, experiments demonstrate that ASL outperforms BCE.
In this work, we set $\lambda_0$ to 2 and $\lambda_1$ to 0.

\definecolor{codeblue}{rgb}{0.25,0.5,0.5}
\newcommand\mycommfont[1]{\footnotesize\ttfamily\textcolor{codeblue}{#1}}
\SetCommentSty{mycommfont}
\begin{algorithm*}[!t]
\small
\DontPrintSemicolon
\SetNoFillComment
\textbf{Input \& Prepare:} Given a multi-label image dataset $\mathcal{D}=\{(\bm{x}_{i}, \bm{y}_{i}) \}_{i=1}^{N}$ with $K$ classes, and their label co-occurrence graph $\mathcal{G}^{+}$ and dis-occurrence graph $\mathcal{G}^{-}$, where $\mathbf{1} = \mathcal{G}^{+}+\mathcal{G}^{-}$. \\
Grouping classes into multiple groups $\{\mathcal{C}_{t}^{+}\}_{t=1}^{T} \leftarrow {\rm Graph Partition}(\mathcal{G}^{+})$ and $\{\mathcal{C}_{t}^{-}\}_{t=1}^{T} \leftarrow {\rm Graph Partition}(\mathcal{G}^{+})$ \tcp*{Apply a clustering algorithm on the graph ($\mathcal{G}^{+})$ and $\mathcal{G}^{-}$)} 
Freeze the ViT parameters and add a set of learnable prompt tokens ($\mathbf{P}^{+}$ / $\mathbf{P}^{-}$) for $\{\mathcal{C}_{t}^{+}\}_{t=1}^{T}$ and $\{\mathcal{C}_{t}^{-}\}_{t=1}^{T}$, respectively, $\mathrm{ViT}(\bm{x}) = \mathrm{ViT}([\mathbf{P}^{+}, \mathbf{P}^{-}], \bm{x})$ \tcp*{Using VPT technology to build a visual encoder model} 
\For {$k=1$ to $\text{MaxEpoch}$}{
    Obtain group-level representations $\mathbf{Z}^{+} \cup \mathbf{Z}^{-} =\mathrm{ViT}(\bm{x})$ \\ 
    Obtain label-aware representations $\{\mathbf{c}^{+}_{k}\}_{k=1}^{K}= \mathrm{MoE}(\mathbf{Z}^{+})$~,~$\{\mathbf{c}^{-}_{k}\}_{k=1}^{K} = \mathrm{MoE}(\mathbf{Z}^{-})$ \tcp*{Using MoE}
    Calculate logits $\hat{\bm{y}}^{+} = \{\hat{{y}}^{+}_{k}\}_{k=1}^{K}=\mathrm{Classifier}^{+}(\{\mathbf{c}^{+}_{k}\}_{k=1}^{K})$~,~$\hat{\bm{y}}^{-}=\{\hat{{y}}^{-}_{k}\}_{k=1}^{K}=\mathrm{Classifier}^{-}(\{\mathbf{c}^{-}_{k}\}_{k=1}^{K})$ \\
    Update model $f(\cdot)$ with $\mathcal{L}_{\mathrm{ASL}}(\bm{\hat{y}}^{+}, \bm{y}) + \mathcal{L}_{\mathrm{ASL}}(\bm{\hat{y}}^{-}, \bm{y})$}
\textbf{Output:} The trained multi-label visual prompt tuning model $f(\cdot)$. \\
\textbf{Predict: $\bm{\hat{y}} = 0.5 \cdot (\bm{\hat{y}}^{+} + \bm{\hat{y}}^{-}) = f(\bm{x})$ }
\caption{Multi-Label Visual Prompt Tuning for Multi-Label Image Classification}
\label{alg:mlvpt}
\end{algorithm*}

\subsection{Pseudo-Code of Proposed Method}
In order to describe our proposed algorithm more clearly, we summarize it in the form of pseudo-code in \cref{alg:mlvpt}.

\section{Supplementary Experiments}
\begin{table}[h]
    \caption{Statistics for the popular benchmark dataset, including the number of training images, test images, categories, and average number of labels per image.}
    \vspace{-1.0em}
    \centering{
    \resizebox{0.90\linewidth}{!}{
	\begin{tabular}{l|cccc}
        \midrule
        \midrule
	Dataset                              &  \# Train     &  \# Test  & \# Classes   & \# Avg. Pos.\\
	\midrule
	Pascal VOC 2007 (\texttt{VOC07})     &      5,011    &    4,952  &     20       &  1.5 \\
	MS-COCO 2014 (\texttt{COCO})         &     82,081    &   40,504  &     80       &  2.9 \\
	NUS-WIDE (\texttt{NUS})              &    126,034    &   84,226  &     81       &  2.4 \\
        Visual-Genome (\texttt{VG256})       &     75,773    &   32,475  &     256      &  7.3 \\
        Objects365 (\texttt{O365})     &  1,742,292    &  193,588  &     365      &  6.1 \\
        \midrule
        \midrule
	\end{tabular}
    }}
    \label{tb:dataset}
    \vspace{-1.0em}
\end{table}
\subsection{Dataset}\label{App:data}
In \cref{tb:dataset}, we present four key characteristics of three benchmark datasets, including the number of training images, the number of test images, and the average number of labels per image.
{Pascal VOC 2007}~\citep{everingham2010pascal} is a popular multi-label dataset containing 20 object categories, divided into a \texttt{trainval} set with 5,011 samples and a \texttt{test} set with 4,952 samples.
{MS-COCO 2014}~\citep{lin2014microsoft} is another widely used multi-label dataset with 80 common categories, consisting of 82,081 training examples and 40,504 validation examples.
Following prior majority work~\citep{ma2023semantic,chen2019gcn}, we use all of its validation examples as the \texttt{test} set, along with the 82,081 training images as the \texttt{train} set.
{NUS-WIDE}~\citep{chua2009nus} is a web image dataset containing 81 categories, with all images sourced from Flickr. In our experiments, we select 126,034 training images as the \texttt{train} set and 84,226 test images as the \texttt{test} set.
{Visual Genome}~\citep{krishna2017visual} is a knowledge base and dataset containing 108,249 images covering 80,138 categories, each of which is annotated by humans with visual concepts.
Given that most categories have very few samples and many categories share similar semantic concepts, we further processed this dataset. 
Following previous work~\citep{xiecounterfactual},
we merged categories with the same meaning and excluded categories with fewer than 500 images. Finally, we obtained a dataset called \texttt{VG256}, comprising 256 classes and 108,249 images, with 70\% of the images used as the \texttt{train} set and 30\% as the \texttt{test} set.
Objects365 (\texttt{O365}) \cite{shao2019objects365} is a large-scale object detection dataset with more than 1,900,000 images covering 365 different categories of everyday objects. It aims to provide a more comprehensive and diverse object recognition scenario. Compared with \texttt{COCO}, \texttt{O365} provides more categories and images, which is more in line with real scenes and is, therefore, more suitable for multi-label image learning. 
Similar to the \texttt{COCO} dataset, annotation information is unavailable for the \texttt{test} set. Consequently, we designated the \texttt{train} set, which comprises 1,742,292 images, for training, and all validation examples as \texttt{test} set, containing 193,588 images, for testing.
\subsection{Evaluation Metrics}
In this work, beyond mean average precision (mAP), the standard metrics reported in the experimental section include overall precision (OP), overall recall (OR), overall F1 score (OF1), as well as per-category precision (CP), per-category recall (CR), and per-category F1 score (CF1). These metrics are computed as follows:
\begin{align*}
&\mathrm{OP}=\frac{\sum\nolimits_i \mathrm{TP}_i}{\sum\nolimits_i \mathrm{TP}_i + \mathrm{FP}_i}~,  &&\mathrm{OR} = \frac{\sum\nolimits_i \mathrm{TP}_i}{\sum\nolimits_i \mathrm{TP}_i + \mathrm{FN}_i}~, \\
&\mathrm{CP}=\frac{1}{C}\sum\nolimits_i\frac{\mathrm{TP}_i}{\mathrm{TP}_i+\mathrm{FP}_i}~,  &&\mathrm{CR} = \frac{1}{C}\sum\nolimits_i\frac{\mathrm{TP}_i}{\mathrm{TP}_i+\mathrm{FN}_i}~, \\
&\mathrm{OF1}=\frac{2\times \mathrm{OP}\times \mathrm{OR}}{\mathrm{OP}+\mathrm{OR}}~,  &&\mathrm{CF1} = \frac{2\times \mathrm{CP}\times \mathrm{CR}}{\mathrm{CP}+\mathrm{CR}}~,
\end{align*}
where $\mathrm{TP}_i$ is true positive of class $i$, $\mathrm{FP}_i$ is false positive of class $i$, $\mathrm{FN}_i$ is false negative of class $i$.
Among the metrics, $\mathrm{OF1}$ and $\mathrm{CF1}$ are the most significant, as they take both recall and precision into account, offering a more comprehensive evaluation.
Moreover, with the exception of mAP, note that these results may be sensitive to the chosen threshold.

\begin{figure*}[t]
    \centering
    \includegraphics[width=0.99\linewidth]{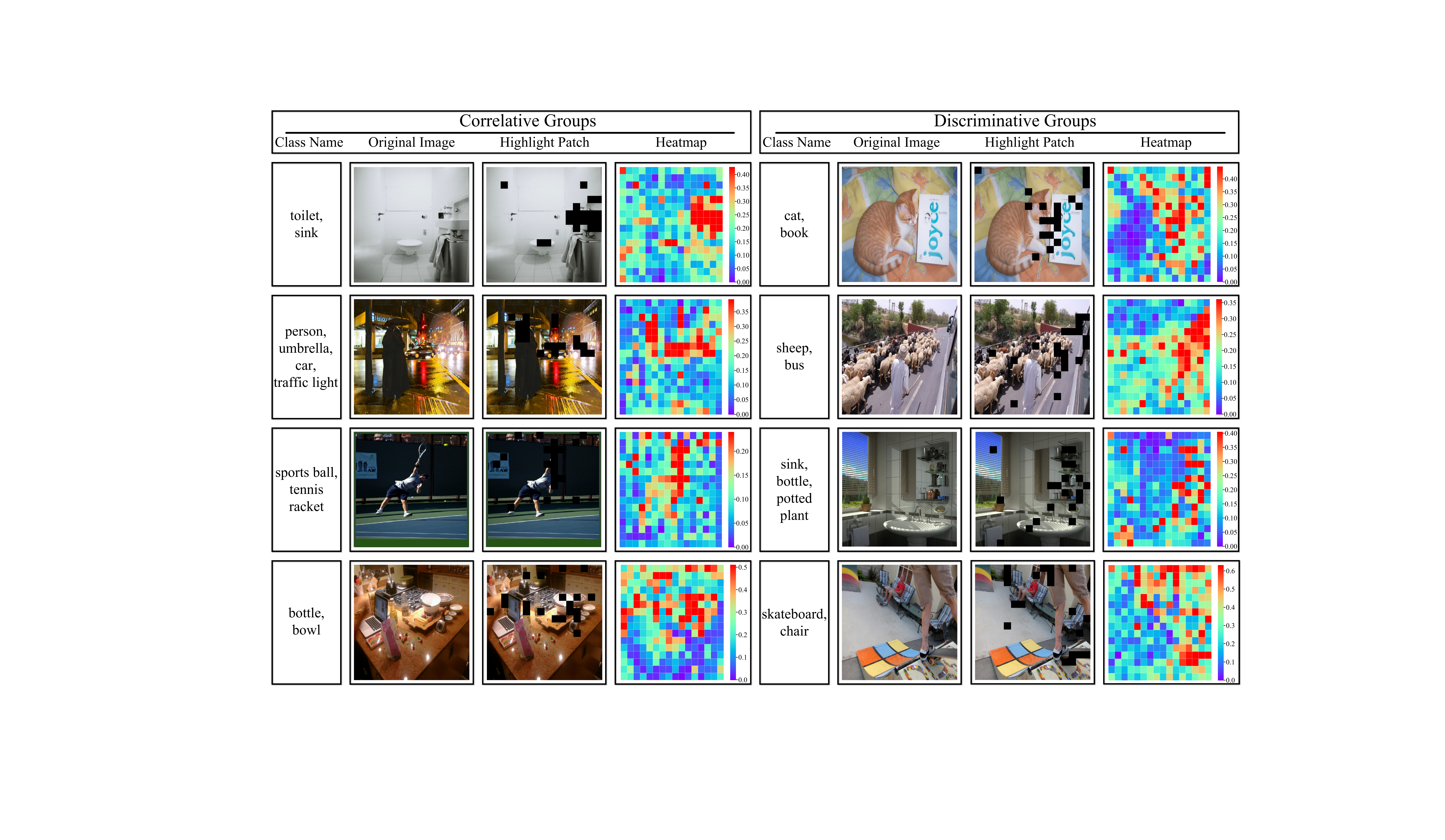}
    \vspace{-1.0em}
    \caption{Visualization of group heatmap from the final layer of the ViT on \texttt{COCO}. The left side is the correlative groups, and the right side is the discriminative groups. As the main body, we also highlight the top-20 image patches based on heatmap scores.}
    \label{fig:app_group_vis}
    \vspace{-1.0em}
\end{figure*}
\subsection{Compared to State-of-the-Art Results}
\noindent \textbf{Performance on Objects365.}
To evaluate our method on a more comprehensive and realistic dataset, we compared our approach with state-of-the-art (SOTA) methods on the \texttt{O365}.
The results are presented in \cref{tab:object365}, demonstrating that our method improves mAP, CF1 and OF1 by 0.9-8.7\%, 0.4-6.6\% and 0.6-5.5\%, respectively.
The effectiveness of our proposed method is verified on a larger dataset with stronger label relationships.

\begin{table}[!t]
\renewcommand\arraystretch{0.50}
\centering
\caption{Comparison of our method with SOTA models on \texttt{O365} at 224 $\times$ 224 resolution. All metrics are in \%.}
\vspace{-1.0em}
\resizebox{0.95\linewidth}{!}
{\begin{tabular}{c|c|c|c|c|c|c|c|c} 
\hline\thickhline
\multicolumn{1}{c|}{\multirow{2}{*}{\cellcolor{igray}}}
&
\multicolumn{1}{c|}{\multirow{2}{*}{\cellcolor{igray}}}
& 
\multicolumn{7}{c}{\cellcolor{igray}Resolution: 224 $\times$ 224} \\ 
\multirow{-2.5}{*}{\cellcolor{igray}Method} & \multirow{-2.5}{*}{\cellcolor{igray}Backbone} & \cellcolor{igray}mAP &  \cellcolor{igray}CP & \cellcolor{igray}CR & \cellcolor{igray}CF1 & \cellcolor{igray}OP & \cellcolor{igray}OR  & \cellcolor{igray}OF1 \\
\hline
\midrule
VPT      & \multicolumn{1}{c|}{\multirow{4}{*}{}}   & 37.6 & 39.3 & 39.4 & 39.4 & 60.4 & 60.6 & 60.5  \\
GateVPT  &                                          & 36.3 & 38.4 & 38.4 & 38.4 & 58.8 & 59.0 & 58.9  \\
E2VPT    &                                          & 37.9 & 39.6 & 39.6 & 39.6 & 60.5 & 60.7 & 60.6  \\
Ours     & \multirow{-5.5}{*}{ViT-B}                & \cellcolor{igray2}40.0 & 41.3 & 41.4 & \cellcolor{igray2}41.3 & 61.7 & 61.8 & \cellcolor{igray2}61.7  \\
\midrule
VPT      & \multicolumn{1}{c|}{\multirow{4}{*}{}}   & 31.5 & 34.3 & 34.4 & 34.4 & 60.2 & 60.4 & 60.3  \\
GateVPT  &                                          & 26.8 & 29.9 & 30.0 & 30.0 & 56.9 & 57.1 & 57.0  \\
E2VPT    &                                          & 29.6 & 32.4 & 32.5 & 32.4 & 59.4 & 59.5 & 59.4  \\
Ours     & \multirow{-5.5}{*}{MAE}                  & \cellcolor{igray2}34.5 & 36.6 & 36.6  & \cellcolor{igray2}36.6 & 62.4 & 62.5 & \cellcolor{igray2}62.5 \\
\midrule
VPT      & \multicolumn{1}{c|}{\multirow{4}{*}{}}   & 31.8 & 34.5 & 34.5 & 34.5 & 58.1 & 58.3 & 58.2  \\
GateVPT  &                                          & 31.8 & 34.5 & 34.5 & 34.5 & 58.1 & 58.3 & 58.2  \\
E2VPT    &                                          & 31.8 & 34.4 & 34.5 & 34.4 & 58.0 & 58.3 & 58.2  \\
Ours     & \multirow{-5.5}{*}{MoCO v3}              & \cellcolor{igray2}33.6 & 35.7 & 35.7 & \cellcolor{igray2}35.7 & 59.1 & 59.2 & \cellcolor{igray2}59.2  \\
\midrule
VPT      & \multicolumn{1}{c|}{\multirow{4}{*}{}}   & 44.3 & 45.0 & 45.1 & 45.1 & 64.1 & 64.3 & 64.2  \\
GateVPT  &                                          & 42.9 & 44.0 & 44.0 & 44.0 & 62.8 & 63.1 & 62.9  \\
E2VPT    &                                          & 44.1 & 44.9 & 44.9 & 44.9 & 64.1 & 64.3 & 64.2  \\
Ours     & \multirow{-5.5}{*}{ViT-B-21k}            & \cellcolor{igray2}45.2 & 45.5 & 45.5 & \cellcolor{igray2}45.5 & 64.7 & 64.8 & \cellcolor{igray2}64.8  \\
\midrule
VPT      & \multicolumn{1}{c|}{\multirow{4}{*}{}}   & 49.3 & 49.3 & 49.3 & 49.3 & 69.6 & 69.8 & 69.7  \\
GateVPT  &                                          & 48.1 & 48.4 & 48.5 & 48.4 & 68.3 & 68.5 & 68.4  \\
E2VPT    &                                          & 49.2 & 49.2 & 49.3 & 49.3 & 69.5 & 69.7 & 69.6  \\
Ours     & \multirow{-5.5}{*}{DINOv2/B}             & \cellcolor{igray2}52.2 & 51.5 & 51.5 & \cellcolor{igray2}51.5 & 70.8 & 71.0 & \cellcolor{igray2}70.9  \\
\midrule
VPT      & \multicolumn{1}{c|}{\multirow{4}{*}{}}   & 40.1 & 41.5 & 41.6 & 41.6 & 64.2 & 64.4 & 64.3  \\
GateVPT  &                                          & 39.0 & 40.8 & 40.8 & 40.8 & 62.7 & 62.9 & 62.8  \\
E2VPT    &                                          & 40.2 & 41.8 & 41.8 & 41.8 & 64.2 & 64.4 & 64.3  \\
Ours     & \multirow{-5.5}{*}{DINOv2/S}             & \cellcolor{igray2}41.4 & 42.7 & 42.8 & \cellcolor{igray2}42.7 & 65.2 & 65.3 & \cellcolor{igray2}65.3  \\
\bottomrule
\end{tabular}}
\label{tab:object365}
\vspace{-1.0em}
\end{table}

\begin{figure}[h]
	\centering
	\begin{subfigure}{0.49\linewidth}
		\includegraphics[width=0.98\linewidth]{figs/COCO_chart_group_ablation_mAP.pdf}
		\caption{mAP on \texttt{COCO}}
		\label{fig:app_coco_map_group}
	\end{subfigure}
	\hfill
	\begin{subfigure}{0.49\linewidth}
		\includegraphics[width=0.98\linewidth]{figs/VOC_chart_group_ablation_mAP.pdf}
		\caption{mAP on \texttt{VOC07}}
		\label{fig:app_vco_map_group}
	\end{subfigure}
	\hfill
	\begin{subfigure}{0.49\linewidth}
		\includegraphics[width=0.98\linewidth]{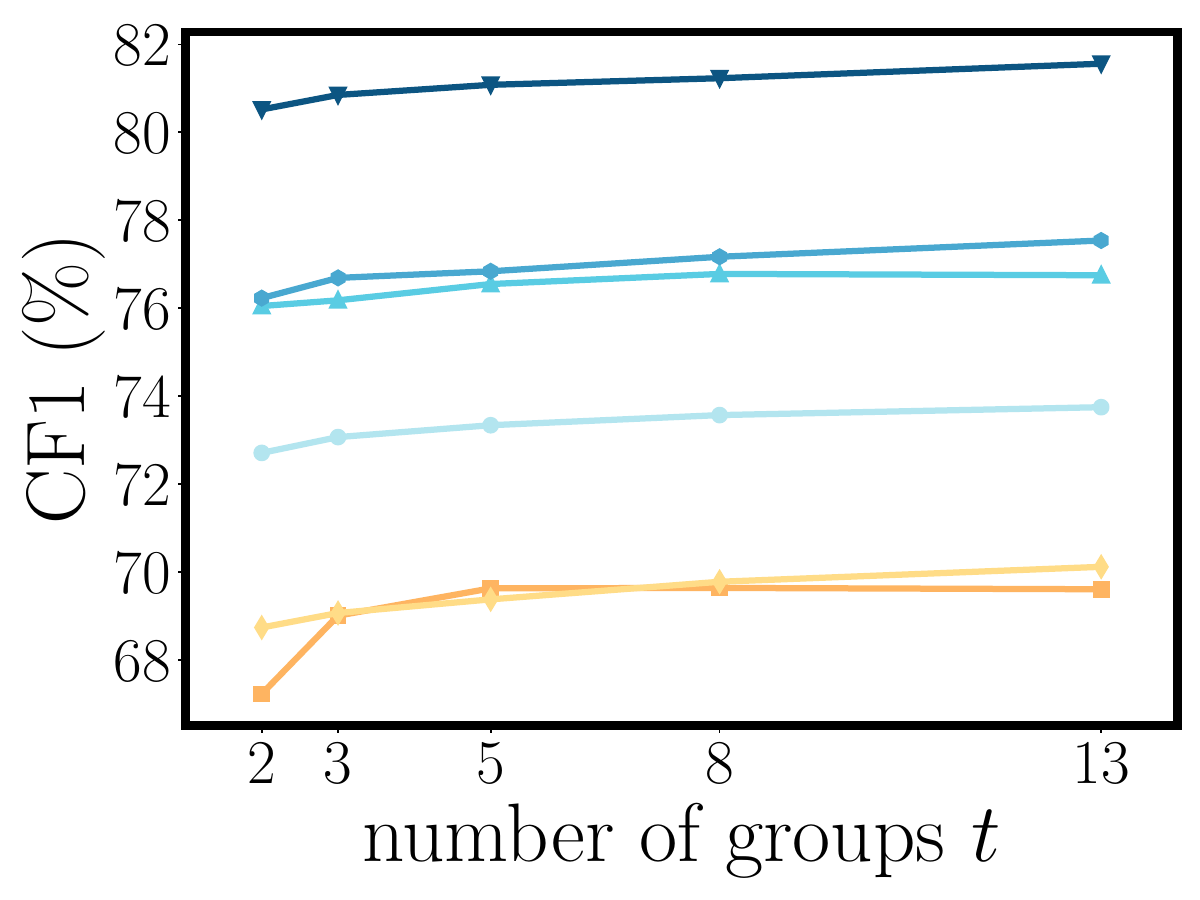}
		\caption{CF1 on \texttt{COCO}}
		\label{fig:app_coco_cf1_group}
	\end{subfigure}
        \hfill
 	\begin{subfigure}{0.49\linewidth}
		\includegraphics[width=0.98\linewidth]{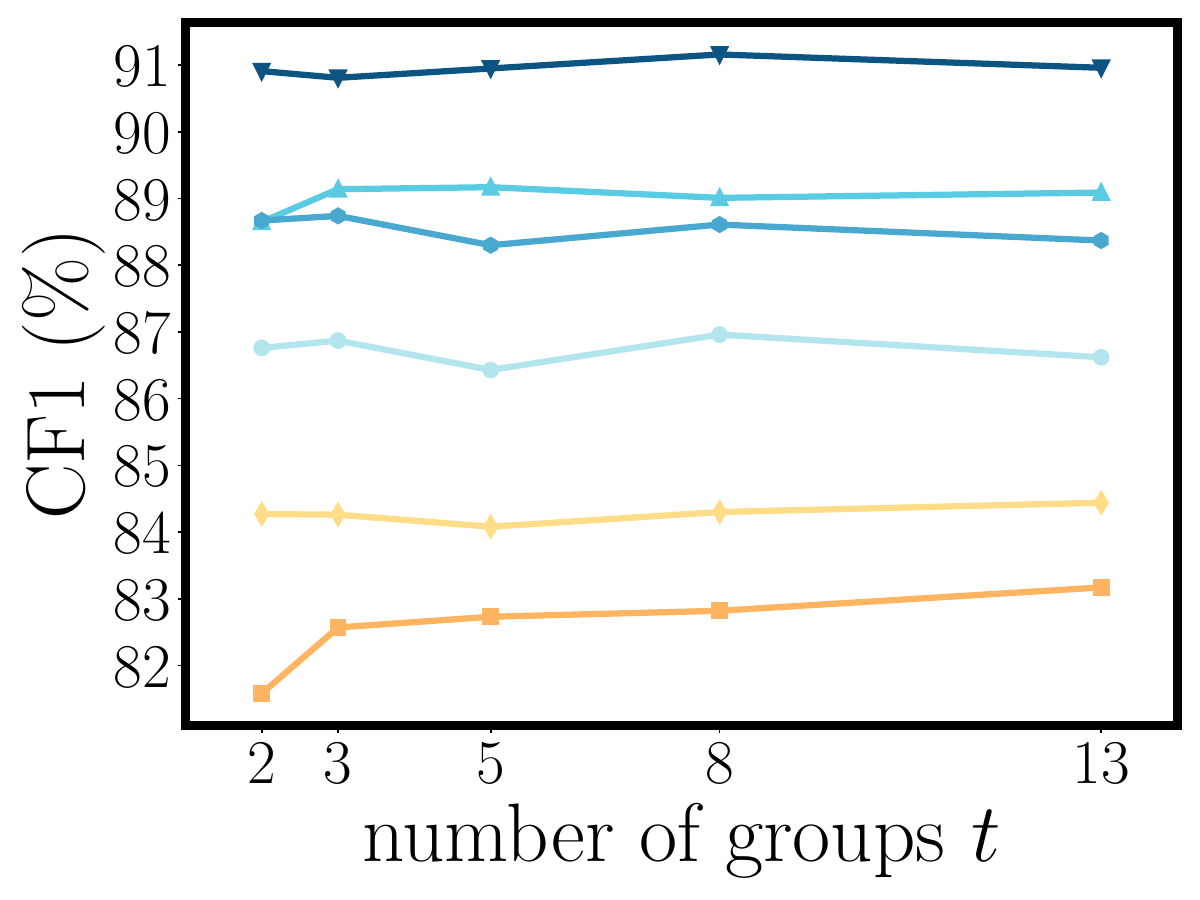}
		\caption{CF1 on \texttt{VOC07}}
		\label{fig:app_voc_cf1_group}
	\end{subfigure}
	\hfill
	\begin{subfigure}{0.49\linewidth}
		\includegraphics[width=0.98\linewidth]{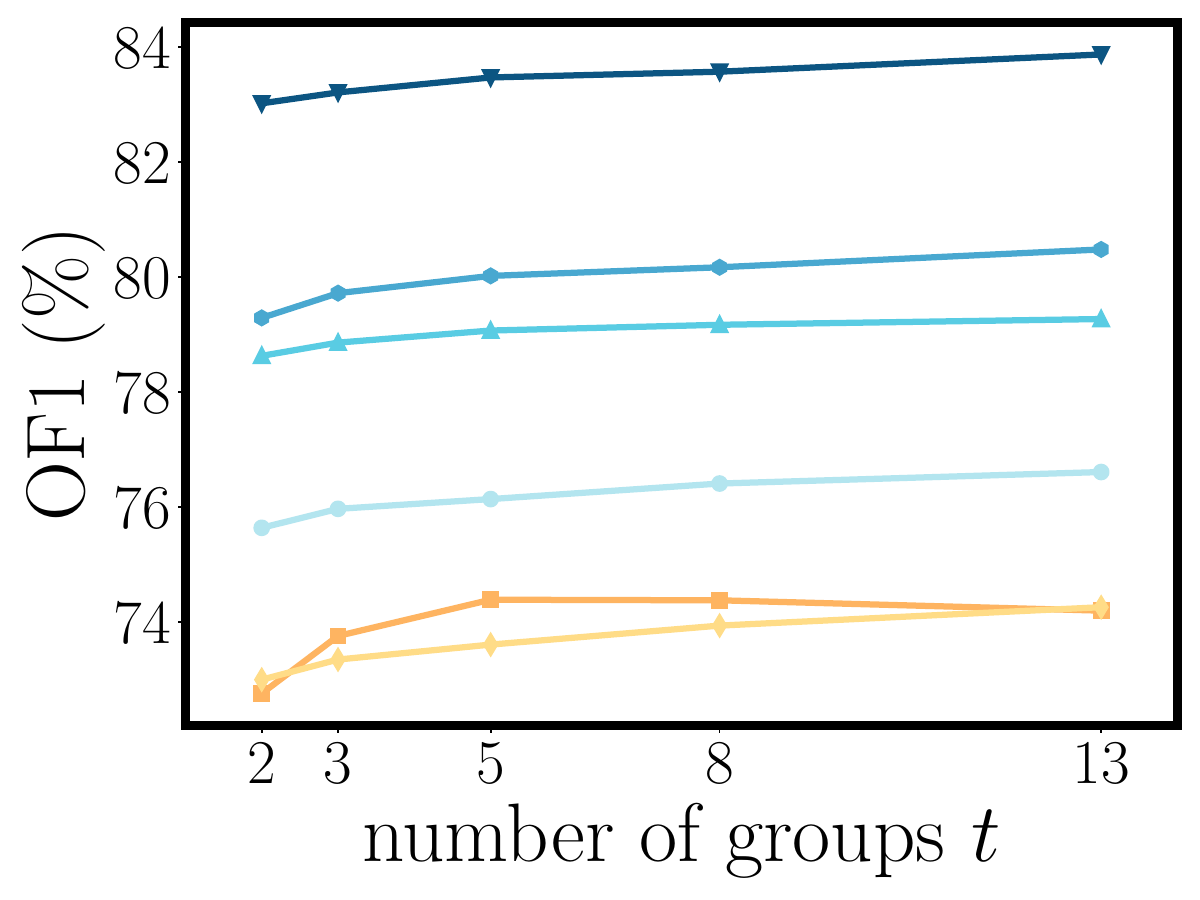}
		\caption{OF1 on \texttt{COCO}}
		\label{fig:app_coco_of1_group}
	\end{subfigure}
	\hfill
	\begin{subfigure}{0.49\linewidth}
		\includegraphics[width=0.98\linewidth]{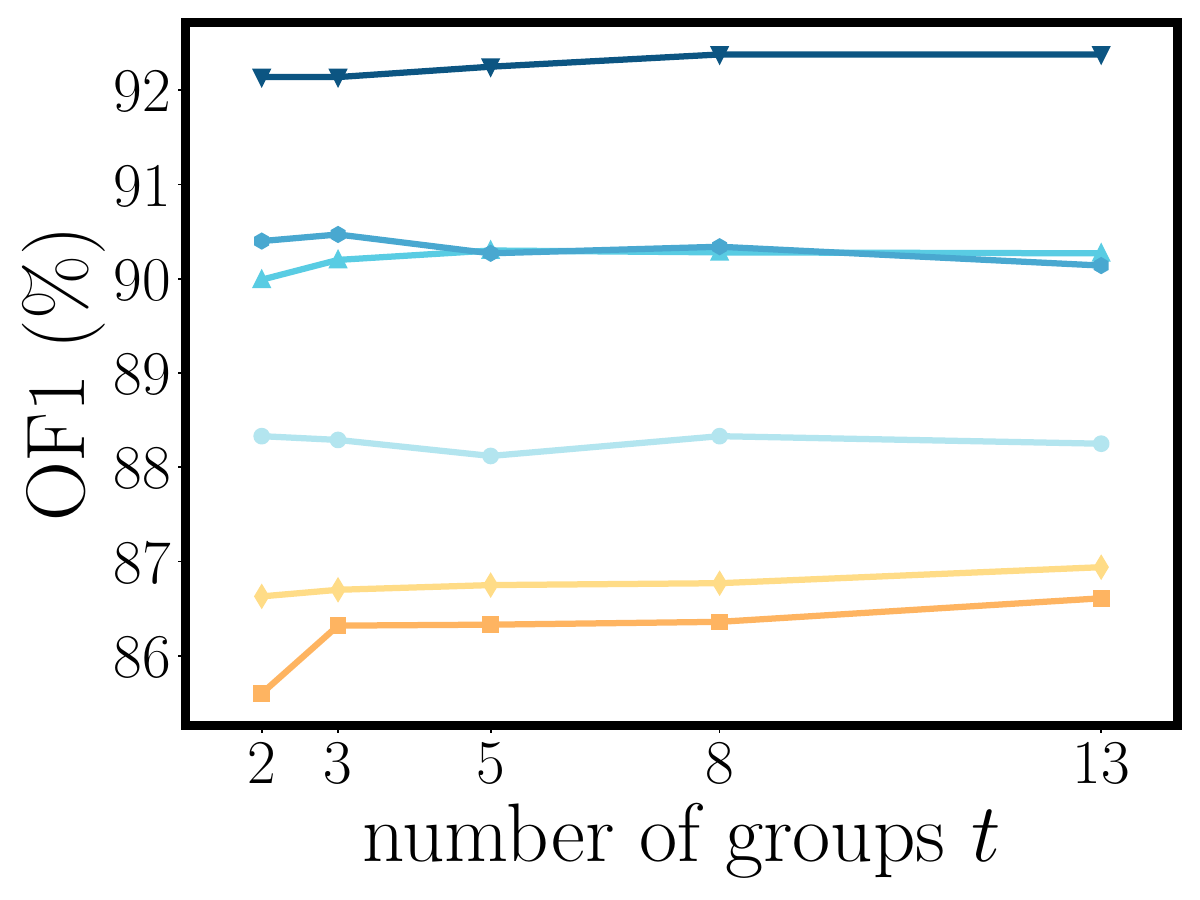}
		\caption{OF1 on \texttt{VOC07}}
		\label{fig:app_voc_of1_group}
	\end{subfigure}
	\caption{The performance curve varies with the increase in the number of groups, in the 3 evaluation metrics: mAP, CF1, and OF1.}
    \label{fig:app_ablation_number_group}
    \vspace{-1.5em}
\end{figure}

\subsection{Diagnostic Experiments}
\noindent \textbf{Number of Groups.} Due to the limitations in the length of the main body, we present a comprehensive evaluation of the number of groups as shown in~\cref{fig:app_ablation_number_group}, which are conducted on \texttt{COCO} and \texttt{VOC07}, including metrics such as mAP, $\mathrm{CF1}$, and $\mathrm{OF1}$. The overall results indicate that increasing the number of groups has improved the performance of our method, although some minor fluctuations were observed.

\begin{figure}[ht]
	\centering
	\begin{subfigure}{0.49\linewidth}
		\includegraphics[width=0.98\linewidth]{figs/COCO_chart_expert_ablation_mAP.pdf}
		\caption{mAP on \texttt{COCO}}
		\label{fig:app_coco_map_expert}
	\end{subfigure}
	\hfill
	\begin{subfigure}{0.49\linewidth}
		\includegraphics[width=0.98\linewidth]{figs/VOC_chart_expert_ablation_mAP.pdf}
		\caption{mAP on \texttt{VOC07}}
		\label{fig:app_voc_map_expert}
	\end{subfigure}
	\hfill
	\begin{subfigure}{0.49\linewidth}
		\includegraphics[width=0.98\linewidth]{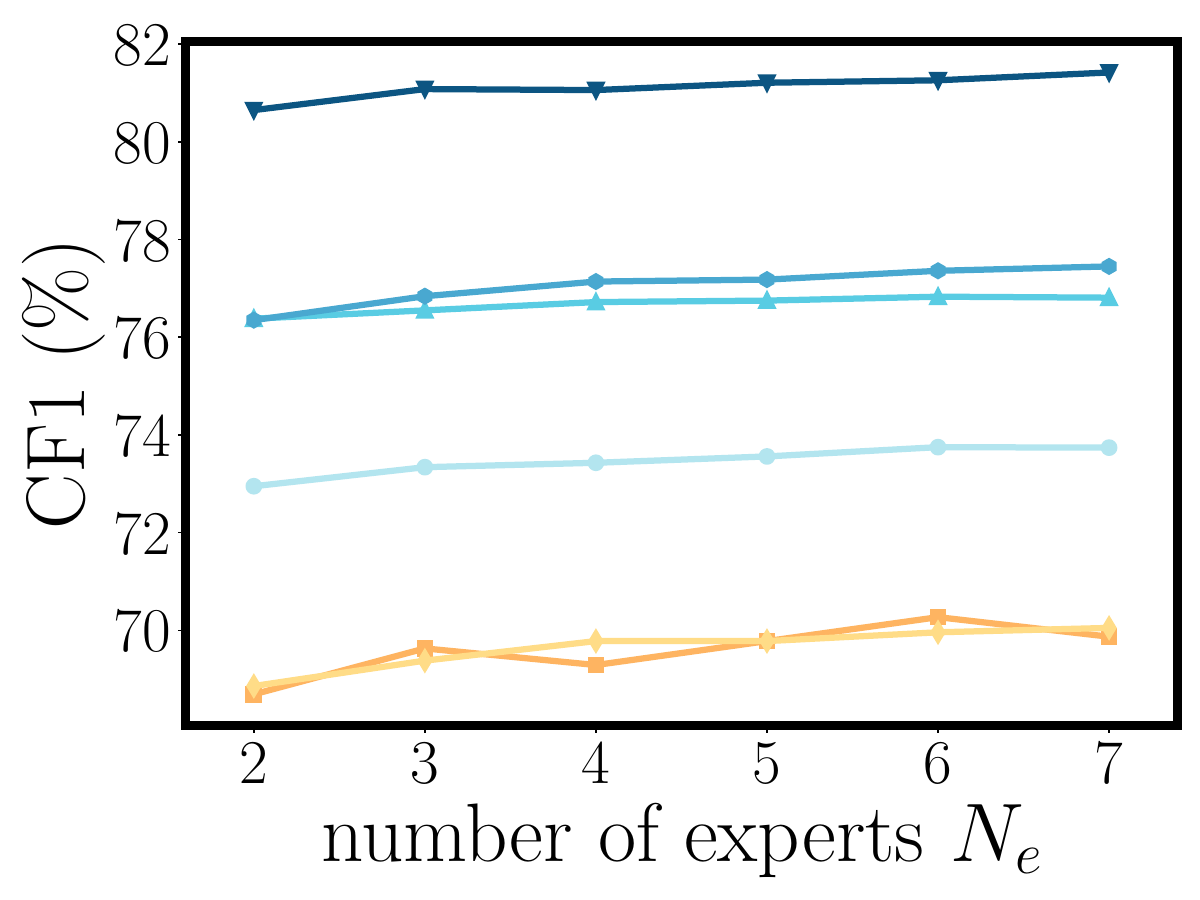}
		\caption{CF1 on \texttt{COCO}}
		\label{fig:app_coco_cf1_expert}
	\end{subfigure}
        \hfill
 	\begin{subfigure}{0.49\linewidth}
		\includegraphics[width=0.97\textwidth]{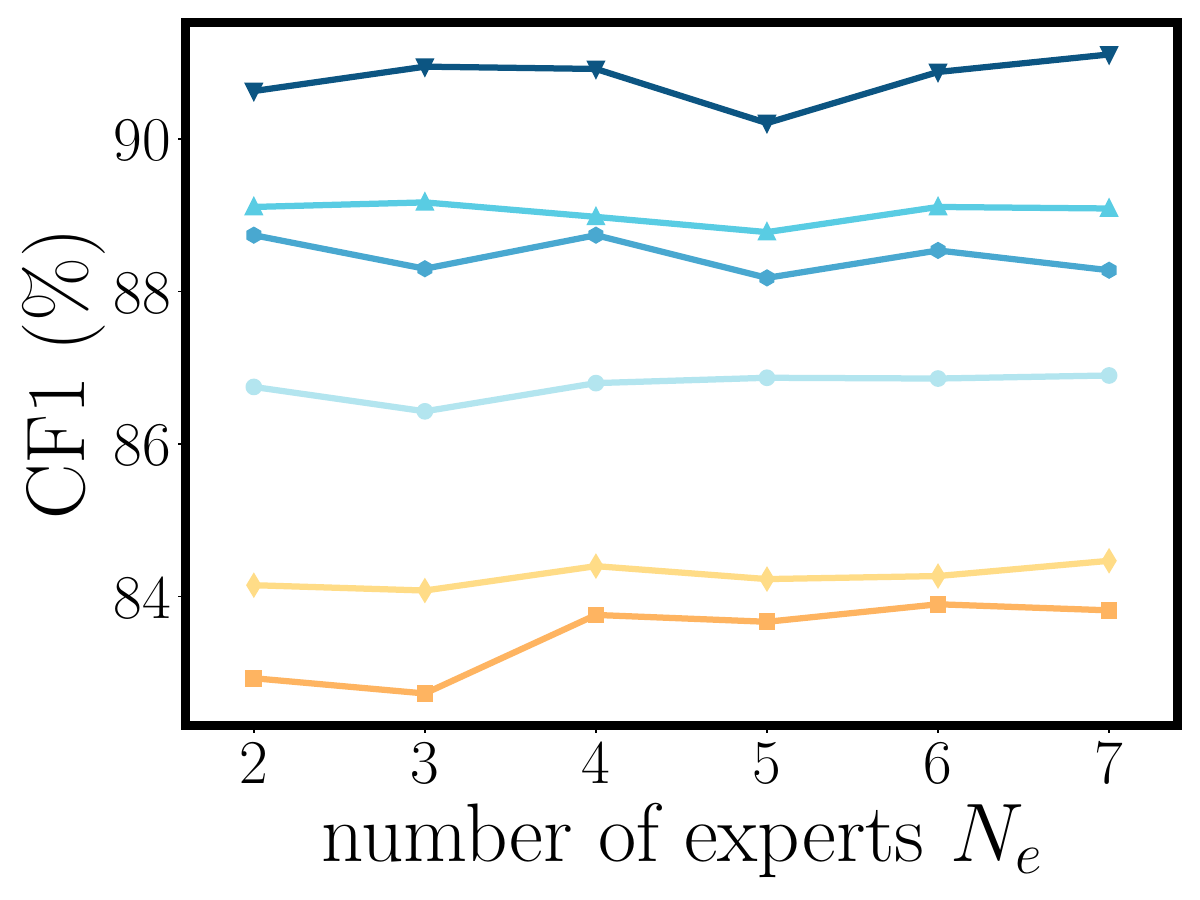}
		\caption{CF1 on \texttt{VOC07}}
		\label{fig:app_voc_cf1_expert}
	\end{subfigure}
	\hfill
	\begin{subfigure}{0.49\linewidth}
		\includegraphics[width=0.98\linewidth]{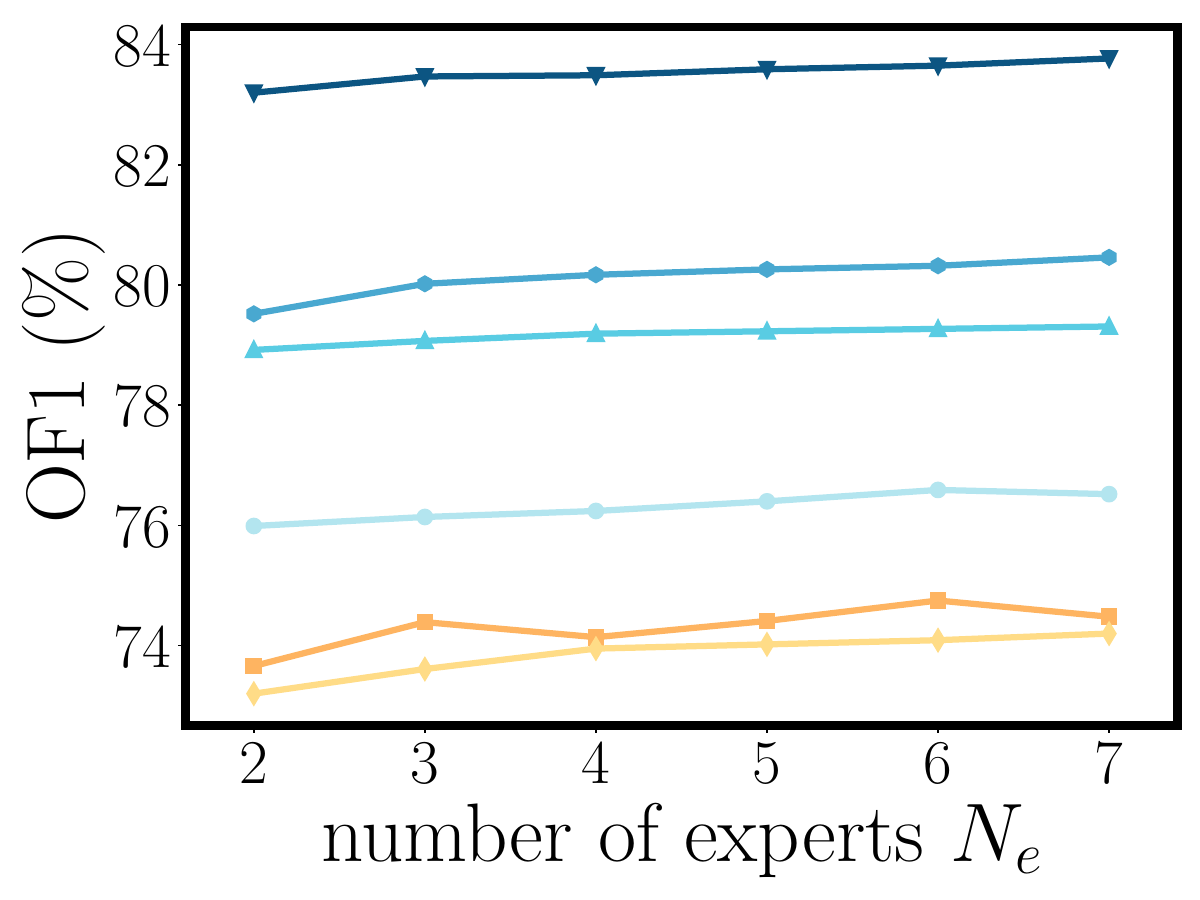}
		\caption{OF1 on \texttt{COCO}}
		\label{fig:app_coco_of1_expert}
	\end{subfigure}
	\hfill
	\begin{subfigure}{0.49\linewidth}
		\includegraphics[width=0.98\linewidth]{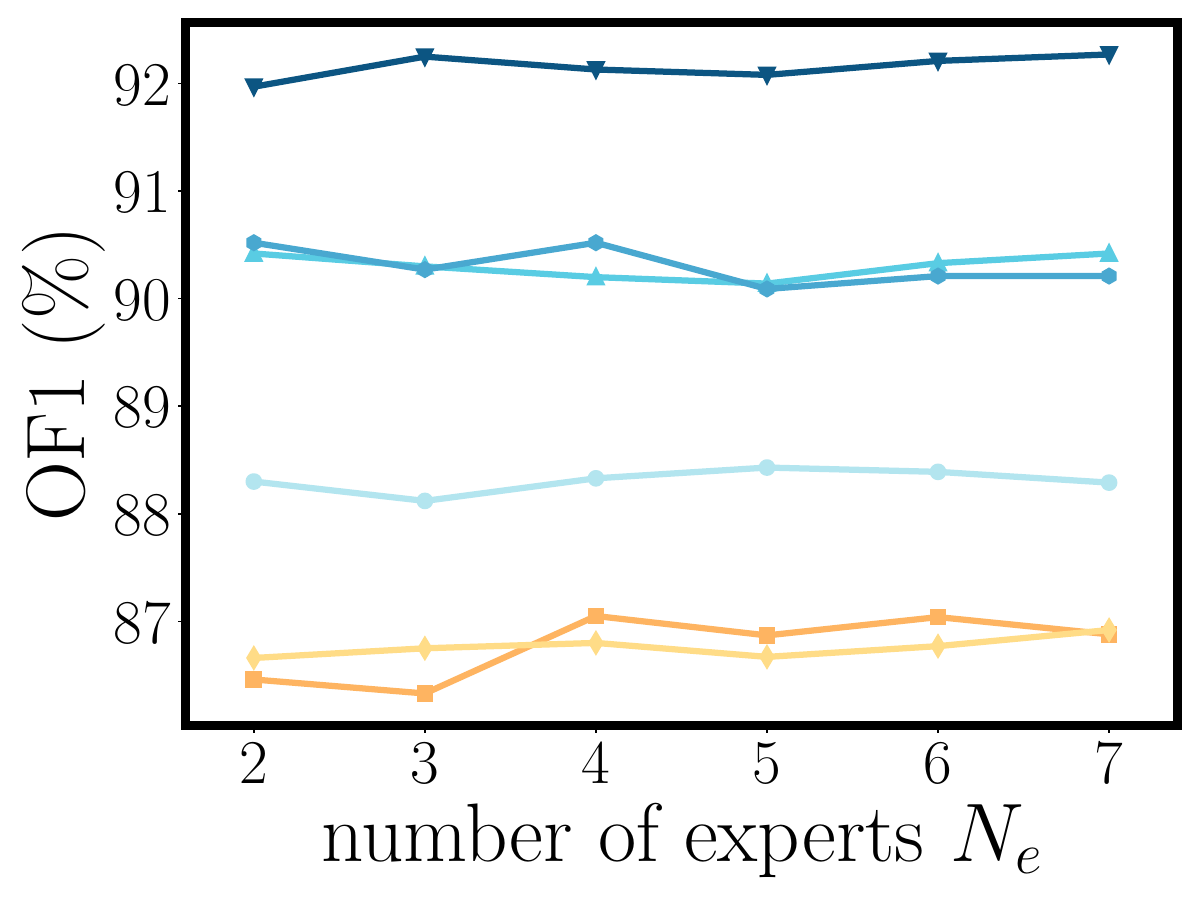}
		\caption{OF1 on \texttt{VOC07}}
		\label{fig:app_voc_of1_expert}
	\end{subfigure}
	\caption{The performance curve varies with the increase in the number of experts, in the mAP, CF1, and OF1.}
    \label{fig:app_ablation_number_expert}
    \vspace{-1.5em}
\end{figure}

\noindent \textbf{Number of Experts.} Due to space constraints in the main body, we present a comprehensive evaluation of the study on the number of experts as shown in \cref{fig:app_ablation_number_expert}, which are conducted on \texttt{COCO} and \texttt{VOC07}, including metrics such as mAP, $\mathrm{CF1}$, and $\mathrm{OF1}$.
The results show that, on the \texttt{COCO} dataset, our method performs better as the number of experts increases. In contrast, on the \texttt{VOC07} dataset, the performance improvement with more experts fluctuates significantly. 
One possible explanation is that the smaller number of categories in each group does not require more prompt tokens to capture label relationships, so the transition from group-aware representation to label-aware representation does not require as many experts.

\begin{figure}[ht]
	\centering
	\begin{subfigure}{0.99\linewidth}
		\includegraphics[width=0.99\linewidth]{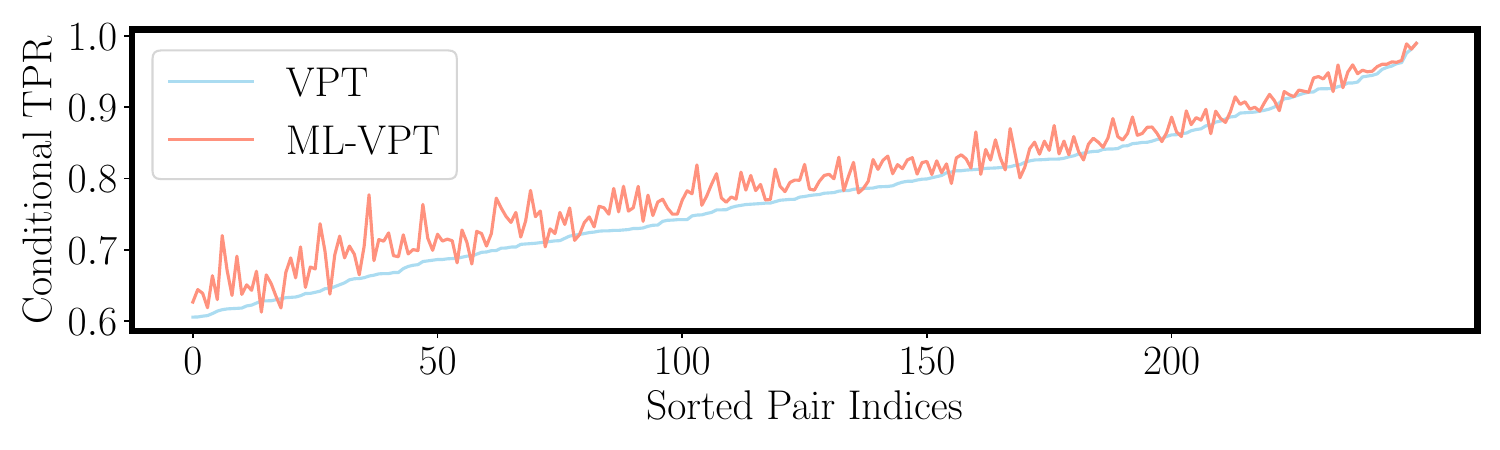}
            \vspace{-0.5em}
		\label{fig:Sorted_TPR}
	\end{subfigure}
	\hfill
	\begin{subfigure}{0.99\linewidth}
		\includegraphics[width=0.99\linewidth]{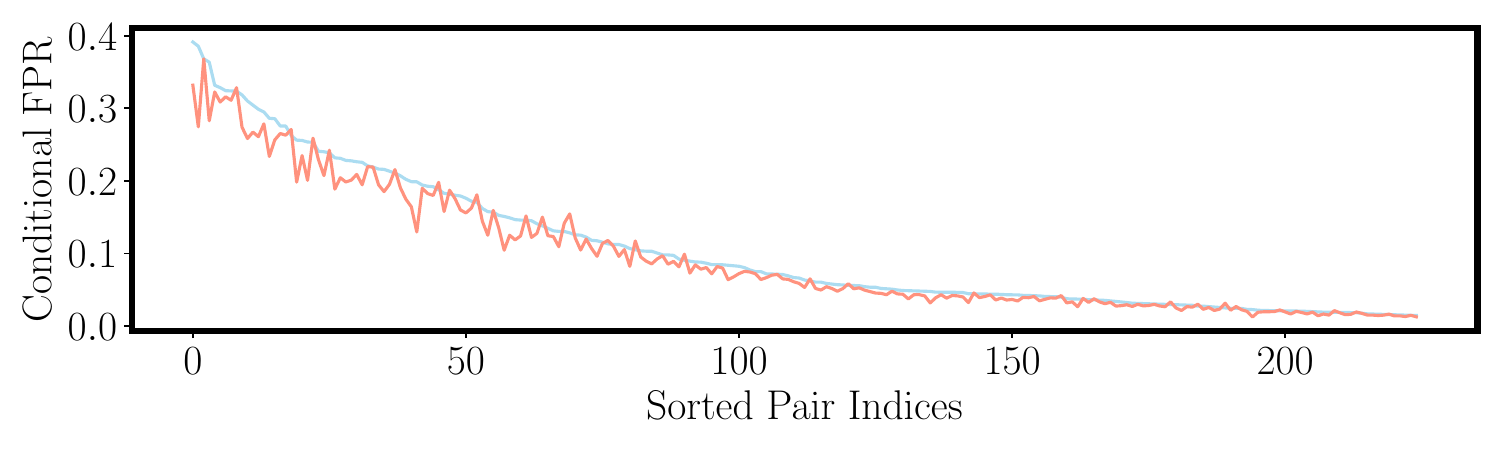}
            \vspace{-0.5em}
		\label{fig:Sorted_FPR}
	\end{subfigure}
	\hfill
        \vspace{-0.5em}
	\caption{Comparison between VPT and ML-VPT (ours) in terms of conditional TPR and FPR, the label pairs with co-occurrence probabilities larger than 0.2 on \texttt{COCO} is selected.}
    \label{fig:app_Sorted_TPRandFPR}
\end{figure}

\noindent \textbf{Effect of Group Strategy.} To further validate the effectiveness of the grouping strategy, we selected label pairs with co-occurrence probabilities greater than 0.2 from on \texttt{COCO}. 
As shown in \cref{fig:app_Sorted_TPRandFPR}, the we proposed ML-VPT outperforms the VPT method in terms of CTPR and CFPR for approximately 88.4\% and 85.3\% of the label pairs respectively.
The higher CTPR indicates that ML-VPT has a stronger ability to capture correlational features, while the lower CFPR suggests that ML-VPT effectively balances both correlational and discriminative features, thereby reducing the risk of overfitting.
Notably, the label pairs in the figure are sorted based on the VPT results for clarity and aesthetic purposes.

\begin{figure}[ht]
	\centering
	\begin{subfigure}{0.99\linewidth}
        \includegraphics[width=0.99\linewidth]{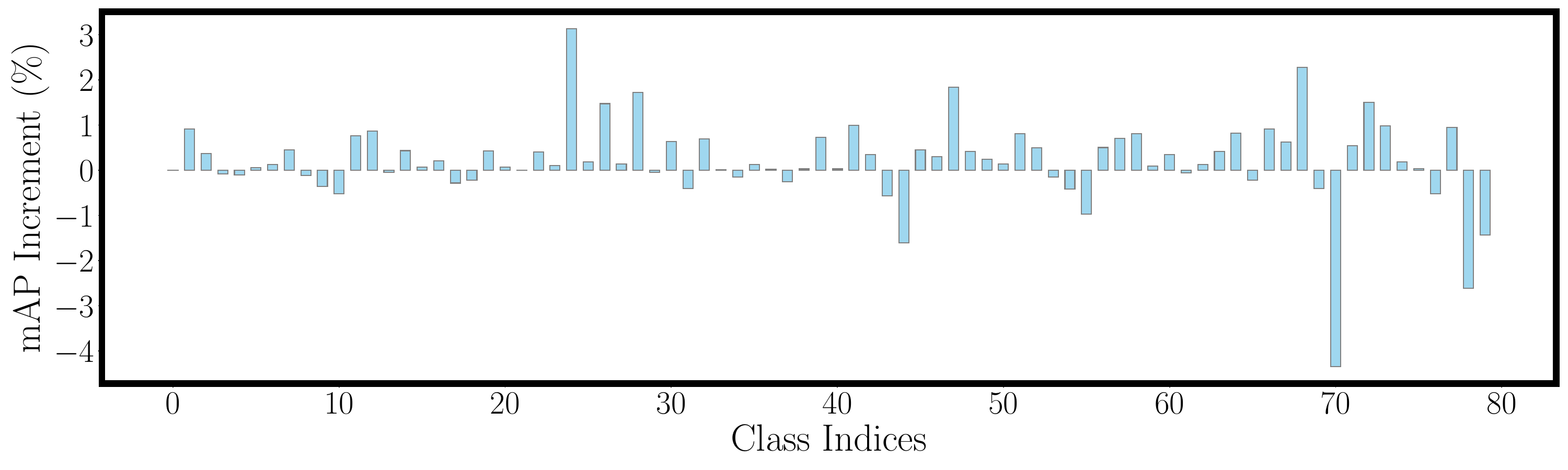}
            \vspace{-0.5em}
		\caption{VPT vs VPT-CO on \texttt{COCO}}
		\label{fig:vpt_co_per_class}
	\end{subfigure}
	\hfill
	\begin{subfigure}{0.99\linewidth}
		\includegraphics[width=0.99\linewidth]{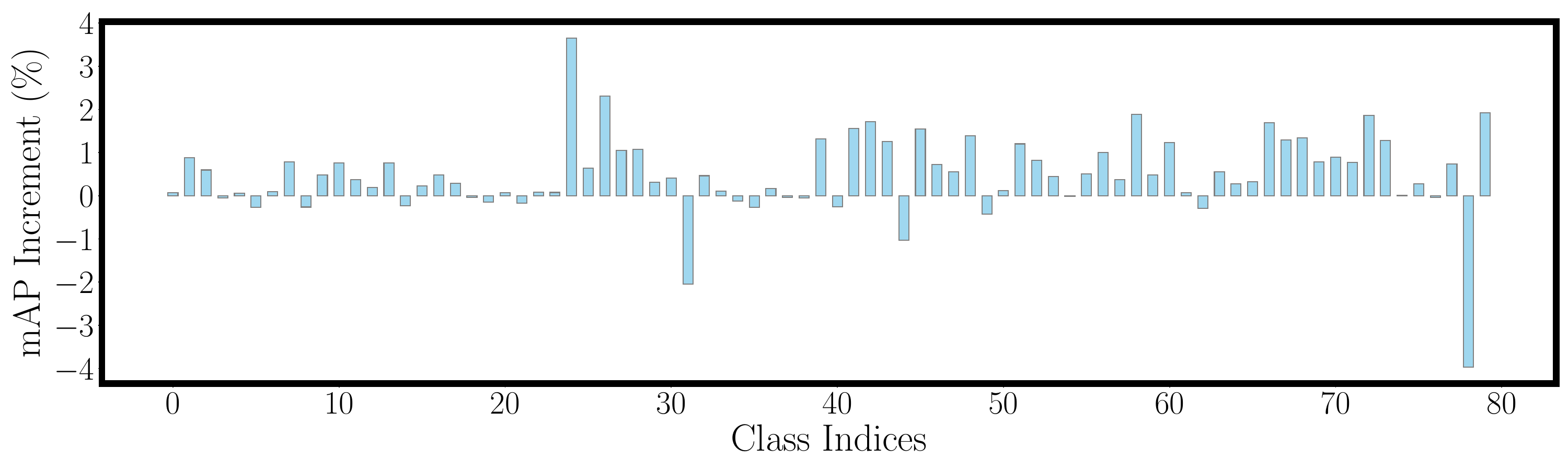}
            \vspace{-0.5em}
		\caption{VPT vs VPT-DC on \texttt{COCO}}
		\label{fig:vpt_do_per_class}
	\end{subfigure}
	\hfill
    	\begin{subfigure}{0.99\linewidth}
		\includegraphics[width=0.99\linewidth]{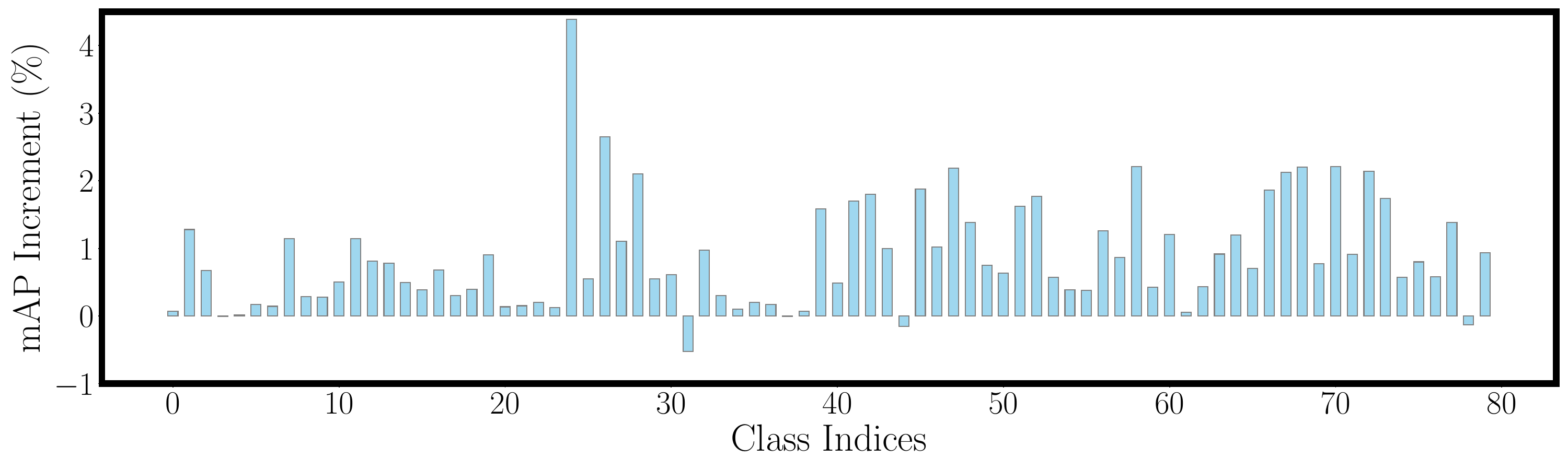}
            \vspace{-0.5em}
		\caption{VPT vs GVPT on \texttt{COCO}}
		\label{fig:vpt_group_per_class}
	\end{subfigure}
	\hfill
        \vspace{-0.5em}
	\caption{Per-class mAP increment for VPT-CO (with only correlative groups), VPT-DC (with only discriminative groups), and GVPT (both correlative and discriminative groups).}
    \label{fig:app_group_per_class}
    \vspace{-0.5em}
\end{figure}

\noindent \textbf{Delve into Grouping Strategy.} Our method divides classes into correlative and discriminative groups to balance their relationships. For comparison, we also conduct experiments with only correlative grouping (VPT-CO) and only discriminative grouping (VPT-DC), as shown in \cref{fig:app_group_per_class}. While both VPT-CO and VPT-DC show an overall improvement in mean Average Precision (mAP) over VPT, they lead to significant accuracy drops for certain classes. Our grouping strategy (GVPT) avoids this issue, providing strong evidence that balancing correlative and discriminative groups is both effective and reasonable.

\begin{figure}[h]
	\centering
	\begin{subfigure}{0.99\linewidth}
		\includegraphics[width=0.99\linewidth]{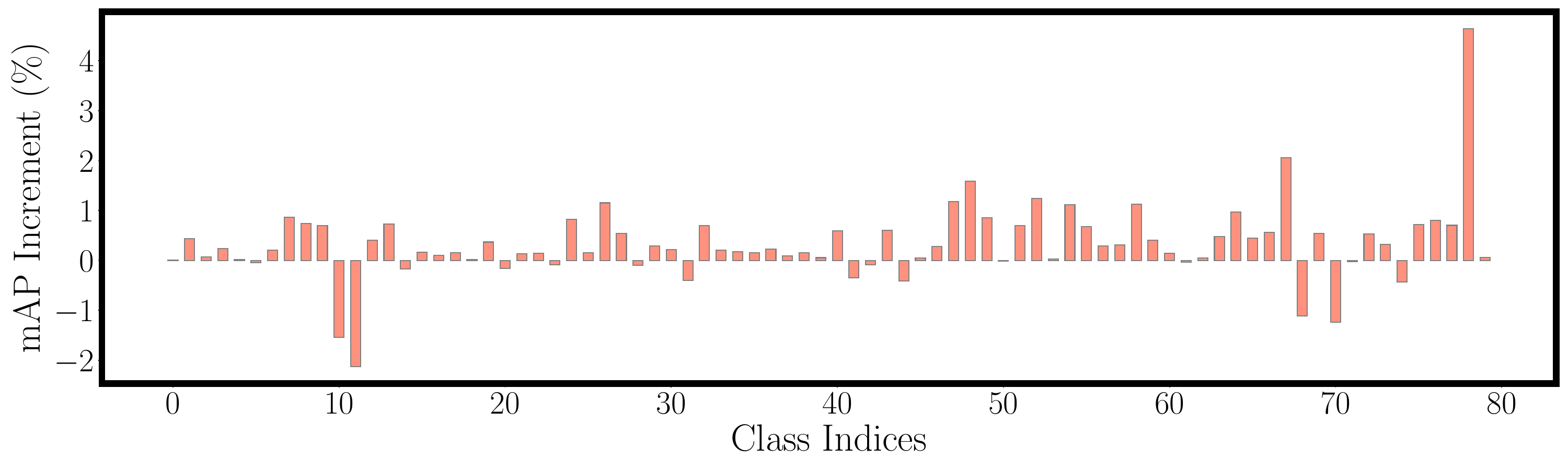}
            \vspace{-0.5em}
		\caption{GVPT vs ML-VPT on \texttt{COCO}}
		\label{fig:vpt_group_moe_per_class}
	\end{subfigure}
	\hfill
        \vspace{-0.5em}
	\caption{Per-class mAP increment for Mixture-of-experts.}
    \label{fig:app_moe_per_class}
\end{figure}

\noindent \textbf{Effect of Mixture-of-Experts} In this work, mixture of experts(MoE) is employed to allocate group-aware representations to label-aware representations, aiming to improve classification performance. To evaluate the effectiveness of the MoE strategy, we present the mAP increment when MoE is incorporated, compared to its absence, as depicted in \cref{fig:app_moe_per_class}, MoE demonstrates a beneficial effect for 93.75\% of the labels, with only 5 labels exhibiting a slight decrement. 

\noindent \textbf{Gating Network Strategy: Label-Aware or Group-Aware.}
To compare the performance of building gating networks at both the label-aware and group-aware, we conducted the following experiments on the \texttt{COCO} dataset using various pre-trained models. The reported results represent the average performance of each method across these models.
As illustrated in \cref{apptab:group_or_label}, the label-aware gating network strategy outperforms the group-aware strategy. This superiority is attributed to the label-aware gating network's ability to select group-aware representations that are most appropriate for the current class, based on the image-specific. In contrast, the group-aware strategy does not account for this selection.
Note that in this work, we choose the label-aware gating networks.
\begin{table}[h]
\centering{
\caption{Comparison of two ways to build gating networks on \texttt{COCO}. All metrics are in \%.}
\vspace{-1em}
\renewcommand\arraystretch{0.50}
\resizebox{0.85\linewidth}{!}{
    \begin{tabular}{c|c|c|c}
    \toprule
    Method        & Avg. mAP  & Avg. CF1  & Avg. OF1  \\
    \midrule
    Group Level   & 79.96     & 73.93     & 77.32  \\
    Label Level   & 80.60     & 74.41     & 77.64  \\ 
    \bottomrule                      	 
    \end{tabular}
}
\label{apptab:group_or_label}
}
\end{table}

\noindent \textbf{Randomization Grouping strategies.}
The results, which are averaged across multiple pre-trained models (including ViT \cite{dosovitskiy2020image}, ViT-21k \cite{dosovitskiy2020image}, MAE \cite{he2022masked}, MoCo v3 \cite{chen2021empirical}, DINOv2/S \cite{dinov22023oquab}, and DINOv2/B \cite{dinov22023oquab}), are presented in \cref{apptab:group_strategies}. Our grouping strategy outperforms all others, including random grouping. 
Note that in this experiment, we do not consider using MoE.
We hypothesize that random grouping might somewhat balance relevance and discrimination relationships; however, it is not the optimal strategy.
\begin{table}[h]
\centering{
\caption{Comparison between multiple grouping strategies on \texttt{COCO}. All metrics are in \%.}
\vspace{-1em}
\renewcommand\arraystretch{0.50}
\resizebox{0.85\linewidth}{!}{
    \begin{tabular}{c|c|c|c}
    \toprule
    Method               & Avg. mAP  & Avg. CF1  & Avg. OF1  \\
    \midrule
    Random-Group         & 78.39     & 72.67     & 76.40  \\
    CO-Group             & 77.79     & 72.21     & 75.97  \\
    DC-Group             & 77.96     & 72.30     & 76.08  \\
    CO-Group\&DC-Group   & 78.84     & 73.00     & 76.58  \\
    \bottomrule                      	 
    \end{tabular}
}
\vspace{-1em}
\label{apptab:group_strategies}
}
\end{table}

\subsection{More Case Study}
\noindent \textbf{Visualization of Group Heatmap.}
To demonstrate that our method effectively learns group-aware representations (group-aware representations) through our grouping strategy, we present a visualization of the group heatmap in \cref{fig:app_group_vis}. These results show that our method can model the relationship between relevant labels and discriminative labels.

\noindent \textbf{Weights Assigned to Experts.}
As shown in \cref{apptab:moe}, the proposed MoE effectively assigns distinct weights to different classes across various images. For instance, although {\ttfamily\text{bottle}} and {\ttfamily\text{broccoli}} are grouped within the same group, the weights required for these two classes by the three experts differ significantly.

\begin{table}[!ht]
\centering{
\caption{Weights assigned to experts for different classes on \texttt{COCO}.
From left to right, they are the image ID, class name, and the weight assigned by the corresponding expert to the corresponding class.
}
\vspace{-1em}
\renewcommand\arraystretch{0.50}
\resizebox{0.90\linewidth}{!}{
    \begin{tabular}{c|c|c|c|c}
    \toprule
    Image ID                     & Classes Name  & Expert 1  & Expert 2  & Expert 3 \\
    \midrule
    000000001000                 & person        & 3.0e-06   & 9.9e-01   & 3.2e-04  \\
    000000010056                 & car           & 9.5e-01   & 2.2e-02   & 2.5e-02  \\
    000000100000                 & cat           & 9.9e-01   & 3.3e-05   & 1.1e-04  \\
    000000100132                 & fork          & 4.4e-04   & 1.9e-03   & 9.9e-01  \\
    000000100238                 & bottle        & 5.1e-01   & 3.1e-01   & 1.8e-01  \\
    000000100624                 & car           & 2.3e-05   & 9.9e-01   & 2.9e-03  \\
    000000100582                 & fork          & 4.3e-01   & 3.1e-01   & 2.5e-01  \\
    000000100811                 & person        & 2.5e-02   & 6.1e-01   & 3.6e-01  \\
    000000113294                 & boat          & 1.5e-01   & 8.5e-01   & 6.8e-05  \\
    000000214919                 & broccoli      & 2.7e-05   & 1.8e-04   & 9.9e-01  \\
    \bottomrule                      	 
    \end{tabular}
}
\vspace{-1em}
\label{apptab:moe}
}
\end{table}
\end{document}